\def\mode{0}

\if 0\mode
    \documentclass[sigconf, nonacm = true]{acmart}
\else
    \documentclass[sigconf]{acmart}
\fi

\usepackage{amsmath, amsthm, mathtools}
\usepackage{subcaption}
\usepackage{enumitem}
\usepackage{makecell}
\usepackage{hyperref}
    
\usepackage[ruled]{algorithm2e} 

\SetAlFnt{\small}
\SetAlCapFnt{\small}
\SetAlCapNameFnt{\small}
\SetAlCapHSkip{0pt}

\newcommand {\newtext}[1]{#1}
\newcommand {\mycaption}[1]{#1}

\DeclareMathOperator*{\argmin}{arg\,min}

\newcommand {\omegain}{\omega_{\mathrm{i}}}
\newcommand {\omegaout}{\omega_{\mathrm{o}}}

\newcommand{\Staircase}{\textsc{Staircase}\xspace}
\newcommand{\Kitchen}{\textsc{Kitchen}\xspace}
\newcommand{\LivingRoom}{\textsc{Living Room}\xspace}
\newcommand{\Living}{\textsc{Living}\xspace}
\newcommand{\Room}{\textsc{Room}\xspace}
\newcommand{\VeachDoor}{\textsc{Veach Door}\xspace}
\newcommand{\Veach}{\textsc{Veach}\xspace}
\newcommand{\Door}{\textsc{Door}\xspace}
\newcommand{\Lego}{\textsc{Lego}\xspace}
\newcommand{\Hotdog}{\textsc{Hotdog}\xspace}
\newcommand{\Ficus}{\textsc{Ficus}\xspace}
\newcommand{\Dragon}{\textsc{Dragon}\xspace}

\acmSubmissionID{papers762}
\copyrightyear{2023} 
\acmYear{2023} 
\setcopyright{rightsretained} 
\acmConference[SIGGRAPH '23 Conference Proceedings]{Special Interest Group on Computer Graphics and Interactive Techniques Conference Conference Proceedings}{August 6--10, 2023}{Los Angeles, CA, USA}
\acmBooktitle{Special Interest Group on Computer Graphics and Interactive Techniques Conference Conference Proceedings (SIGGRAPH '23 Conference Proceedings), August 6--10, 2023, Los Angeles, CA, USA}
\acmDOI{10.1145/3588432.3591553}
\acmISBN{979-8-4007-0159-7/23/08}

\begin{document}

\title{Inverse Global Illumination using a Neural Radiometric Prior}


\author{Saeed Hadadan}
\affiliation{%
  \institution{University of Maryland, College Park}
  \institution{NVIDIA}  
  \state{MD}
  \postcode{20740}
  \country{USA}}
\email{saeedhd@umd.edu}

\author{Geng Lin}
\affiliation{%
 \institution{University of Maryland, College Park}
 \state{MD}
 \country{USA}}
\email{geng@umd.edu}

\author{Jan Novák}
\affiliation{%
 \institution{NVIDIA}
 \country{Czech Republic}}
\email{jnovak@nvidia.com}

\author{Fabrice Rousselle}
\affiliation{%
 \institution{NVIDIA}
 \country{Switzerland}}
\email{frousselle@nvidia.com}

\author{Matthias Zwicker}
\affiliation{%
 \institution{University of Maryland, College Park}
 \state{MD}
 \country{USA}}
\email{zwicker@cs.umd.edu}

\renewcommand\shortauthors{Hadadan, S. et al}

%
%


%
%

\begin{abstract}
Inverse rendering methods that account for global illumination are becoming more popular, but current methods require evaluating and automatically differentiating millions of path integrals by tracing multiple light bounces, which remains expensive and prone to noise. Instead, this paper proposes a radiometric prior as a simple alternative to building complete path integrals in a traditional differentiable path tracer, while still correctly accounting for global illumination. Inspired by the Neural Radiosity technique, we use a neural network as a radiance function, and we introduce a prior consisting of the norm of the residual of the rendering equation in the inverse rendering loss. \newtext{We train our radiance network and optimize scene parameters simultaneously using a loss consisting of both a photometric term between renderings and the multi-view input images, and our radiometric prior (the residual term).} This residual term enforces a physical constraint on the optimization that ensures that the radiance field accounts for global illumination. We compare our method to a vanilla differentiable path tracer, and more advanced techniques such as Path Replay Backpropagation. Despite the simplicity of our approach, we can recover scene parameters with comparable and in some cases better quality, at considerably lower computation times.
\end{abstract}
\begin{teaserfigure}
\centering
  \includegraphics[width=\textwidth]{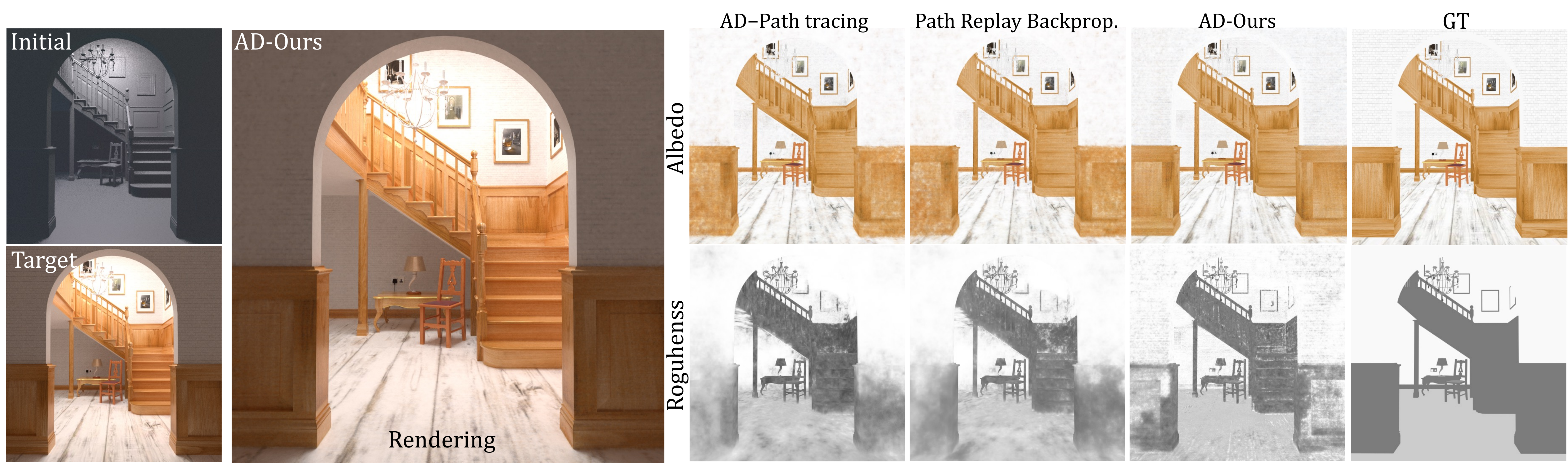}
  \caption{\mycaption{We propose an inverse rendering method that uses a radiometric prior to account for global illumination as opposed to building and differentiating path integrals. Our method uses standard automatic differentiation (AD) to compute gradients with respect to the scene parameters, while satisfying the rendering equation using our radiometric prior, which is represented by a neural network. Here we compare a traditional auto-differentiable path tracer (AD-PT), an advanced technique (Path Replay Backpropagation, or PRB), and our method (AD-Ours) for recovering non-diffuse spatially varying BRDF properties (also represented as neural networks) under known illumination and geometry from 26 views of the \Staircase scene. Despite its simplicity, our approach takes into account global illumination, and recovers albedo and roughness with similar accuracy as differentiable path tracing and PRB. \newtext{Each method used a total of 16384 $\times$ 16 $\times$ 18000 (batch size $\times$ spp $\times$ steps) = 4.7B samples, i.e., 690 training samples per pixel (26 views $\times$ 512 $\times$ 512 pixels). We conducted all experiments with a single RTX3090 GPU, and the total runtimes for AD-PT, PRB, and our method were 760, 970, and 260 minutes,} respectively.}}
  \label{fig:teaser}
\end{teaserfigure}

\maketitle

\section{Introduction}

Inverse rendering---the task of recovering the parameters of a 3D scene from 2D images---has seen rapid progress and adoption in recent years.
Two key components of an inverse rendering approach are a \emph{differentiable renderer} and an algorithm for propagating gradients of the objective function to scene parameters. Differentiable rendering is a challenging problem, especially if global illumination effects are to be taken into account, since this process typically involves Monte Carlo path tracing and calculating derivatives of millions of path integrals with respect to scene parameters.


In principle, path integrals can be differentiated using standard \emph{automatic differentiation} (AD), a universal approach that constructs a graph of operations, which is then traversed in a second pass to compute gradients of the objective function with respect to scene parameters.
However, AD suffers from high memory requirements when differentiating multi-bounce path integrals in complex scenes, which limits its use to simple light transport models, such as direct illumination or one-bounce indirect illumination. Yet such simplified light transport models cannot explain appearance changes caused by multiple interreflections, and as a result, the ignored interreflections will be ``baked'' into the recovered scene parameters.

To reduce the memory requirements of AD, a group of algorithms called \emph{radiative back-propagation} (RB) 
develop a problem-specific automatic differentiation approach that removes the need to store the enormous graph of operations in standard AD. Thus RB is more efficient in its use of memory, but relies on sophisticated light transport algorithms that require familiarity with transport theory and have not yet been fully commoditized.
The main benefit of RB, thanks to its light transport foundation, is its ability to accurately reconstruct the scene under a global illumination rendering model.

Instead, our main contribution is a simple approach for differentiable rendering under a global illumination model that does not require computing multi-bounce light path integrals. \newtext{We leverage a neural network both to represent the radiance function, and to account for global illumination without building path integrals. We achieve this via a radiometric prior, inspired by Neural Radiosity \cite{hadadan2021neural}, that represents the norm of the residual of the rendering equation ~\cite{kajiya_rendering_equation}. To recover scene parameters, we simultaneously optimize our radiance network and the scene parameters, using a loss that takes into account both the radiometric prior and the difference between the multi-view ground truth images and rendered images. Because this process does not require computing multi-bounce light paths, we can efficiently differentiate it using standard automatic differentiation to compute gradients. 

The radiometric prior achieves the following benefits: (1) it ensures that the radiance field represented by our radiance network satisfies the rendering equation and accurately accounts for global illumination, and the optimization of scene parameters is guided towards a physically valid solution; (2) it enables automatic differentiation to yield accurate results without having to simulate many bounces of light using path integrals, which is key to curbing the memory requirements of AD and also results in significant performance gains; (3) it allows us to achieve accurate reconstruction quality similar as radiative back-propagation, but with the simple and straightforward approach of standard automatic differentiation.
Please note we will use the terms radiometric prior, radiometric term, and residual term interchangeably throughout the paper.
}


{\em Scope.}
Our work presents a contribution that lifts some of the constraints of inverse rendering under a global illumination model with standard automatic differentiation.  We analyze the method on synthetic scenes, reconstructing either the material parameters, or lighting; geometry reconstruction and joint optimization are outside of our scope. We use a neural radiance field instead of Monte Carlo estimators of (differential) radiance. Our gradients are thus biased, yet the optimization reaches comparable reconstruction accuracy, often faster than prior works. \newtext{Explicitly handling  visibility-related discontinuities are outside of our scope, but we believe that the smoothing property of our radiance network can reduce the negative effects of discontinuities in the integrands of the rendering equation.}


\section{Related Work}
For brevity, this section focuses on differentiable rendering techniques, radiance field representations, and neural inverse rendering techniques.
We refer to existing resources for an overview of classical~\cite{PBRT3e} and neural~\citep{Tewari:STAR:2020,Tewari:STAR:2022} rendering.

{\em Differentiable Rendering.}
Many techniques in visual computing rely on differentiable rendering to extract scene properties from images.
Techniques building on differentiable rasterizers
~\cite{versatile_scene_model,OpenDR, Kato_2018_CVPR, liu2019soft, petersen2019pix2vex, laine2020modular}
are fast, but do not produce accurate results as effects due to indirect illumination tend to be incorrectly attributed to scene parameters.
Naively differentiating advanced rendering algorithms, such as path tracing, via automatic differentiation is severely limited by the memory requirements of AD.
\emph{Radiative Backpropagation} (RB)~\cite{radiative} sidesteps this issue via an adjoint approach that propagates gradients of the objective function to scene parameters using Monte Carlo path sampling.
The quadratic cost of the algorithm has been addressed by Vicini et al.~\shortcite{path_replay} (PRB), who propose to ``replay'' paths to save cost. 
Both of these methods yield accurate scene reconstructions, but their implementation is intricate and  sampling strategies for building paths optimally are not yet fully developed. 

Conversely, our method accounts for global illumination by enforcing a radiometric prior instead of building and differentiating complete, multi-bounce path integrals, and hence we can use standard AD to compute gradients. Automatic differentiation and PRB are among the baselines we compare our method to and there is more discussion about it throughout this paper. 

Recently, a neural-network based representation of the \emph{differential radiance} was proposed by Hadadan et al. \shortcite{hadadan2022differentiable}, which is also inspired by Neural Radiosity. \newtext{However, the two methods are very different. The prior work requires an adjoint estimator and caches \emph{differential} radiance. As a result, it needs one network output per scene parameter and scales poorly to practical scenarios. In contrast, our method caches radiance, and can optimize large parameter sets as we only require AD.}


{\em Radiance field representations.}
The computation of radiometric quantities is a significant burden in rendering, which also applies to the computation of our radiometric prior.
Various caching methods have been proposed to tackle this problem, starting with the seminal work of ~\citet{Ward:1988:Irradiance} on diffuse interreflection, later extended to volumes~\cite{Greger:1998:Irradiance} and glossy surfaces~\cite{Krivanek:2005:Radiance}.
A common challenge for these techniques is the design of cache data structures, as these must accommodate well to any scene geometry and support view-dependent queries.
We also utilize a caching mechanism, but we sidestep the need for a sophisticated data structure by representing the radiance cache with a neural network as proposed in previous works~\cite{Ren:2013:rrf,nerf,Mueller:2020:ncv,Mueller2021NRC}.
Specifically, our radiometric prior builds upon the Neural Radiosity model introduced by~\citet{hadadan2021neural}.

{\em Neural inverse rendering.}
NeRF, introduced by \citet{nerf}, proposes to represent a scene as a neural radiance and density field encoded with an MLP.
While this approach achieves state-of-the-art accuracy in novel view synthesis of complex scenes, it effectively bakes the scene parameters, such as reflectance and lighting, into the neural representation.
The NeRV method of~\citet{nerv} proposes to separately recover the reflectance of the scene elements, while accounting for indirect illumination, to enable relighting applications.
Their technique, however, explicitly differentiates the path integral and is therefore limited in practice to one-bounce indirect illumination to fit a reasonable memory and computational budget.
In contrast, our radiometric prior captures the steady-state of light transport, thereby modelling the full scope of indirect illumination at a fixed memory and computational overhead.
Another work, NerFactor~\citep{nerfactor} extracts a geometric representation from the NeRF density field and proceeds to recover the albedo, visibility and surface normals in the scene, albeit under a simplified direct illumination model.
More recently, Zhang et al. \shortcite{inverse_rendering_gi} proposed to estimate the global illumination from a radiance field obtained with an existing off-the-shelf technique.
The accuracy of their approach, however, is inherently bounded by the accuracy of the underlying neural radiance field, whereas our approach explicitly applies physically-based constraints to the optimization of scene parameters.

\newtext{
{\em Visibility-related Discontinuities.}
Differentiating the rendering equation without taking into account discontinuities may result in incorrect gradients, and special treatment is necessary for discontinuities caused by occlusions \cite{edgesampling,reparam, Path-Space, BangaruMichel2021DiscontinuousAutodiff}. Li et al~\shortcite{edgesampling} have addressed this issue by separating the rendering equation into continuous and discontinuous parts, and proposing a silhouette edge sampling approach. Loubet et al ~\shortcite{reparam} have suggested a reparametrization of the rendering integrals so that the positions of discontinuities are independent of the scene parameters. Although this paper does not explicitly address discontinuities, and therefore does not study the issue, the authors suggest that the smoothing property of a radiance neural network may help mitigate the problems arising from not handling discontinuities in the pixel and light integrals.
}
\section{Background}

For differentiable rendering under a global illumination model, we build on the rendering equation~\cite{kajiya_rendering_equation},
which defines the radiometric equilibrium between the outgoing, emitted, and incident radiance that is scattered at each point in the scene, 
\begin{multline}
\label{eq:rendering-equation}
L(x,\omegaout) = E(x,\omegaout) + \int_{\mathcal{H}^2} f(x,\omegain, \omegaout) L(r(x,\omegain),-\omegain) d\omegain^{\perp},
\end{multline}
where $x$ is a surface point, $\omegain$ and $\omegaout$ are directions of incidence and exitance at $x$, $L(x,\omegaout)$ is the outgoing radiance as a function over surface locations and outgoing directions, $E(x,\omegaout)$ is emitted radiance, $f(x,\omegain, \omegaout)$ is the bidirectional reflectance distribution function (BRDF), $r(x,\omegain)$ is the ray tracing operator returning the closest surface intersection of ray $(x,\omegain)$, 
$\mathcal{H}^2$ is the upper hemisphere, and $d\omegain^{\perp}$ is the differential projected solid-angle measure.

The rendering equation is written more concisely in operator notation as $L(x,\omegaout) = E(x,\omegaout) + \mathcal{T}(L)(x,\omegaout)$, where the transport operator $\mathcal{T}$ represents the hemispherical integral in Equation~\eqref{eq:rendering-equation}. Its solution is given by the Liouville–Neumann series $L(x,\omegaout)=\sum_{i=0}^{\infty} \mathcal{T}^i(E)(x,\omegaout)$, where each term $\mathcal{T}^i(E)(x,\omegaout)$ is a $2i$-dimensional integral over multi-bounce light paths with $i$ segments. 

Denoting an image consisting of a set of pixels as $I=\{I_k\}$, each pixel $I_k$ is given by the so-called measurement integral,
\begin{align}
\label{eq:measurement-equation}
I_k = \int_{\mathcal{A}}\int_{\mathcal{H}^2} W_k(x,\omega) L(r(x,\omega),-\omega) dx d\omega^{\perp},
\end{align}
where $W_k(x,\omega)$ models the response of a sensor pixel to incident radiance over its area $\mathcal{A}$ and the hemisphere of directions $\mathcal{H}^2$. In practice, the integrals in Equation~\eqref{eq:measurement-equation} are most commonly estimated using Monte Carlo sampling and path tracing.

\subsection{\newtext{Neural Radiosity}}
\newtext{
Instead of building on the series expansion and path integrals, Neural Radiosity \cite{hadadan2021neural} finds a solution of the rendering equation using a neural network, without requiring multi-bounce light path integrals. The radiance function $L(x,\omega_o)$ in Equation \eqref{eq:rendering-equation} is represented by a neural network with a set of parameters $\theta$, denoted as $L_{\theta}(x,\omega_o)$, that takes location $x$ and direction $\omega_o$ and returns the outgoing radiance. Neural Radiosity then determines the network parameters $\theta$ by minimizing the norm  of the \emph{residual} of the rendering equation, which is
\begin{multline}
\label{eq:renderingresidual}
r_{\theta}(x,\omega_o) =  L_{\theta}(x,\omega_o) - E(x,\omega_o) \\
 - \int_{\mathcal{H}^2} f(x,\omega_i, \omega_o) L_{\theta}(x'(x,\omega_i),-\omega_i) d\omega_i^{\perp},
\end{multline}
that is, the difference between the left and right-hand sides of Equation \eqref{eq:rendering-equation} when the radiance function $L$ is substituted by $L_{\theta}$. Monte Carlo sampling is used to estimate the norm of the residual as in
\begin{align}
    \mathcal{L}(\theta) \approx \frac{1}{N}\sum_{j=1}^{N} \frac{r_{\theta}(x_j,\omega_{o,j})^2}{p(x_j,\omega_{o,j})},
\end{align}
where $N$ is the number of samples, and samples of surface locations $x_j$ and outgoing directions $\omega_{o,j}$ are distributed according to probability density $p(x,\omega)$.

%

%

Neural Radiosity is a self-training approach, that is, there is no use of ground truth data to supervise the training. 
The resulting radiance network accounts for global illumination effects without integrating over the path space.
Our method builds on Neural Radiosity as we will elaborate in Section~\ref{sec:method}.
}
\subsection{Inverse Rendering under Global Illumination}
\label{inverse rendering}

Inverse rendering is the problem of finding unknown scene parameters $\phi$ in 3D such that, under a given rendering model, they match with a set of given 2D images. We denote an image rendered using certain scene parameters $\phi$ as $I(\phi)$, where $\phi$ may represent surface geometry, appearance, and lighting.

The objective of inverse rendering is to recover optimal parameters $\phi^*$ by minimizing a photometric loss function $\mathcal{L}(I(\phi))$, 
\begin{align}
\label{eq:inverse-rendering-objective}
\phi^* = \argmin_{\phi} \mathcal{L}(I(\phi)),
\end{align}
where $\mathcal{L}(I(\phi))$ quantifies the similarity of a rendered image $I(\phi)$ and a given input image $I^{\mathrm{GT}}$. For simplicity our notation assumes a single input image, and the photometric loss can be defined for example as the $L_2$ distance $\mathcal{L}(I(\phi)) = \|I(\phi) - I^{\mathrm{GT}}\|$, but in practice images from multiple viewpoints are typically involved.

In a practical approach, we minimize the loss $\mathcal{L}(I(\phi))$ using a gradient-based optimization algorithm, and in principle, the gradient with respect to scene parameters $\partial \mathcal{L}(I(\phi)) / \partial \phi$ can be computed using standard automatic differentiation. Because rendering $I(\phi)$ using Equation~\eqref{eq:measurement-equation} requires computing millions of pixels and path integrals, however, standard AD creates large computation graphs that easily exceed available memory.

{\em Adjoint Methods.}
To reduce memory requirements of standard AD, a group of algorithms called \emph{radiative backpropagation} \cite{radiative} use the chain rule to separate the computation of an adjoint image, consisting of the derivatives of the loss with respect to image pixels, and the differentiation of image pixels with respect to scene parameters,
\begin{align}
\label{eq:inverse-chain-rule}
\frac{\partial \mathcal{L}}{\partial \phi} = \frac{\partial \mathcal{L}}{\partial I}. \frac{\partial I}{\partial \phi}.
\end{align}
%
These methods require differentiated versions of Equations~\eqref{eq:rendering-equation} and \eqref{eq:measurement-equation} with respect to the scene parameters $\phi$,
\begin{align}
\label{eq:differential-measurement-equation}
\partial_{\phi}I_k = \int_{\mathcal{A}}\int_{\mathcal{H}^2} W_k(x,\omega) \partial_\phi L(x,\omega) dx d\omega^{\perp}
\end{align}
and 
\begin{equation}
\begin{split}
\label{eq:differntial-rendering-equation}
\partial_{\phi}L(x,\omega_o) = \partial_{\phi}E(x,\omega_o) + \int_{\mathcal{H}^2} f(x,\omega_i, \omega_o) \partial_{\phi}L(x'(x,\omega_i),-\omega_i) d\omega_i^{\perp} \\
+ \int_{\mathcal{H}^2} \partial_{\phi}f(x,\omega_i, \omega_o) L(x'(x,\omega_i),-\omega_i) d\omega_i^{\perp},
\end{split}
\end{equation}

where $\partial_\phi$ is a short form for $\partial / \partial_\phi$. Nimier David~\shortcite{radiative} call the latter the \emph{differential rendering equation}.

The differential rendering equation can be solved using a series expansion similar as the original rendering equation, leading to an expression for $\partial_{\phi}L$ given by the path integrals
\begin{align}
\label{eq:differential-rendering-equation-series}
\partial_{\phi}L = \sum_{i=0}^{\infty} \mathcal{T}^i(\partial_{\phi}E) + \sum_{i=0}^{\infty} \mathcal{T}^i(\mathcal{T}'(\sum_{k=0}^{\infty} \mathcal{T}^k(E))),
\end{align}
where $\mathcal{T}'$ is the operator notation for the second integral in Equation~\eqref{eq:differntial-rendering-equation}. Path replay backpropagation (PRB)~\cite{path_replay} samples and evaluates these integrals efficiently by reusing computations in a clever way. 
In particular, one first evaluates a so-called primal path to compute samples of the term $\sum_k \mathcal{T}^k(E)$ (one sample for each path length $k$, paths are terminated using Russian roulette as usual), which is the same as the series expansion of the usual rendering equation. Then, one reuses this same path together with various terms evaluated and stored along the path, to evaluate samples of the other integrals in Equation~\eqref{eq:differential-rendering-equation-series}. 

Even with advanced techniques such as PRB, the main issue remains to be the need to build long path integrals to correctly and efficiently evaluate the integrands in Equation~\eqref{eq:differential-rendering-equation-series} by reusing computations. In addition, variance in the Monte Carlo estimates will lead to noisy gradients that may lead to slow convergence.

\section{Method}
\label{sec:method}

\begin{figure}
    \centering
    \includegraphics[width=0.49\textwidth]{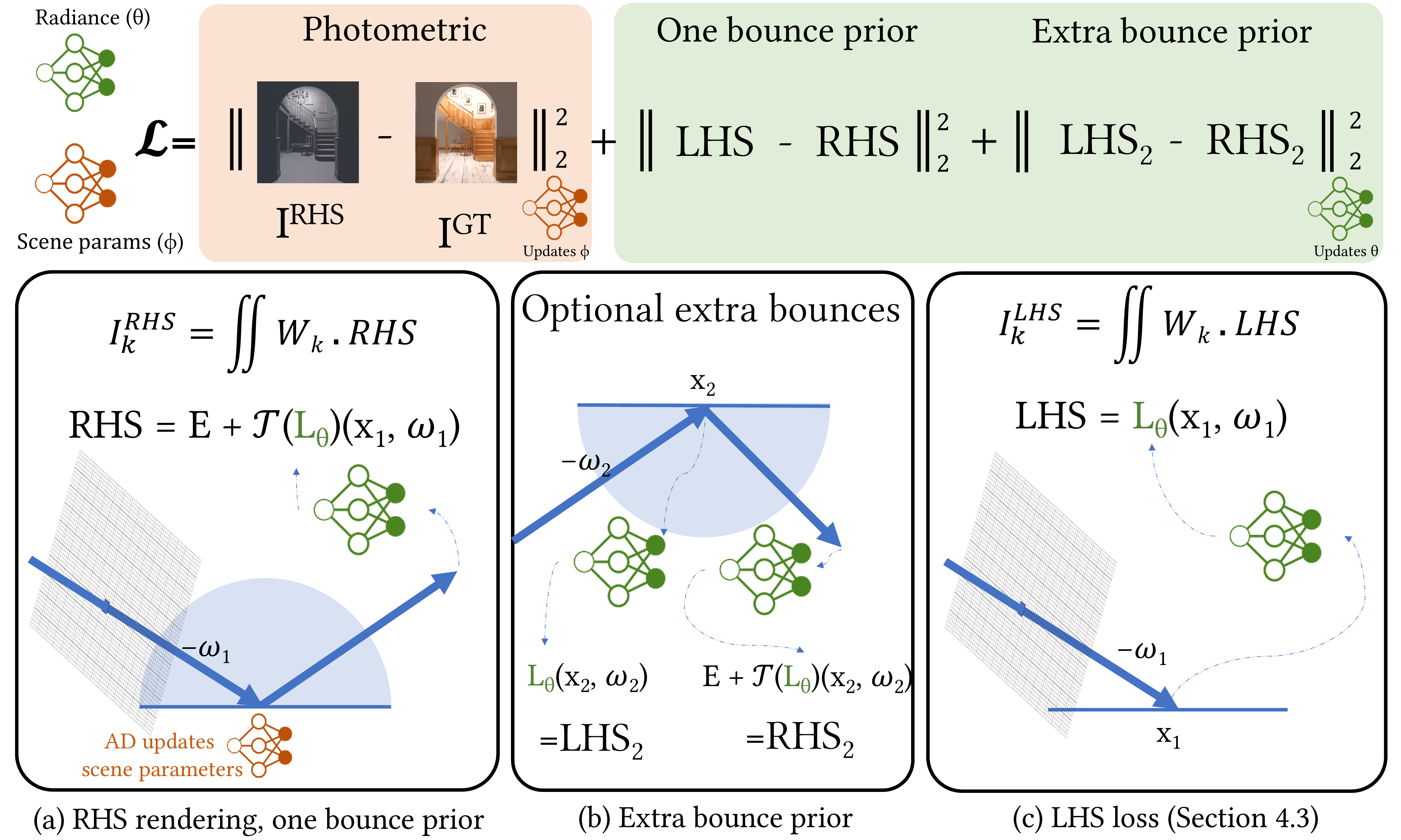}
    \vspace{-1mm}
    \caption{\textbf{Method overview}. \mycaption{ \textbf{(a)} We use RHS rendering in the photometric loss (top-left) whose gradients are back-propagated using AD to the scene parameters at the primary bounce.
    The radiometric prior (top right) acts as a regularizer to ensure the solution satisfies the rendering equation, thereby accounting for global illumination; its gradients are back-propagated to the radiance network.
    \textbf{(b)} The prior can model global illumination using a single bounce, but it is not restricted to it. Additional radiometric terms can be defined and optimized at later bounces to impose radiometric constraints on locations not directly observed by the camera (Section \ref{sec:sampling}).
    \textbf{(c)} Computing a photometric loss based on LHS can further improve results (Section \ref{sec:lhs-loss}).}}
    \label{fig:overviewfig}
    \vspace{-2mm}
\end{figure}

We avoid the drawbacks above by completely circumventing the estimation of path integrals to solve the (differential) rendering equation. Instead, our key idea is that we use the rendering equation as a radiometric prior, which we will add as an additional loss term to the inverse rendering optimization objective in Equation~\eqref{eq:inverse-rendering-objective}. 

To formulate the radiometric prior, we represent the radiance field in the scene as a parametric function $L_{\theta}(x,\omega_o)$ implemented as a neural network, where $\theta$ represents the network weights. 
Similarly to \citet{hadadan2021neural}, the radiometric prior $\mathcal{L}_{\mathrm{prior}}$ 
is given by the norm of the residual of the rendering equation, 
\begin{align}
\label{eq:radiometric prior}
    \mathcal{L}_{\mathrm{prior}}(\theta) = \| L_{\theta}(x,\omega_o) - (E(x,\omega_o) + \mathcal{T}(L_{\theta})(x,\omega_o))\|.
\end{align}
The residual is simply the difference between the left-hand side (LHS) and right-hand side (RHS) of Equation~\eqref{eq:rendering-equation}. Note that the prior depends both on the radiance field and scene parameters, $\theta$ and $\phi$ respectively, although we omit the dependence on $\phi$ in our notation for simplicity.
For known scene parameters, \citet{hadadan2021neural} have shown that minimizing this term on its own can be used to solve the rendering equation.
In contrast, we leverage this term for inverse rendering to recover scene parameters that are unknown. 

In our inverse rendering approach, we simultaneously minimize the photometric loss $\mathcal{L}(I(\phi))$ to ensure the rendering $I(\phi)$ matches the input image, while enforcing the radiometric prior $\mathcal{L}_{\mathrm{prior}}(\theta)$ to ensure the radiance field satisfies the rendering equation, 
\begin{align}
\label{eq:objective}
\phi^*, \theta^* = \argmin_{\phi,\theta} \mathcal{L}(I(\phi)) + \mathcal{L}_{\mathrm{prior}}(\theta).
\end{align}
Crucially, having the radiance field $L_{\theta}(x,\omega_o)$, which is constrained by the radiometric prior, allows us to formulate a simple image formation model  $I(\phi) = \{I_k\}$ that correctly captures global illumination. Instead of using Equation~\ref{eq:measurement-equation}, which requires estimating path integrals to calculate the radiance $L$, we compute pixels using the neural radiance field $L_{\theta}(x,\omega_o)$ as
\begin{align}
\label{eq:RHS-measurement}
I_k = \int_{\mathcal{A}}\int_{\mathcal{H}^2} W_k (E + \mathcal{T}(L_{\theta})(x,\omega)) dx d\omega^{\perp} \eqqcolon I^{\mathrm{RHS}}_k.
\end{align}
%
The rendering $I(\phi)$ and its pixels $I_k$ depend on scene and radiance field parameters, $\phi$ and $\theta$ respectively, but we keep our notation as before for simplicity. We also call this {\em RHS rendering} (Figure~\ref{fig:overviewfig}), because it involves evaluating the right-hand side (RHS) of Equation~\ref{eq:rendering-equation}. 

As a key advantage, the RHS rendering model does not involve any path integrals or path tracing. As a consequence, we can easily compute the derivatives $\partial \mathcal{L}(I(\phi)) / \partial{\phi}$ with standard AD and we do not need to use the adjoint approach.

\subsection{Gradient Computation}
\label{sec:stopgrad}

As mentioned above, the radiometric prior both depends on the radiance field and scene parameters, $\theta$ and $\phi$. However, the prior needs to be satisfied for any choice of scene parameters; that is, irrespective of the scene parameters, the radiance field needs to satisfy the rendering equation. Hence we need to treat the scene parameters as given constants when minimizing the prior.
We also observed experimentally that if the radiometric prior influences the scene properties $\phi$, this detrimentally affects our results (Figures \ref{fig:ablation-materials} and \ref{fig:ablation-envmap}). In practice, we block backpropagating gradients from the radiometric prior term to the scene parameters, and we update scene parameters only using gradients of the photometric loss $\mathcal{L}(I(\phi))$. Similarly, the photometric loss depends both on $\theta$ and $\phi$. However, we obtain better results in practice by not backpropagating gradients of $\mathcal{L}(I(\phi))$ to radiance field parameters $\theta$.



\subsection{Sampling}
\label{sec:sampling}

We numerically evaluate the integrals in Equation~\eqref{eq:radiometric prior} and~\eqref{eq:RHS-measurement} using Monte Carlo sampling. We sample the pixel measurement and transport operator in Equation~\eqref{eq:RHS-measurement} using standard ray tracing with a given number of samples per pixel. We trace two consecutive ray segments, one from the camera to a primary hit point, and a second one from the primary hit point to a secondary hit point, and we use the two ray segments to evaluate the transport operator. 

In theory, the radiometric prior in Equation~\eqref{eq:radiometric prior} should be estimated by sampling over all surface locations and pairs of incident and outgoing directions (to evaluate the transport operator). For simplicity, however, we reuse the primary hit point and the two ray segments obtained from sampling the measurement equation to also sample the radiometric prior. We call this approach the one bounce prior. While this simple approach fails to minimize the radiometric prior at locations not directly visible from the camera, it often works well in practice and faithfully captures global illumination effects as our experiments show. To improve results at a small additional cost, we can also ray trace a third consecutive ray segment and sample the radiometric prior both at the first and secondary hit points, which leads to better coverage of surface locations. We call this extended approach the extra bounce prior.

\subsection{Ground Truth Radiance Field Constraint}
\label{sec:lhs-loss}
So far, the radiance network $L_{\theta}$ is only being optimized using the radiometric prior, which uses no ground truth data--implementing a 
\emph{self-training} approach similar as in Neural Radiosity. In inverse rendering problems, however, we may have high dynamic range ground truth 
images $I^{\mathrm{GT}}$ that can also be used to constrain the radiance network. To leverage $I^{\mathrm{GT}}$ to constrain $L_{\theta}$, we can 
directly render the radiance field into an image $I^{\mathrm{LHS}}(\theta)$ consisting of pixels 
\begin{align}
\label{eq:LHS-measurement}
I^{\mathrm{LHS}}_k = \int_{\mathcal{A}}\int_{\mathcal{H}^2} W_k L_{\theta}(x,\omega) dx d\omega^{\perp},
\end{align}
and define and additional loss term 
\begin{align}
    \mathcal{L}_{\mathrm{LHS}}(\theta)= \left\|I^{LHS}(\theta) - I^{\mathrm{GT}} \right\|^2.
    \label{eq:lhs-recons-loss}
\end{align}
Our experiments show that adding this loss to our objective in Equation~\eqref{eq:objective} produces small additional improvements (Figures \ref{fig:ablation-materials} and \ref{fig:ablation-envmap}), hence we include this term in our final results. 

\section{Implementation}



{\em Representing Scene Parameters.}
We use the Burley BRDF~\cite{Burley:2012:brdf} in a spatially varying configuration and parameterize its albedo and roughness parameters using separate MLPs that receive a surface location as input.
When optimizing for lighting (Figure \ref{fig:ablation-envmap}), we store it as a learned environment map texture.

{\em Use of Automatic Differentiation.}
Having all terms of the objective computed, we rely on AD to propagate gradients of the objective to the albedo and roughness MLPs, and the gradients of the radiometric term to the radiance MLP. Technically, computing unbiased gradients using AD requires two forward passes to avoid correlation between the derivatives of the loss and the derivatives of the measurement equation. We take the same approach in this paper when using AD with our method, as well as path tracing and direct illumination. Additionally, we used detached sampling \cite{gradestimate} across all AD based methods.



{\em System \& Architecture.}
We implement our differentiable renderer using Mitsuba 3 \cite{jakob2022mitsuba3}, which provides primitives for differentiable rendering using automatic differentiation and radiative backpropagation. 
%
%
We implement our MLPs in PyTorch \cite{pytorch} and embed them inside the Mitsuba renderer, while allowing the gradients to flow between the two frameworks. 
For PRB we use the official Python implementation.
We use an MLP with 3 hidden layers, 256 neurons in each, to represent the radiance field.
The roughness and albedo are each modeled using a dedicated MLP with a single hidden layer and 256 neurons. 
We equip all MLPs with a hash grid~\citep{muller2022instant} to facilitate high spatial resolutions;
we use the following configuration: resolutions = $[2^1-2^{16}]$, features per level = $2$, hash table size = $2^{17}$.
The radiance network receives location, direction, surface normal, and albedo as input. Our code and data are publicly available online\footnote{\url{https://inverse-neural-radiosity.github.io}}.

{\em Training.}
We use the Adam optimizer with learning rate $5 \cdot 10^{-4}$ in all experiments.
At each training step, we select a random view from the scene, and randomly crop to a certain size from the image. For NeRF scenes with transparent background, at least half of the samples in every training batch are foreground pixels. The patch is 
rendered to compute the loss and gradients for the parameter updates. Detailed statistics for all scenes can be found in Table \ref{tab:scene_stats}. We use the relative $L_2$ loss function for the photometric and radiometric terms. 
%
The \mbox{albedo-}, \mbox{roughness-}, and \mbox{radiance-}predicting MLPs are trained jointly, starting with randomly initialized weights.



\section{Results \& Analysis}
\label{sec:results}
In this section, we discuss our experimental setup and compare our proposed automatic differentiation using a radiometric prior (AD-Ours) to three baselines: automatic differentiation of direct illumination (AD-Direct),
path replay backpropagation (PRB)~\citep{path_replay}, \newtext{and an approach, labeled as AD-Ours w/o prior, where we only train the radiance field using the ground truth images (Section~\ref{sec:lhs-loss}), without including the radiometric prior. This is similar in spirit to the idea proposed by Zhang et al. \shortcite{inverse_rendering_gi}}.


{\em Multi-view datasets.}
We evaluated our method on a set of synthetic scenes, including indoor scenes~\cite{benedict_scenes} and scenes from the NeRF dataset~\cite{nerf}. All scenes were altered to use the Burley BRDF. In the \Staircase scene we increased the albedo to evaluate the methods in the presence of strong indirect illumination. For each indoor scene, we rendered a multi-view dataset using the Mitsuba 3 path tracer~\citep{jakob2022mitsuba3}, placing cameras manually to cover the entire scene. For the NeRF scenes, we sampled cameras on the hemisphere surrounding the object.

{\em Experimental setup.}
All results for all methods use the same optimization hyper-parameters. The path length for PRB (and AD-PT) is limited to 15 bounces and paths are terminated using Russian Roulette with probability=0.95~\cite{Arvo:1990:Russian}; we report the mean path length and the number of views used for optimization for each scene in \autoref{tab:scene_stats}.

\begin{table}[t]
    \centering
    \caption{\textbf{Training statistics per scene.}}
    \setlength{\tabcolsep}{1.2pt}
    \begin{small}
    \begin{tabular}{lcccccccc}
        \toprule
        {} & {} & {} & \footnotesize{\Living} & \footnotesize{\Veach} \\ [-3pt]
        {} & \footnotesize{\Staircase} & \footnotesize{\Kitchen} & \footnotesize{\Room} & \footnotesize{\Door} & \footnotesize{\Lego} & \footnotesize{\Hotdog} & \footnotesize{\Ficus} & \footnotesize{\Dragon} \\
        \cmidrule{2-9}
        Mean path len.     & 4.74 & 3.67 & 4.10 & 3.33 & 2.32 & 1.46 & 1.43 & 1.26 \\
        \# of views & 26   & 62   & 64   & 19   & 25   & 25   & 25   & 25   \\
        \newtext{Training spp} & 16   & 64   & 64   & 64   & 64   & 64   & 64   & 64   \\
        \newtext{Batch size} $2^x$ & 14   & 12 & 12   & 12   & 12   & 12   & 12   & 12   \\
        \newtext{Total steps} & 18K   & 32K   & 32K   & 32K   & 10K   & 10K   & 10K   & 10K   \\
        \bottomrule
    \end{tabular}
    \end{small}
    \label{tab:scene_stats}
\end{table}

\begin{figure}
    \vspace{-3mm}
    \centering
    \includegraphics[width=0.47\textwidth]{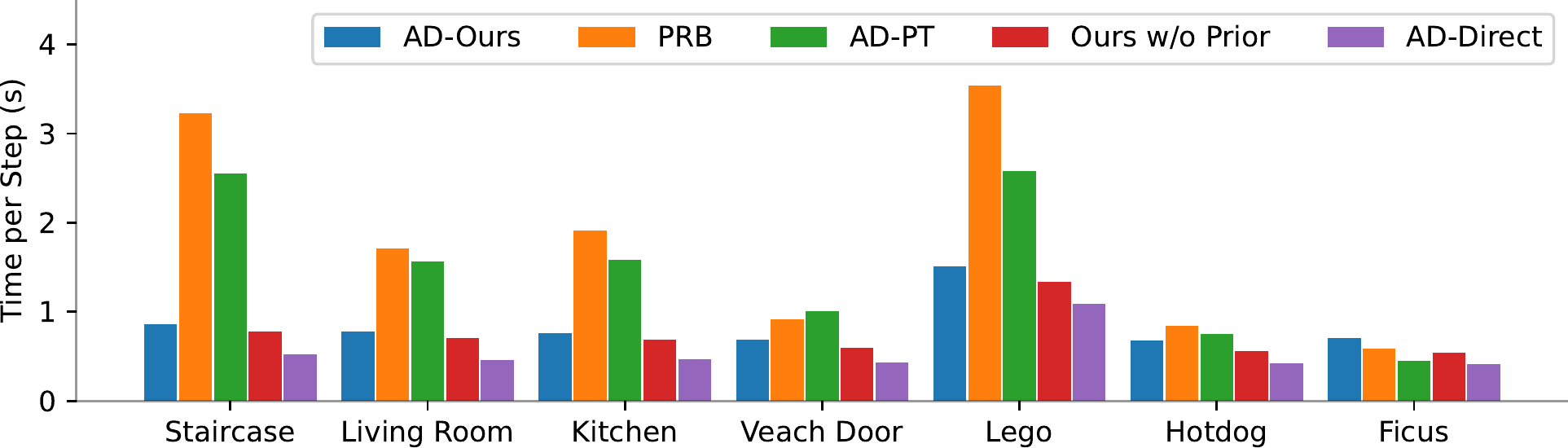}
    \vspace{-3mm}        
    \caption{\newtext{\textbf{Total training time per step.}}}
    \label{fig:train_times}
    \vspace{-3mm}    
\end{figure}


\subsection{Comparison to previous techniques}
\begin{figure*}
    \centering
    \captionsetup[subfigure]{labelformat=empty}
    \begingroup
\renewcommand{\arraystretch}{0.6}

\makebox[5pt]{\rotatebox{90}{\hspace{-10pt} \footnotesize{\Staircase}}}
\begin{subfigure}[b]{0.98\textwidth}
\begin{tabular}{ccccccccccc}

{\footnotesize{AD-Direct}}
 & 
{\footnotesize{PRB}}
 & 
{\footnotesize{\begin{tabular}{@{}c@{}}AD-Ours \\ w/o Prior\end{tabular}}}
 & 
{\footnotesize{AD-Ours}}
 & 
{\footnotesize{GT}}
 &  & 
{\footnotesize{AD-Direct}}
 & 
{\footnotesize{PRB}}
 & 
{\footnotesize{\begin{tabular}{@{}c@{}}AD-Ours \\ w/o Prior\end{tabular}}}
 & 
{\footnotesize{AD-Ours}}
 & 
{\footnotesize{GT}}
 \\

\includegraphics[width=0.077\textwidth]{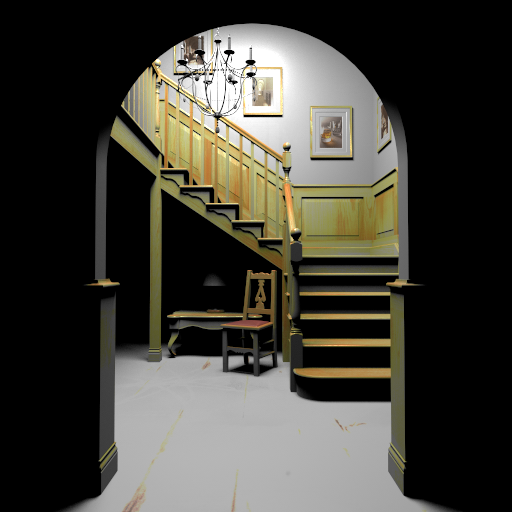}
 & 
\includegraphics[width=0.077\textwidth]{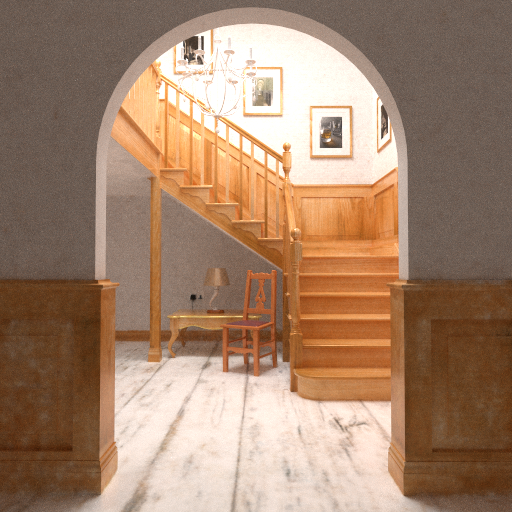}
 & 
\includegraphics[width=0.077\textwidth]{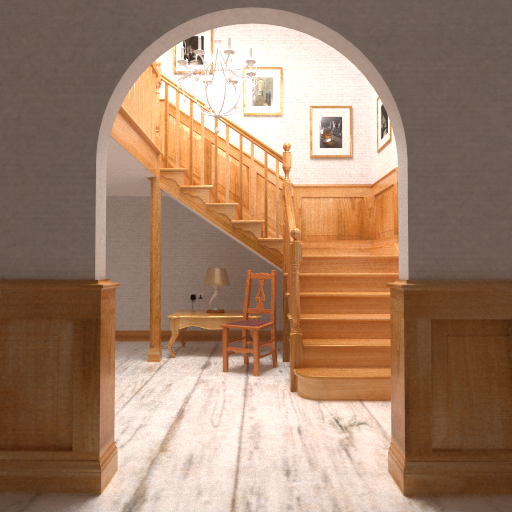}
 & 
\includegraphics[width=0.077\textwidth]{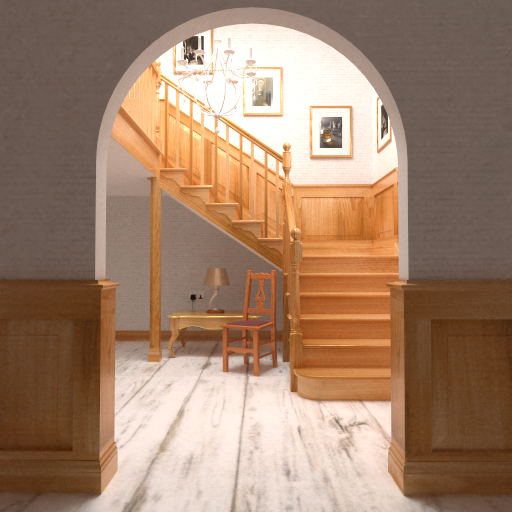}
 & 
\includegraphics[width=0.077\textwidth]{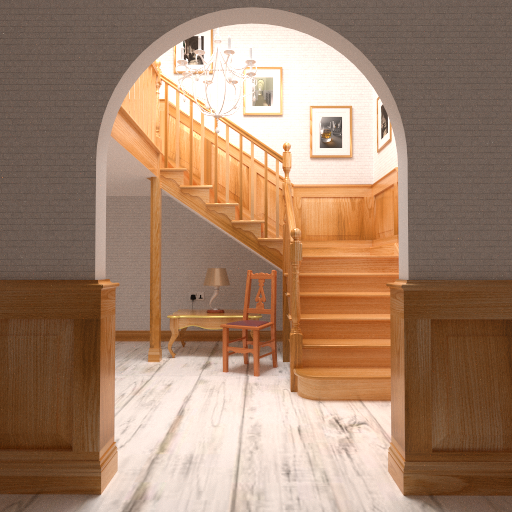}
 &  & 
\includegraphics[width=0.077\textwidth]{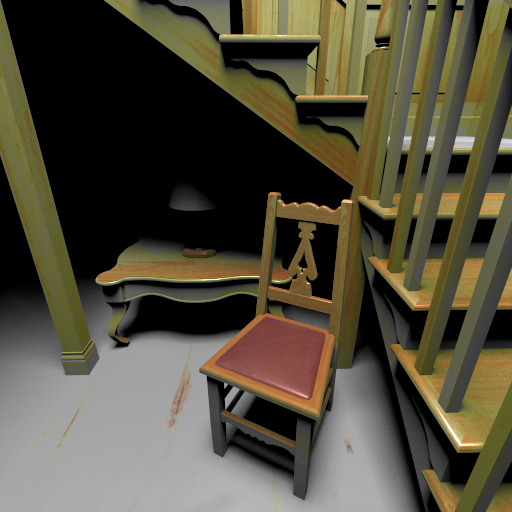}
 & 
\includegraphics[width=0.077\textwidth]{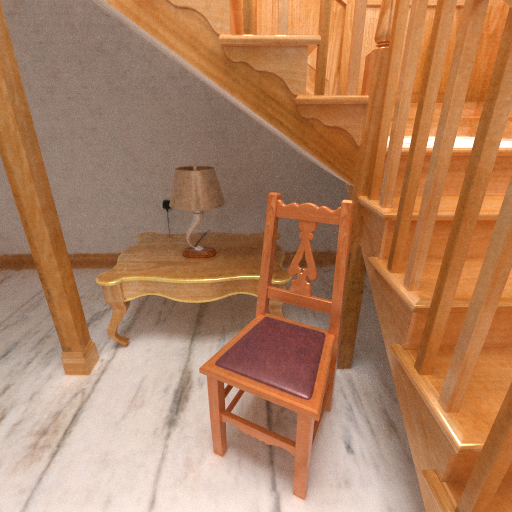}
 & 
\includegraphics[width=0.077\textwidth]{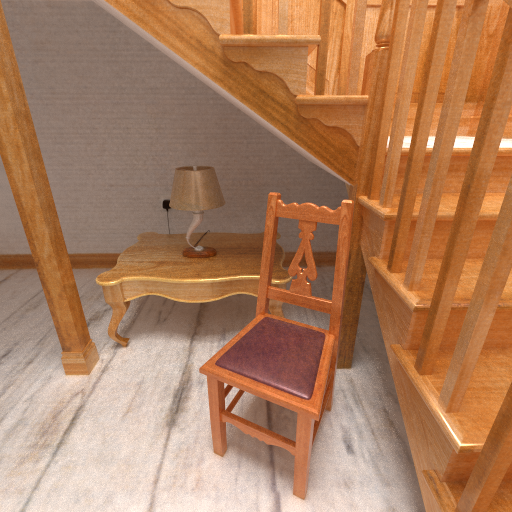}
 & 
\includegraphics[width=0.077\textwidth]{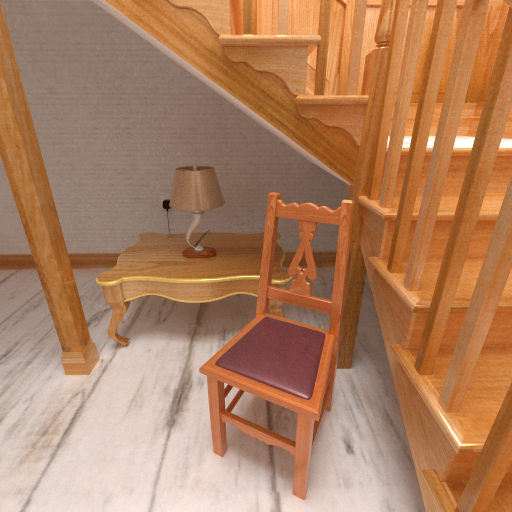}
 & 
\includegraphics[width=0.077\textwidth]{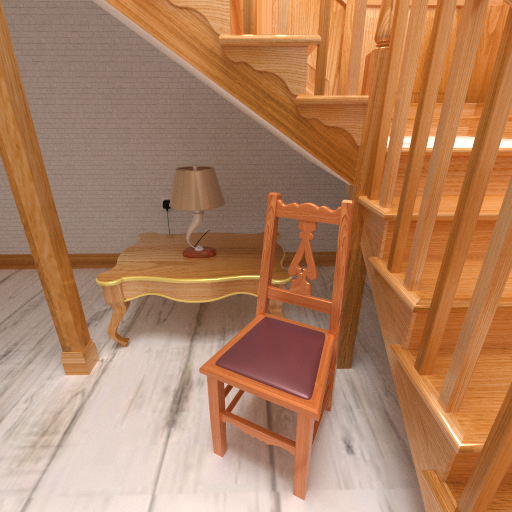}

\makebox[5pt]{\rotatebox{-90}{\hspace{-36pt}\footnotesize{Rendering}}}

 \\

{\footnotesize{11.72}}
 & 
{\footnotesize{30.34}}
 & 
{\footnotesize{29.04}}
 & 
{\footnotesize{\textbf{33.00}}}
 &  &  & 
{\footnotesize{13.19}}
 & 
{\footnotesize{31.28}}
 & 
{\footnotesize{29.55}}
 & 
{\footnotesize{\textbf{33.50}}}
 &  \\

\includegraphics[width=0.077\textwidth]{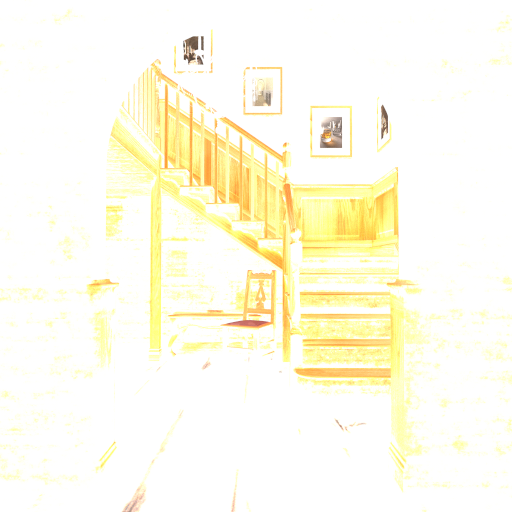}
 & 
\includegraphics[width=0.077\textwidth]{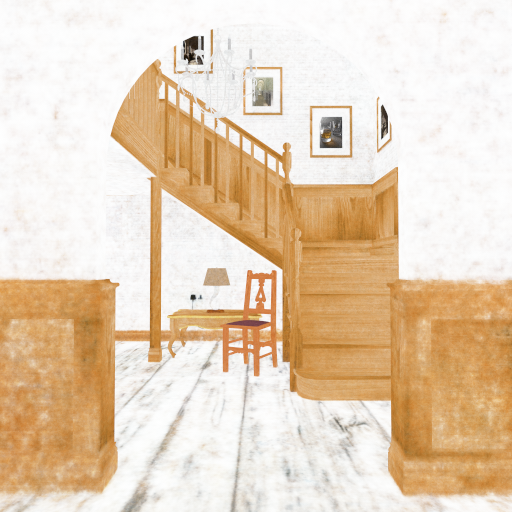}
 & 
\includegraphics[width=0.077\textwidth]{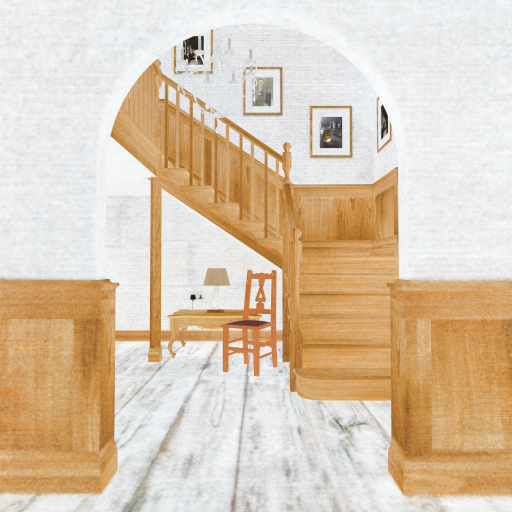}
 & 
\includegraphics[width=0.077\textwidth]{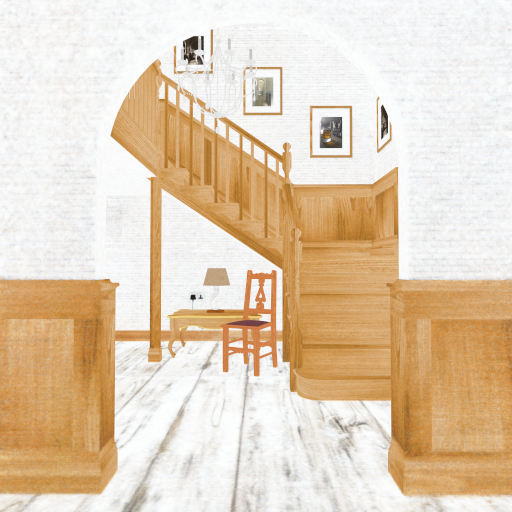}
 & 
\includegraphics[width=0.077\textwidth]{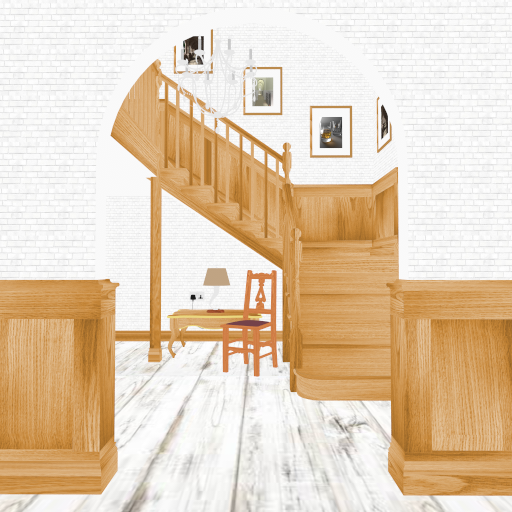}
 &  & 
\includegraphics[width=0.077\textwidth]{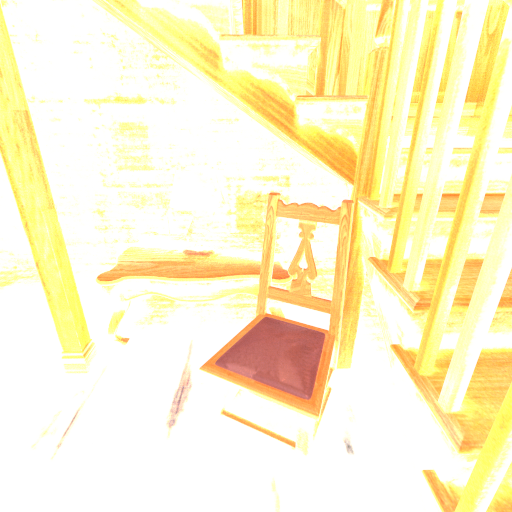}
 & 
\includegraphics[width=0.077\textwidth]{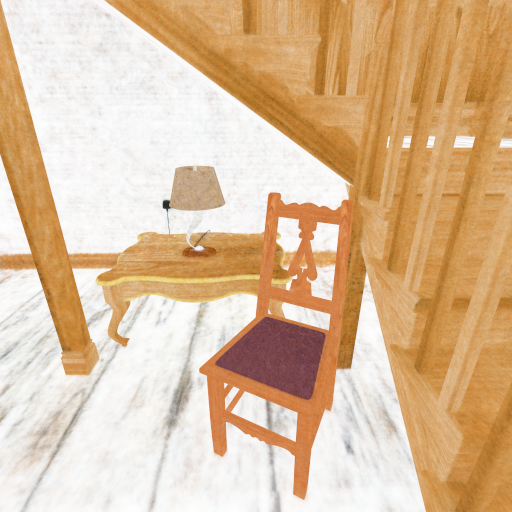}
 & 
\includegraphics[width=0.077\textwidth]{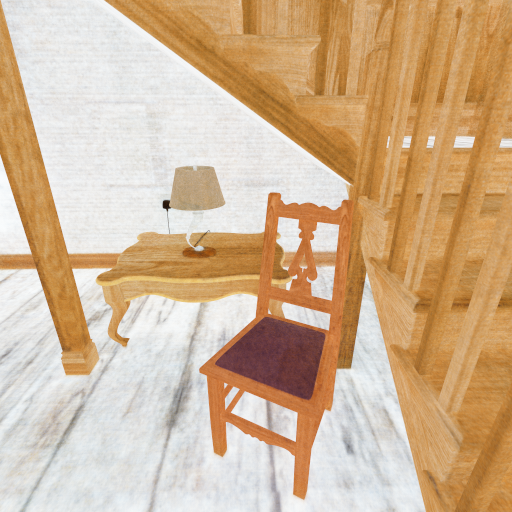}
 & 
\includegraphics[width=0.077\textwidth]{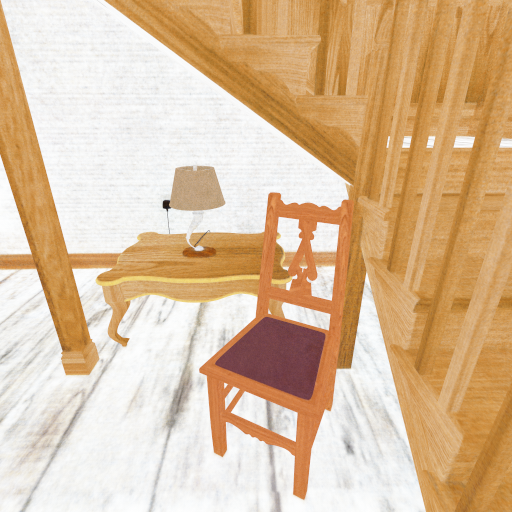}
 & 
\includegraphics[width=0.077\textwidth]{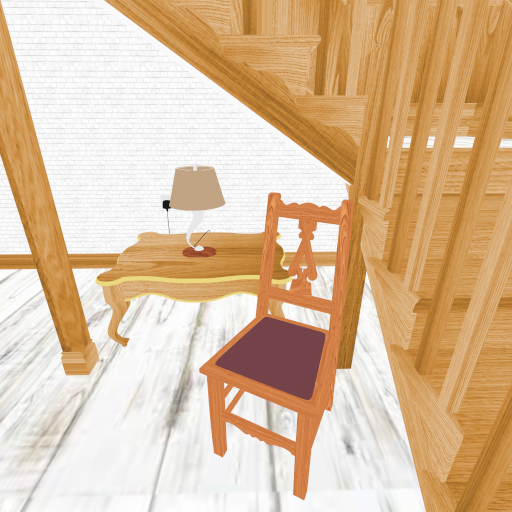}

\makebox[5pt]{\rotatebox{-90}{\hspace{-33pt}\footnotesize{Albedo}}}

 \\

{\footnotesize{8.65}}
 & 
{\footnotesize{24.93}}
 & 
{\footnotesize{20.89}}
 & 
{\footnotesize{\textbf{27.03}}}
 &  &  & 
{\footnotesize{8.39}}
 & 
{\footnotesize{27.26}}
 & 
{\footnotesize{20.99}}
 & 
{\footnotesize{\textbf{29.24}}}
 &  \\

\includegraphics[width=0.077\textwidth]{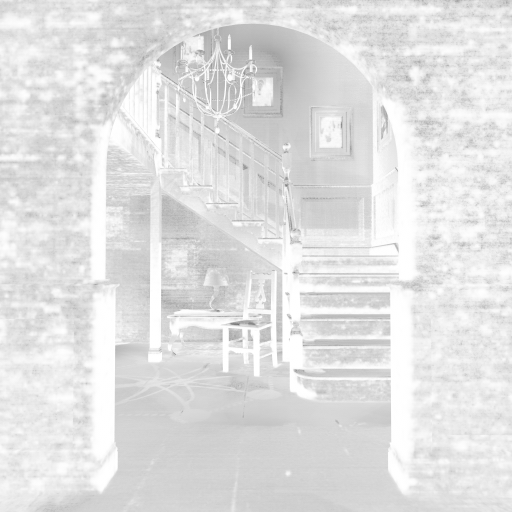}
 & 
\includegraphics[width=0.077\textwidth]{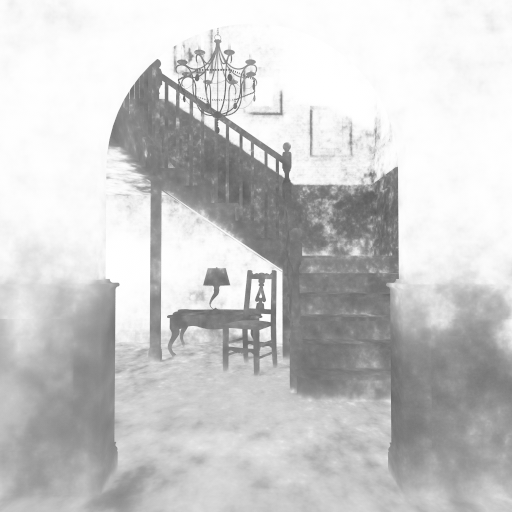}
 & 
\includegraphics[width=0.077\textwidth]{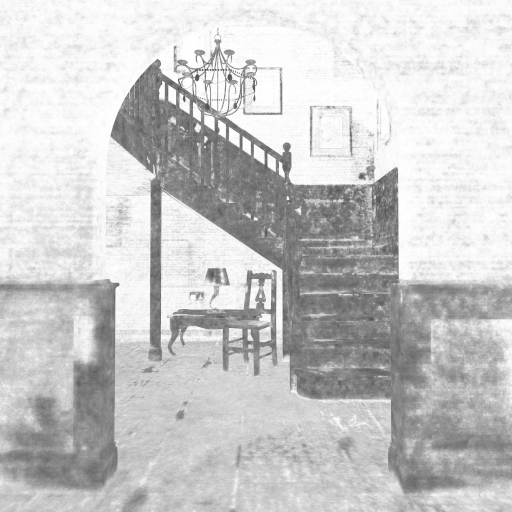}
 & 
\includegraphics[width=0.077\textwidth]{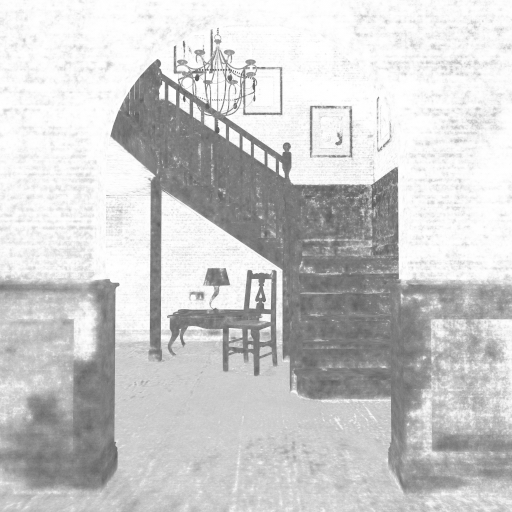}
 & 
\includegraphics[width=0.077\textwidth]{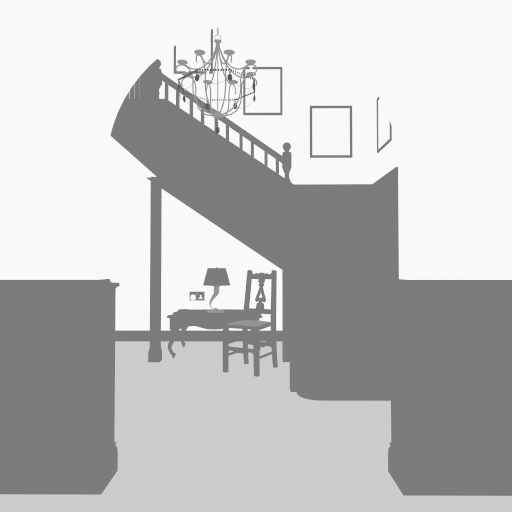}
 &  & 
\includegraphics[width=0.077\textwidth]{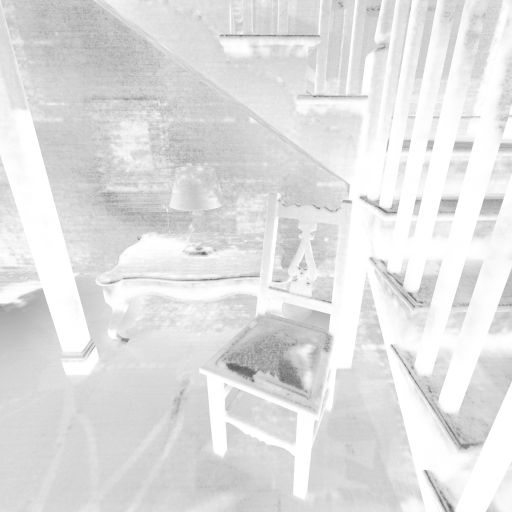}
 & 
\includegraphics[width=0.077\textwidth]{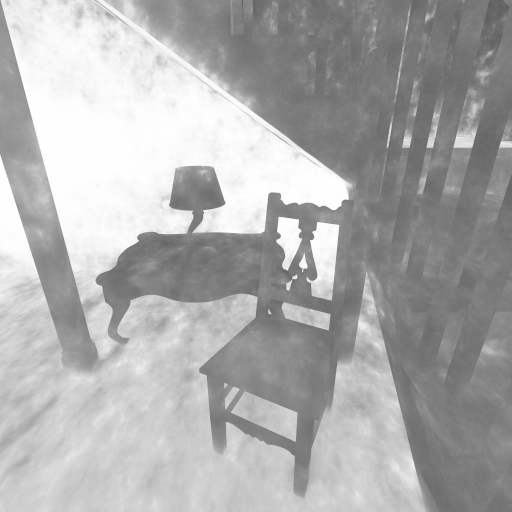}
 & 
\includegraphics[width=0.077\textwidth]{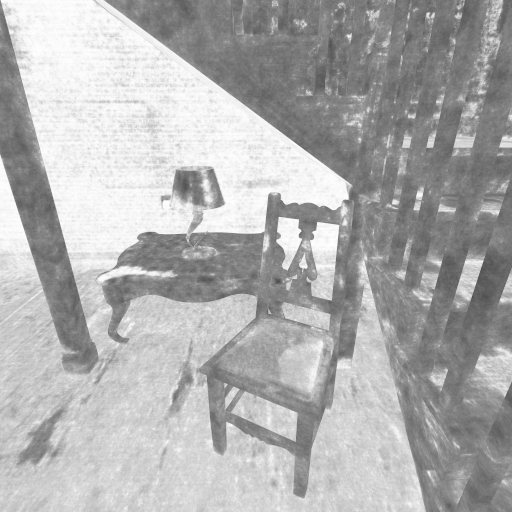}
 & 
\includegraphics[width=0.077\textwidth]{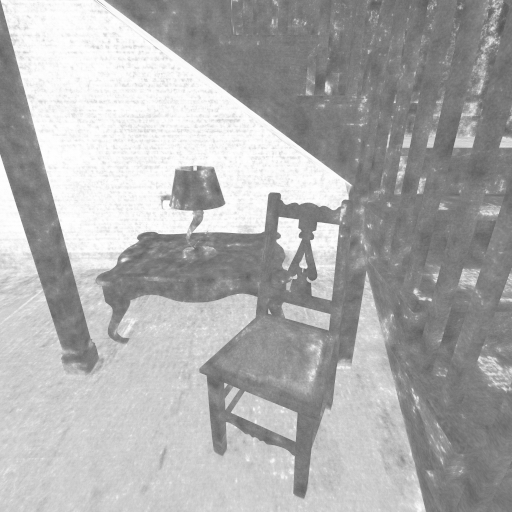}
 & 
\includegraphics[width=0.077\textwidth]{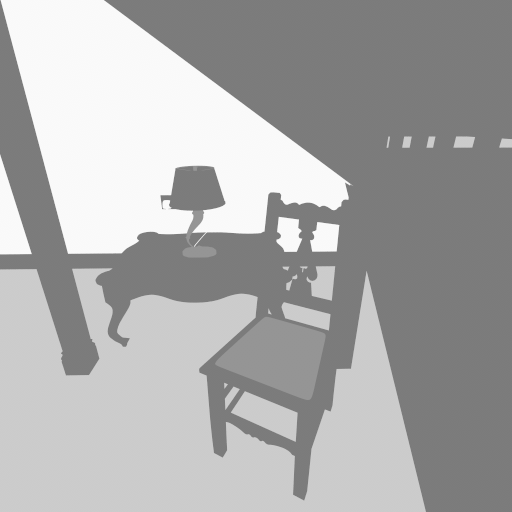}

\makebox[5pt]{\rotatebox{-90}{\hspace{-36pt}\footnotesize{Roughness}}}

 \\

{\footnotesize{7.54}}
 & 
{\footnotesize{\textbf{17.25}}}
 & 
{\footnotesize{16.37}}
 & 
{\footnotesize{16.38}}
 &  &  & 
{\footnotesize{6.00}}
 & 
{\footnotesize{20.80}}
 & 
{\footnotesize{18.49}}
 & 
{\footnotesize{\textbf{22.25}}}
 &  \\
\midrule
\end{tabular}
\end{subfigure}

\makebox[5pt]{\rotatebox{90}{\hspace{-10pt} \footnotesize{\LivingRoom}}}
\begin{subfigure}[b]{0.98\textwidth}
\begin{tabular}{ccccccccccc}

\includegraphics[width=0.077\textwidth]{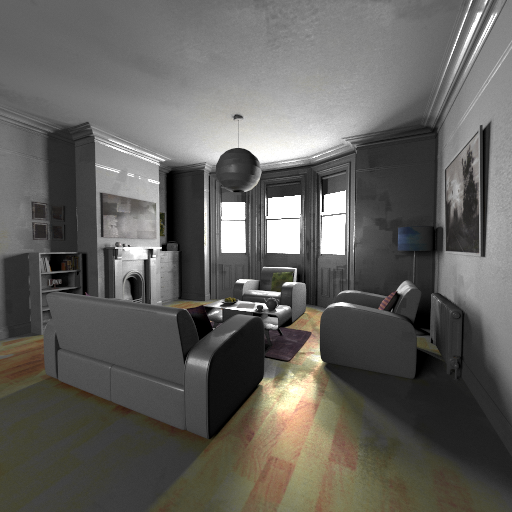}
 & 
\includegraphics[width=0.077\textwidth]{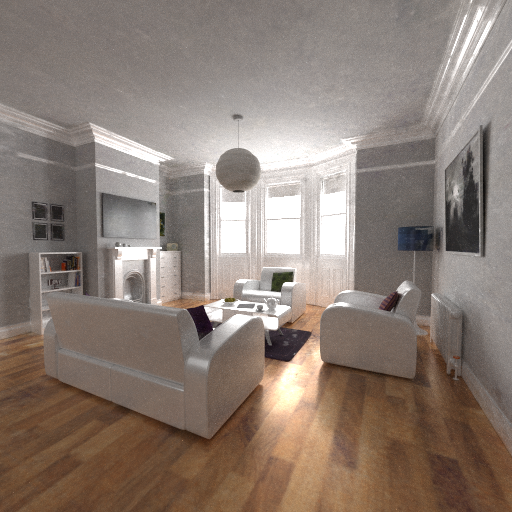}
 & 
\includegraphics[width=0.077\textwidth]{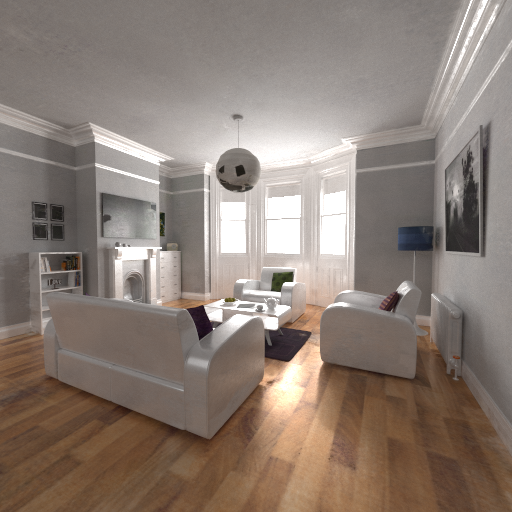}
 & 
\includegraphics[width=0.077\textwidth]{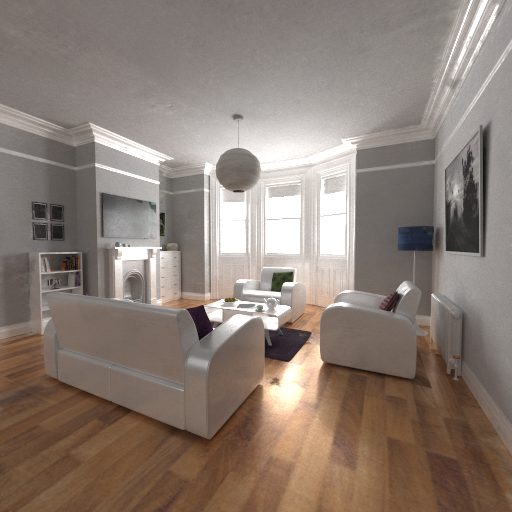}
 & 
\includegraphics[width=0.077\textwidth]{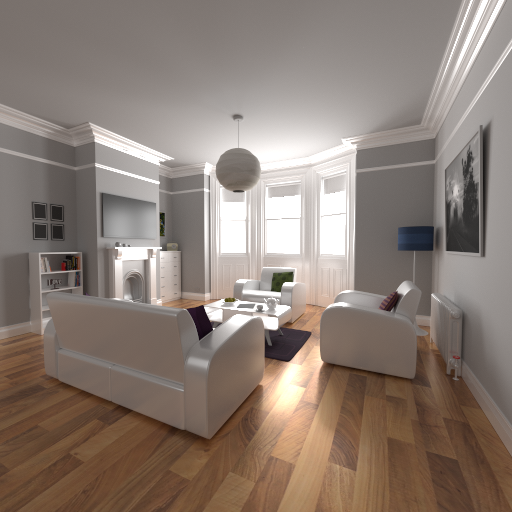}
 &  & 
\includegraphics[width=0.077\textwidth]{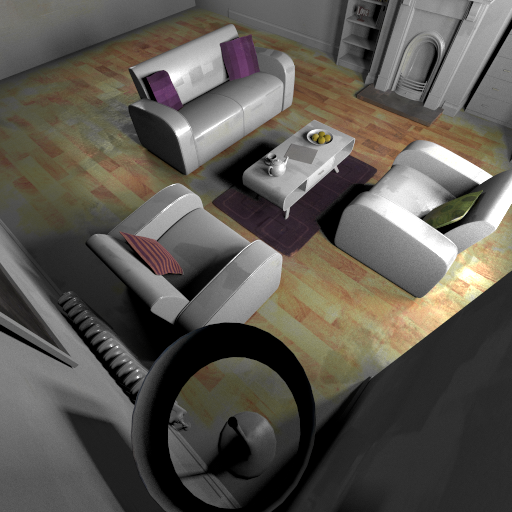}
 & 
\includegraphics[width=0.077\textwidth]{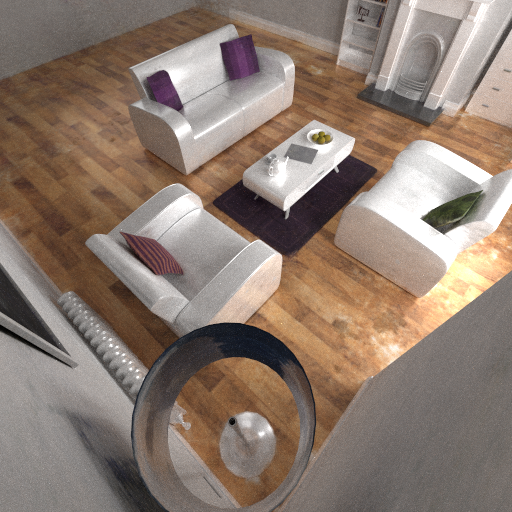}
 & 
\includegraphics[width=0.077\textwidth]{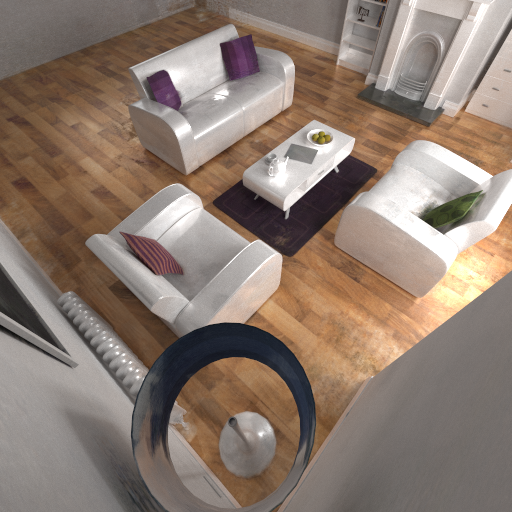}
 & 
\includegraphics[width=0.077\textwidth]{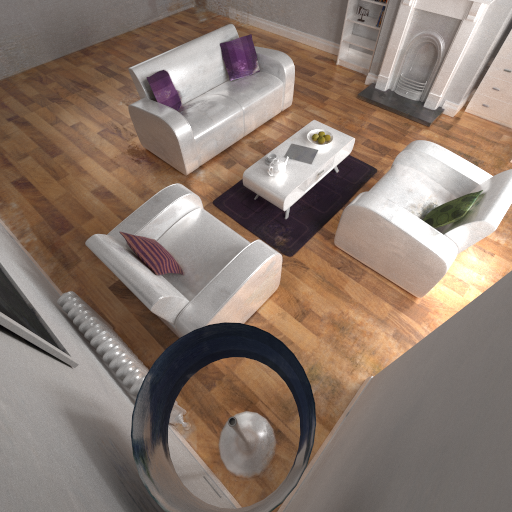}
 & 
\includegraphics[width=0.077\textwidth]{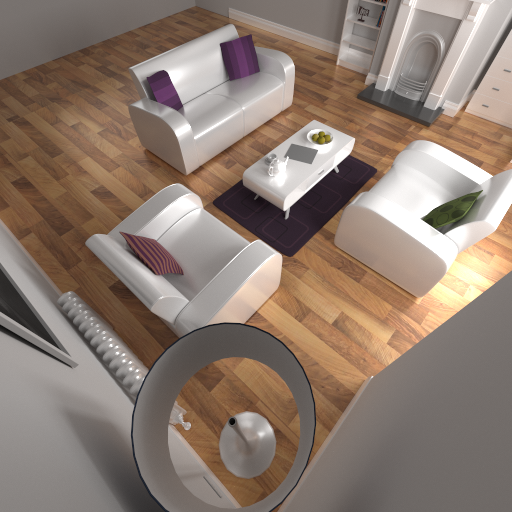}

\makebox[5pt]{\rotatebox{-90}{\hspace{-36pt}\footnotesize{Rendering}}}

 \\

{\footnotesize{12.41}}
 & 
{\footnotesize{\textbf{23.84}}}
 & 
{\footnotesize{20.95}}
 & 
{\footnotesize{23.23}}
 &  &  & 
{\footnotesize{14.63}}
 & 
{\footnotesize{\textbf{27.18}}}
 & 
{\footnotesize{25.00}}
 & 
{\footnotesize{26.05}}
 &  \\

\includegraphics[width=0.077\textwidth]{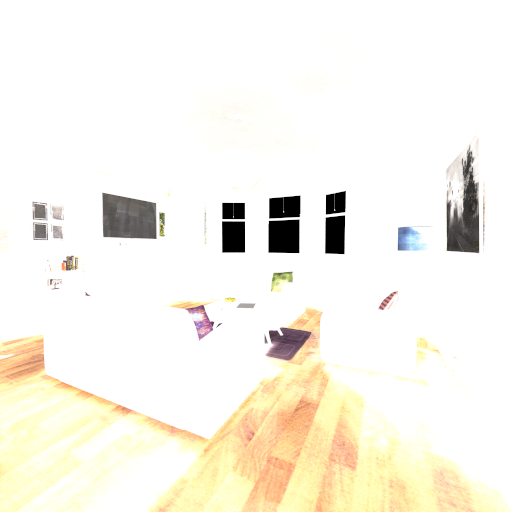}
 & 
\includegraphics[width=0.077\textwidth]{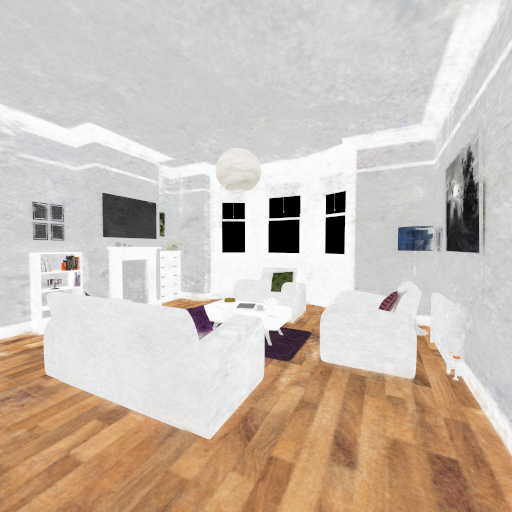}
 & 
\includegraphics[width=0.077\textwidth]{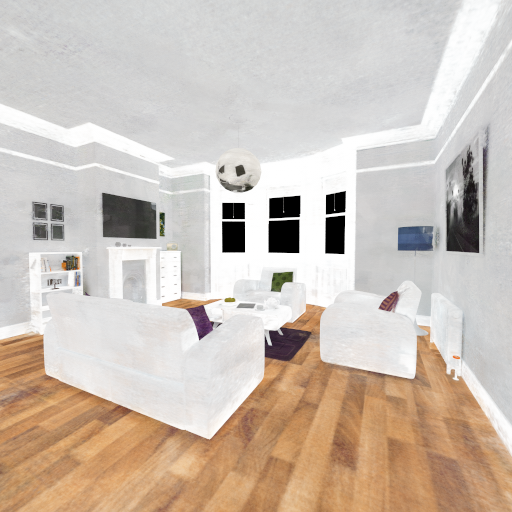}
 & 
\includegraphics[width=0.077\textwidth]{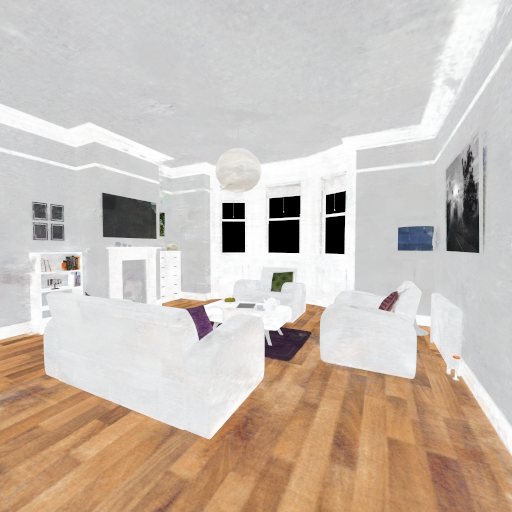}
 & 
\includegraphics[width=0.077\textwidth]{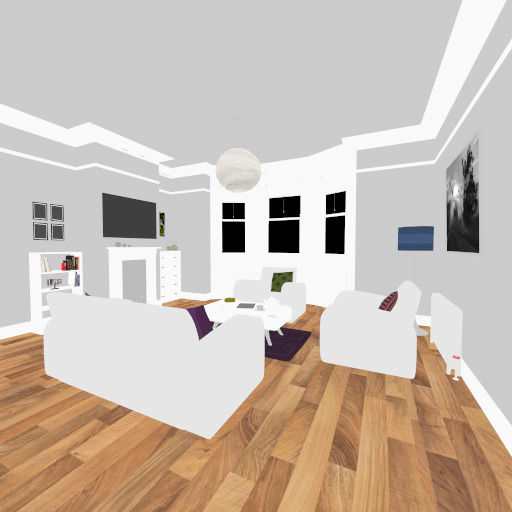}
 &  & 
\includegraphics[width=0.077\textwidth]{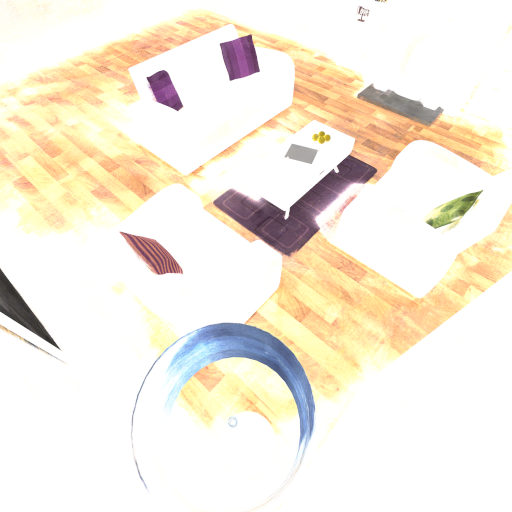}
 & 
\includegraphics[width=0.077\textwidth]{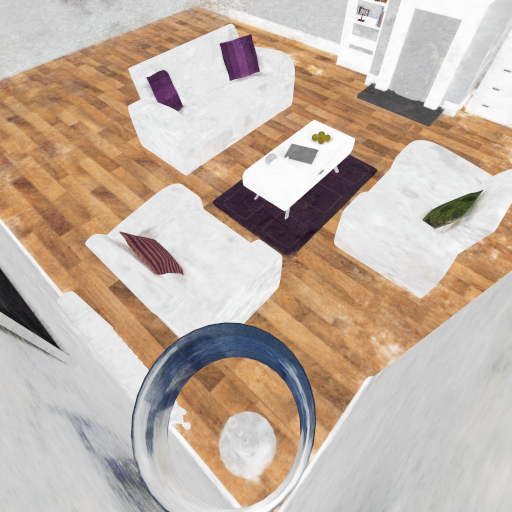}
 & 
\includegraphics[width=0.077\textwidth]{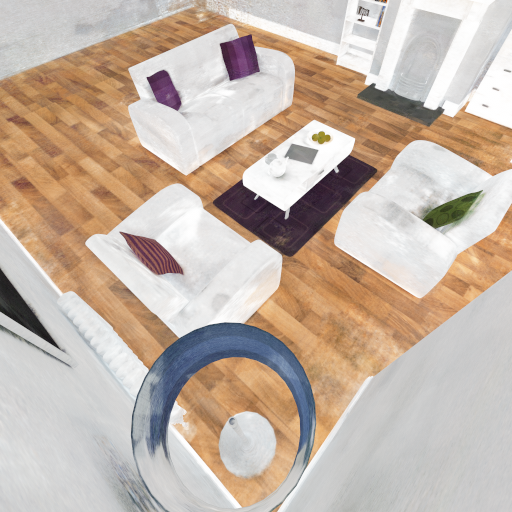}
 & 
\includegraphics[width=0.077\textwidth]{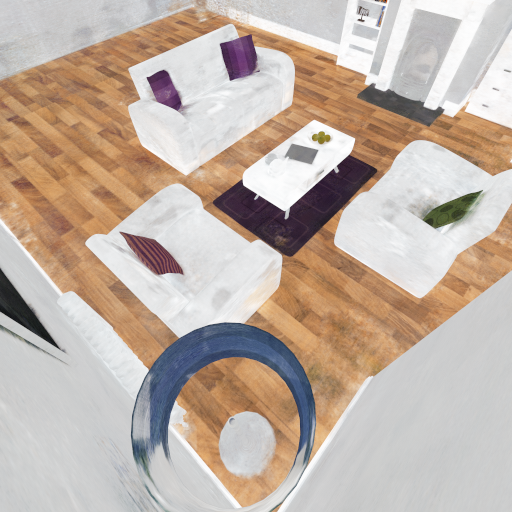}
 & 
\includegraphics[width=0.077\textwidth]{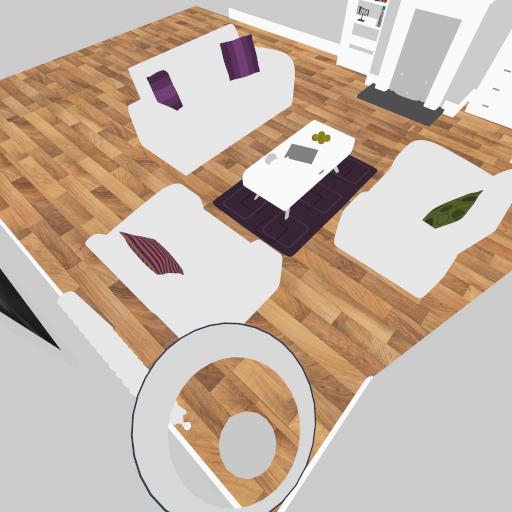}

\makebox[5pt]{\rotatebox{-90}{\hspace{-33pt}\footnotesize{Albedo}}}

 \\

{\footnotesize{7.73}}
 & 
{\footnotesize{26.56}}
 & 
{\footnotesize{23.64}}
 & 
{\footnotesize{\textbf{27.16}}}
 &  &  & 
{\footnotesize{7.76}}
 & 
{\footnotesize{\textbf{18.89}}}
 & 
{\footnotesize{18.43}}
 & 
{\footnotesize{18.72}}
 &  \\

\includegraphics[width=0.077\textwidth]{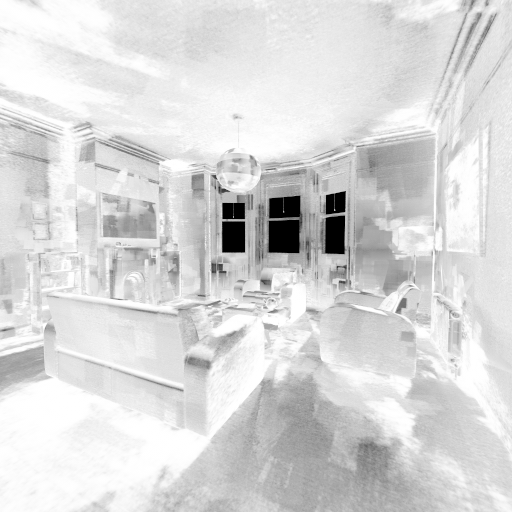}
 & 
\includegraphics[width=0.077\textwidth]{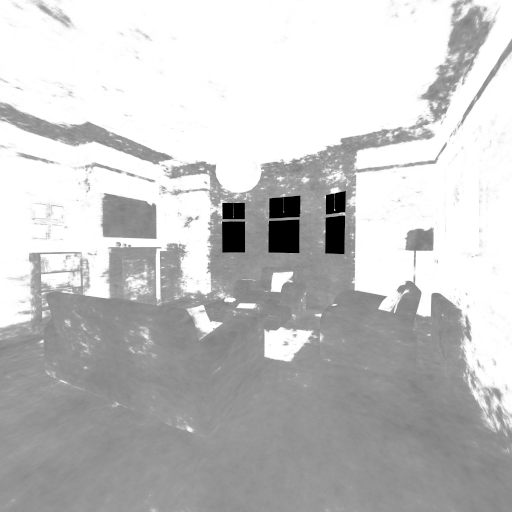}
 & 
\includegraphics[width=0.077\textwidth]{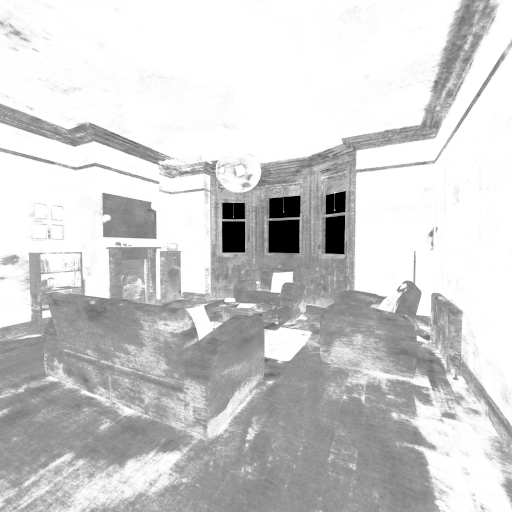}
 & 
\includegraphics[width=0.077\textwidth]{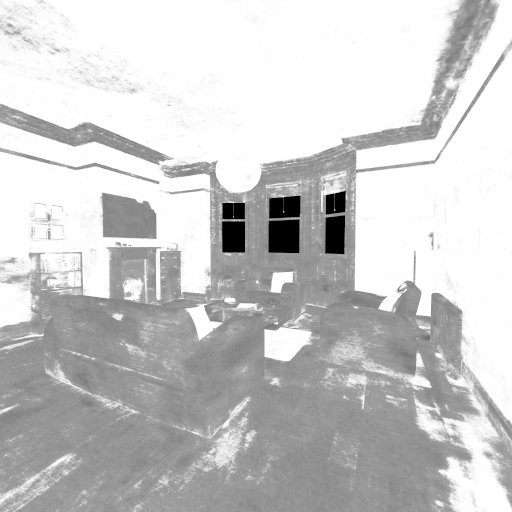}
 & 
\includegraphics[width=0.077\textwidth]{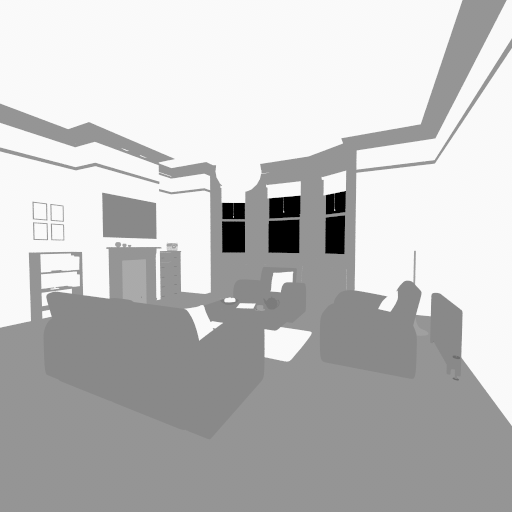}
 &  & 
\includegraphics[width=0.077\textwidth]{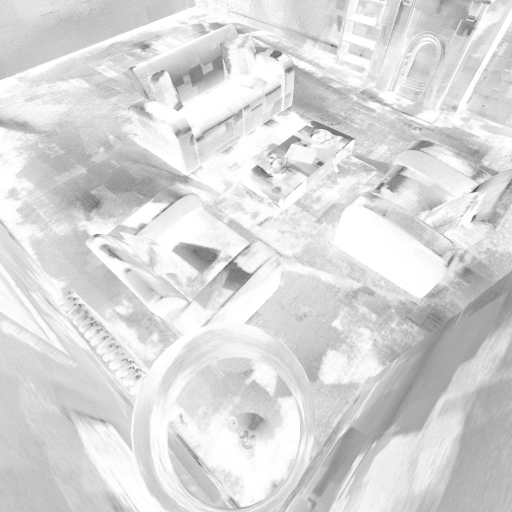}
 & 
\includegraphics[width=0.077\textwidth]{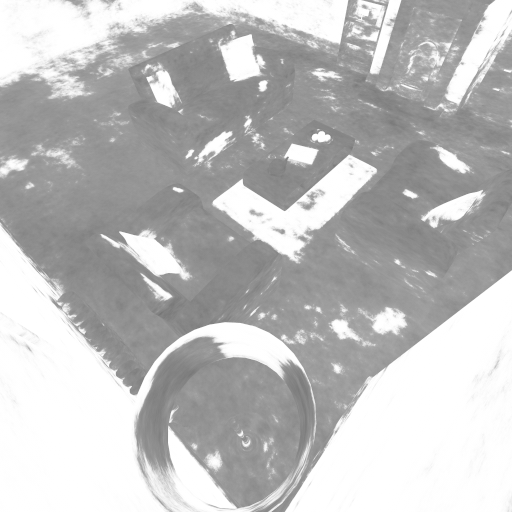}
 & 
\includegraphics[width=0.077\textwidth]{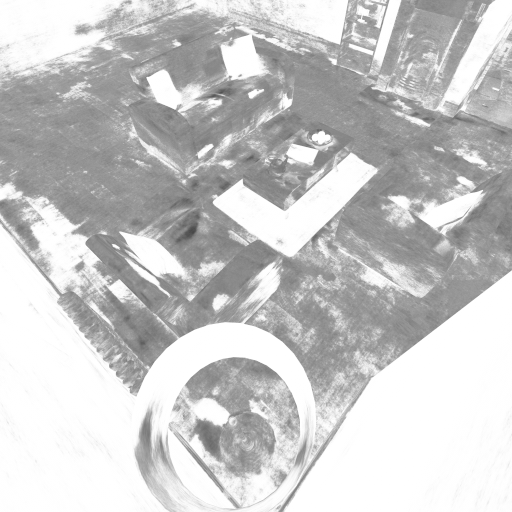}
 & 
\includegraphics[width=0.077\textwidth]{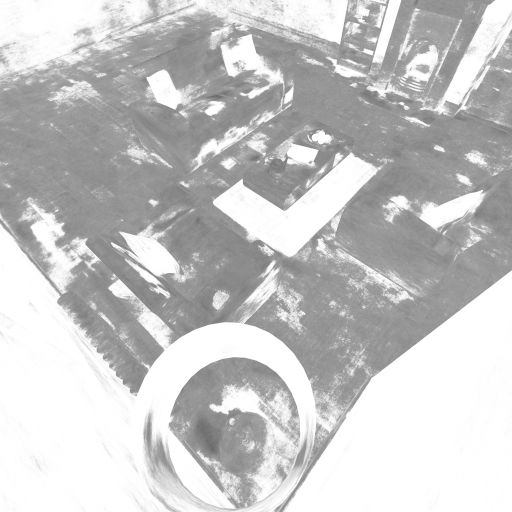}
 & 
\includegraphics[width=0.077\textwidth]{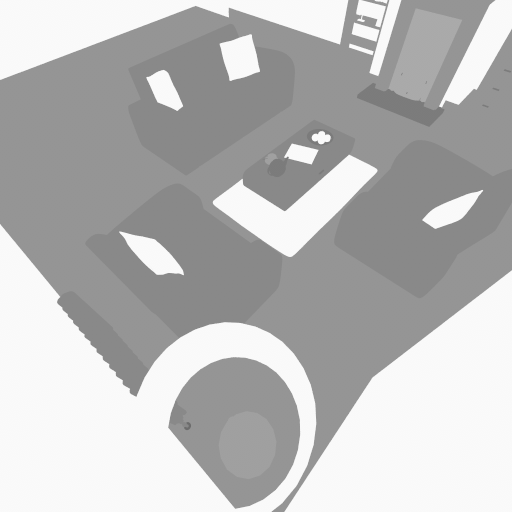}

\makebox[5pt]{\rotatebox{-90}{\hspace{-36pt}\footnotesize{Roughness}}}

 \\

{\footnotesize{8.88}}
 & 
{\footnotesize{\textbf{19.69}}}
 & 
{\footnotesize{14.83}}
 & 
{\footnotesize{17.32}}
 &  &  & 
{\footnotesize{7.31}}
 & 
{\footnotesize{\textbf{15.92}}}
 & 
{\footnotesize{14.32}}
 & 
{\footnotesize{15.13}}
 &  \\
\midrule
\end{tabular}
\end{subfigure}

\makebox[5pt]{\rotatebox{90}{\hspace{-10pt} \footnotesize{\Kitchen}}}
\begin{subfigure}[b]{0.98\textwidth}
\begin{tabular}{ccccccccccc}

\includegraphics[width=0.077\textwidth]{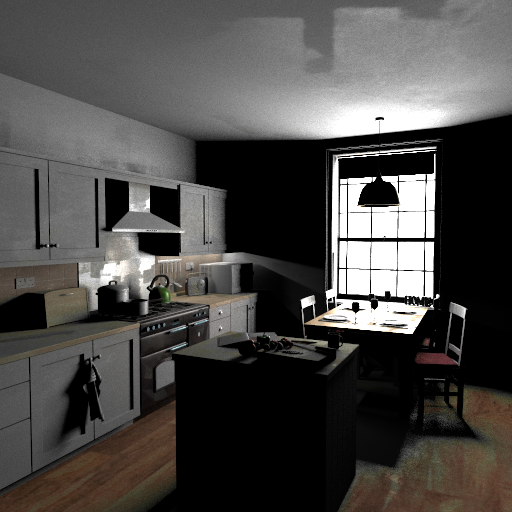}
 & 
\includegraphics[width=0.077\textwidth]{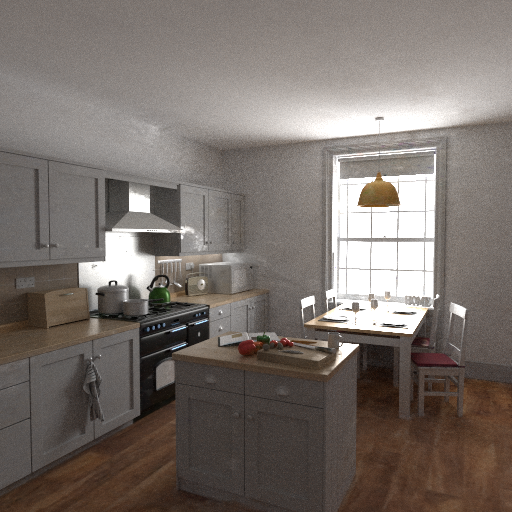}
 & 
\includegraphics[width=0.077\textwidth]{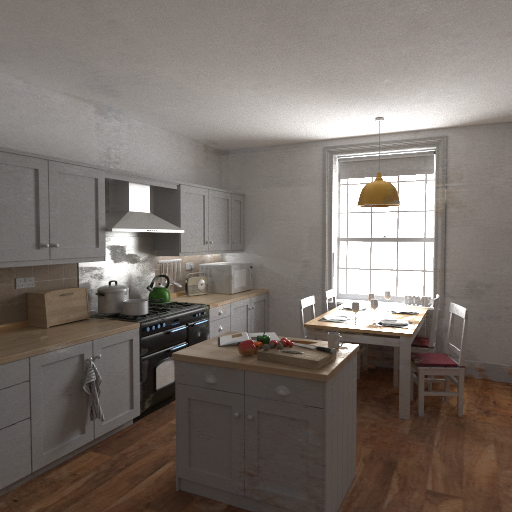}
 & 
\includegraphics[width=0.077\textwidth]{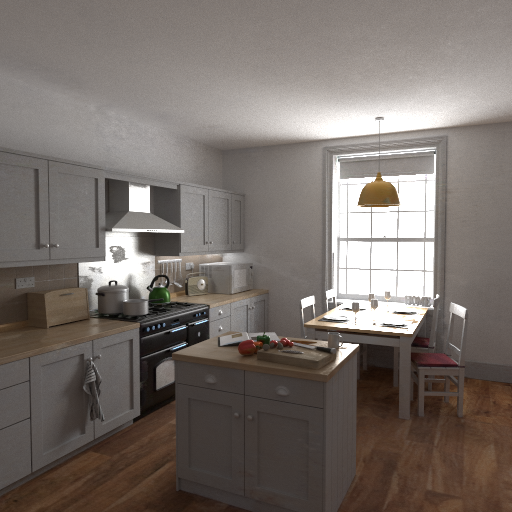}
 & 
\includegraphics[width=0.077\textwidth]{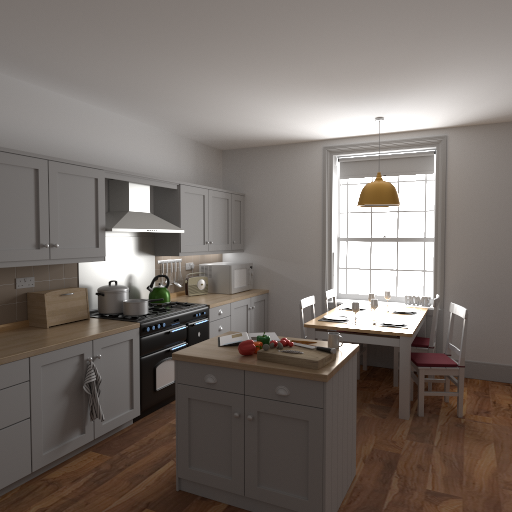}
 &  & 
\includegraphics[width=0.077\textwidth]{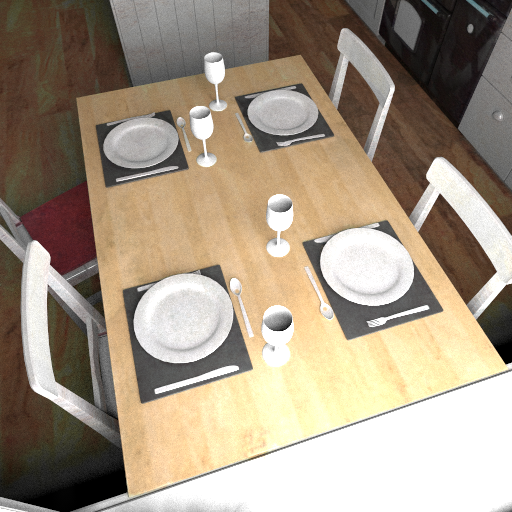}
 & 
\includegraphics[width=0.077\textwidth]{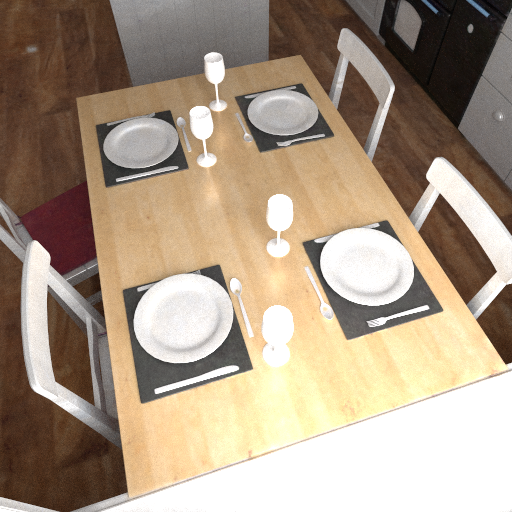}
 & 
\includegraphics[width=0.077\textwidth]{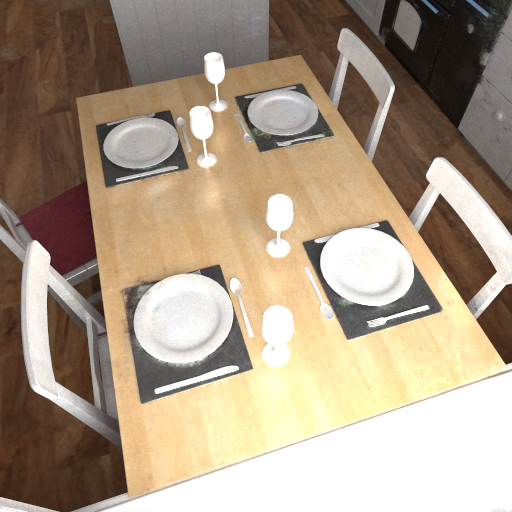}
 & 
\includegraphics[width=0.077\textwidth]{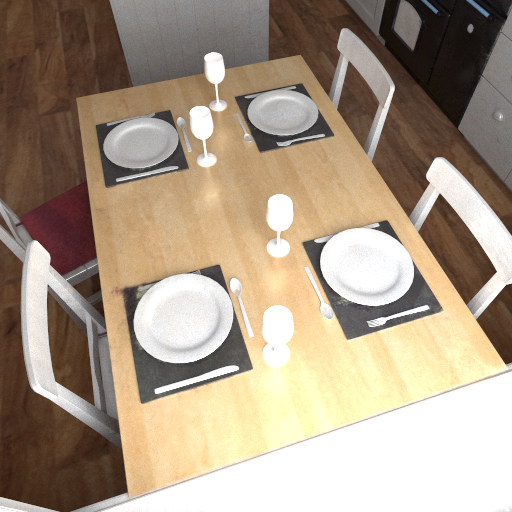}
 & 
\includegraphics[width=0.077\textwidth]{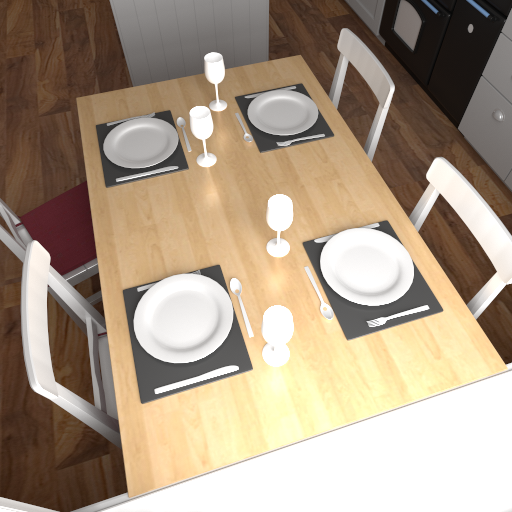}

\makebox[5pt]{\rotatebox{-90}{\hspace{-36pt}\footnotesize{Rendering}}}

 \\

{\footnotesize{14.26}}
 & 
{\footnotesize{\textbf{26.23}}}
 & 
{\footnotesize{16.38}}
 & 
{\footnotesize{21.64}}
 &  &  & 
{\footnotesize{15.49}}
 & 
{\footnotesize{\textbf{23.00}}}
 & 
{\footnotesize{20.68}}
 & 
{\footnotesize{21.64}}
 &  \\

\includegraphics[width=0.077\textwidth]{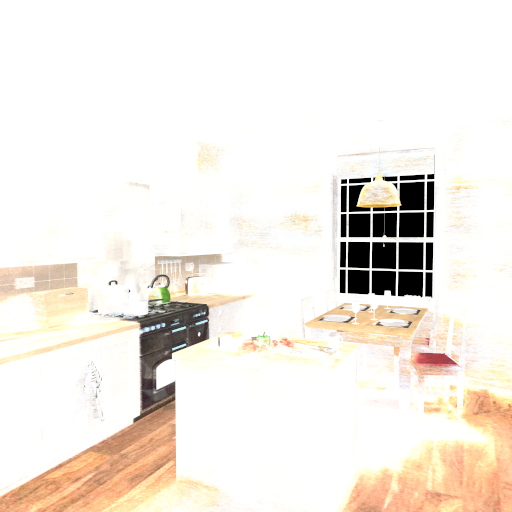}
 & 
\includegraphics[width=0.077\textwidth]{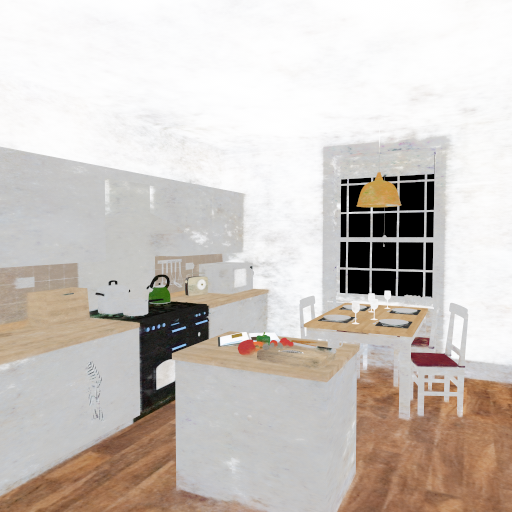}
 & 
\includegraphics[width=0.077\textwidth]{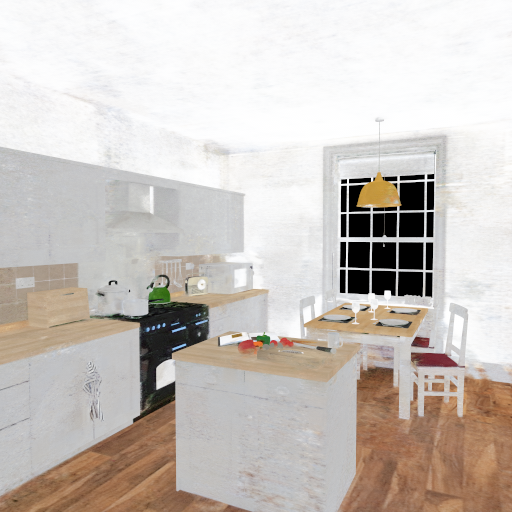}
 & 
\includegraphics[width=0.077\textwidth]{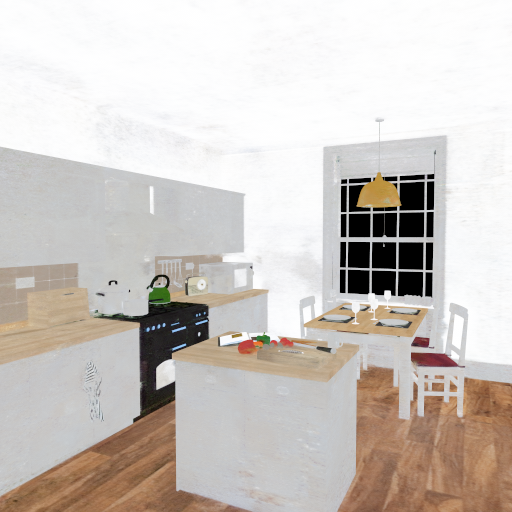}
 & 
\includegraphics[width=0.077\textwidth]{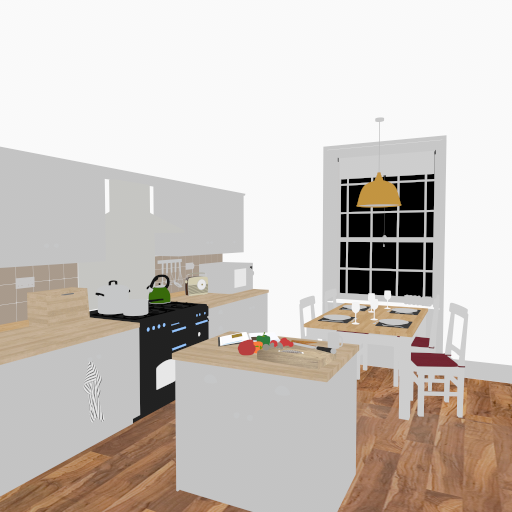}
 &  & 
\includegraphics[width=0.077\textwidth]{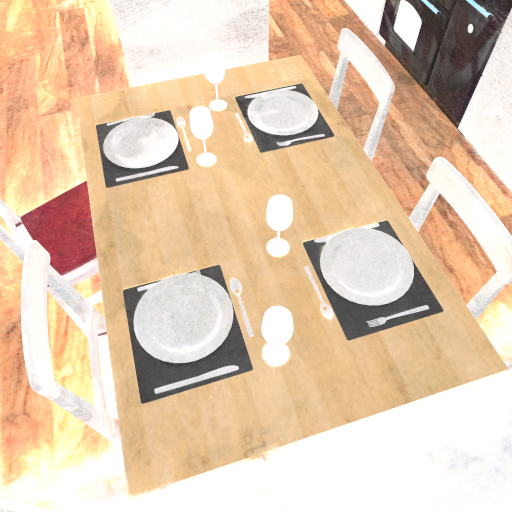}
 & 
\includegraphics[width=0.077\textwidth]{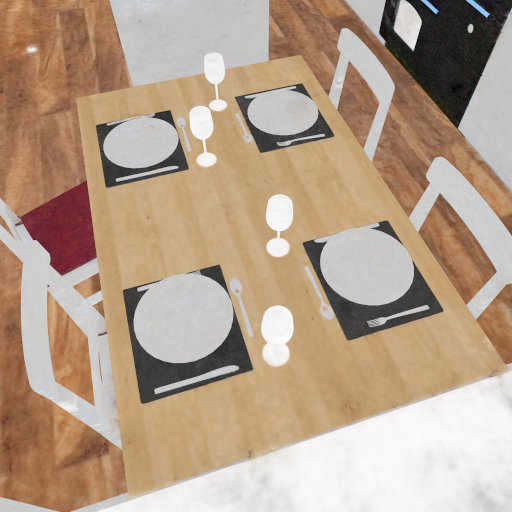}
 & 
\includegraphics[width=0.077\textwidth]{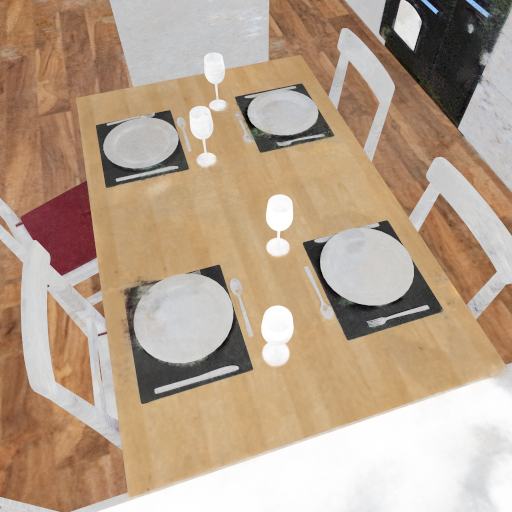}
 & 
\includegraphics[width=0.077\textwidth]{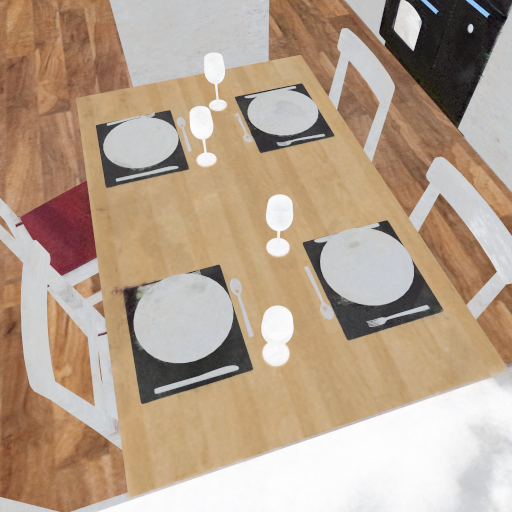}
 & 
\includegraphics[width=0.077\textwidth]{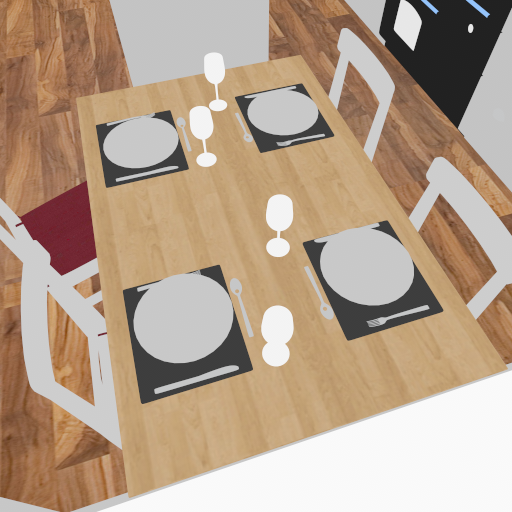}

\makebox[5pt]{\rotatebox{-90}{\hspace{-33pt}\footnotesize{Albedo}}}

 \\

{\footnotesize{9.32}}
 & 
{\footnotesize{28.03}}
 & 
{\footnotesize{24.58}}
 & 
{\footnotesize{\textbf{29.05}}}
 &  &  & 
{\footnotesize{10.43}}
 & 
{\footnotesize{\textbf{27.85}}}
 & 
{\footnotesize{27.02}}
 & 
{\footnotesize{27.70}}
 &  \\

\includegraphics[width=0.077\textwidth]{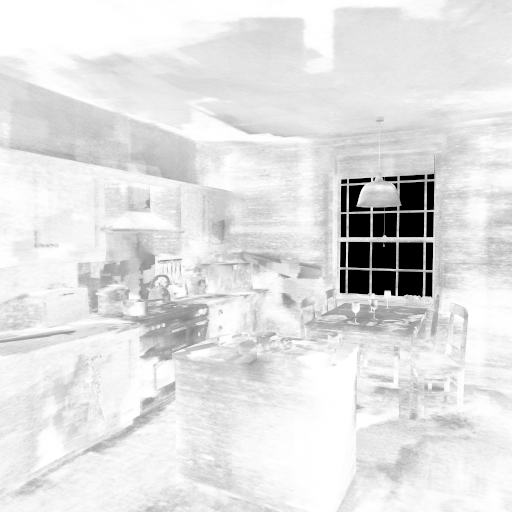}
 & 
\includegraphics[width=0.077\textwidth]{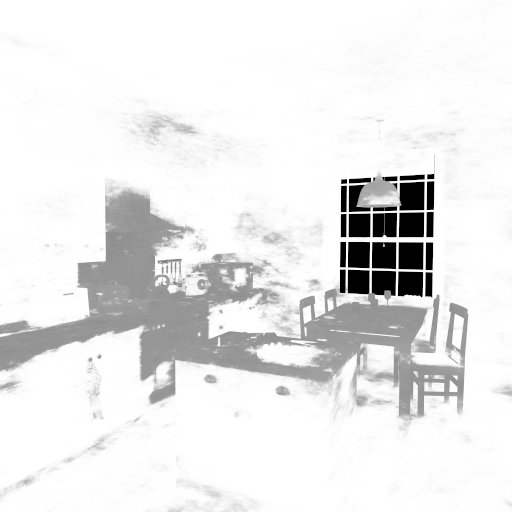}
 & 
\includegraphics[width=0.077\textwidth]{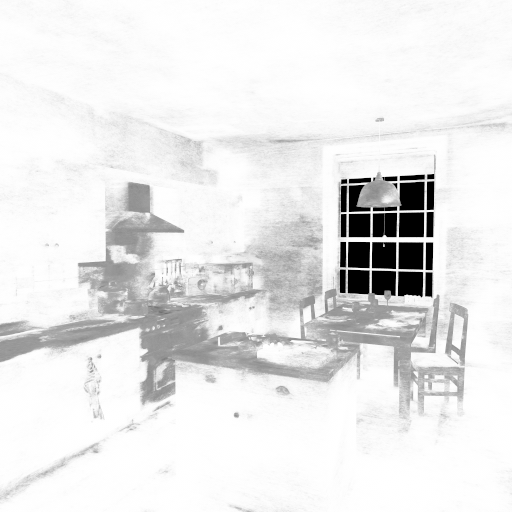}
 & 
\includegraphics[width=0.077\textwidth]{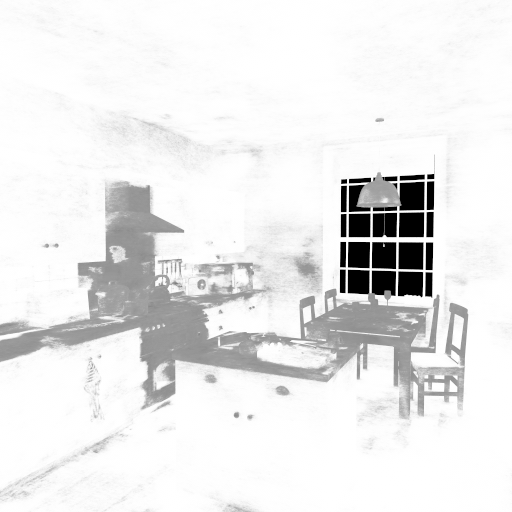}
 & 
\includegraphics[width=0.077\textwidth]{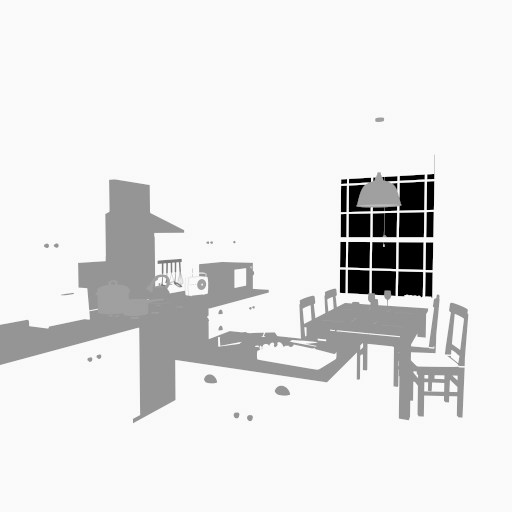}
 &  & 
\includegraphics[width=0.077\textwidth]{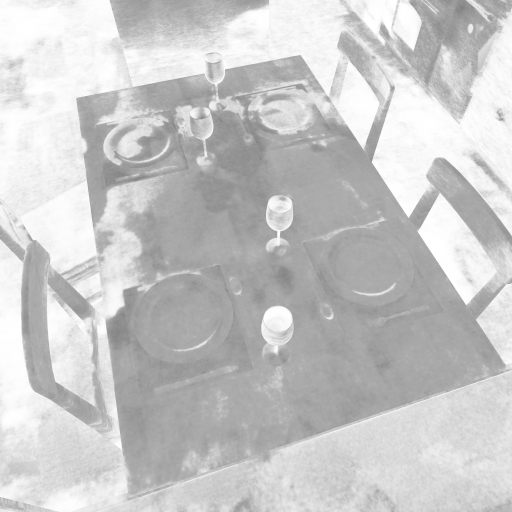}
 & 
\includegraphics[width=0.077\textwidth]{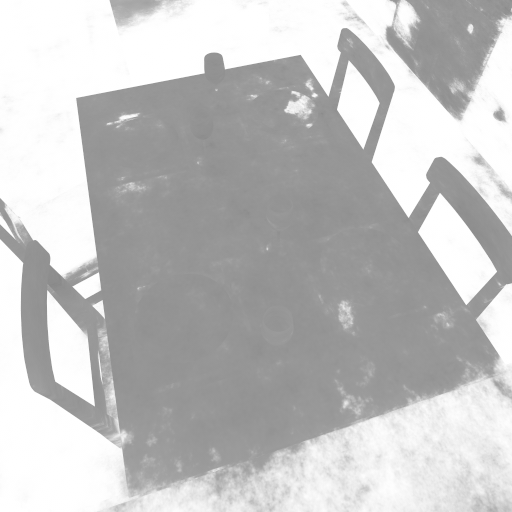}
 & 
\includegraphics[width=0.077\textwidth]{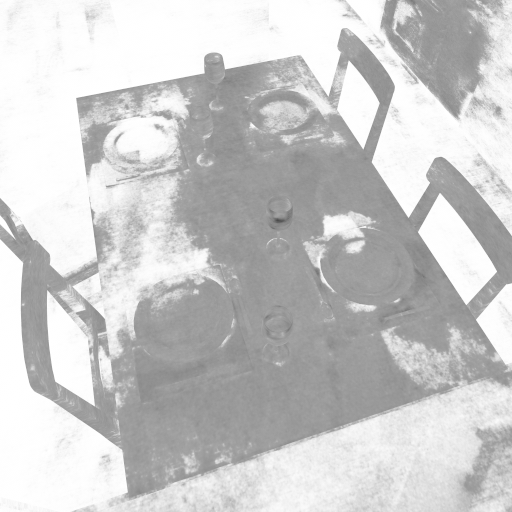}
 & 
\includegraphics[width=0.077\textwidth]{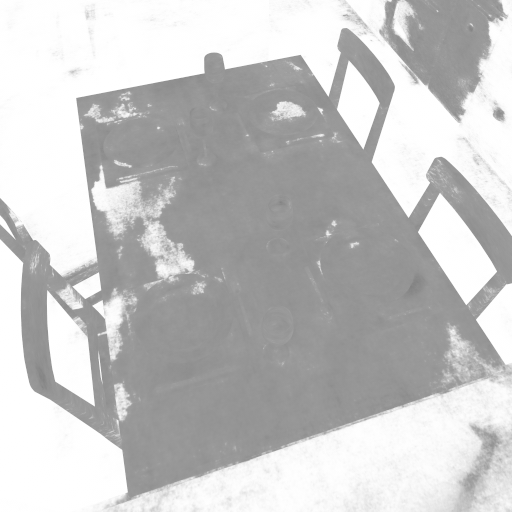}
 & 
\includegraphics[width=0.077\textwidth]{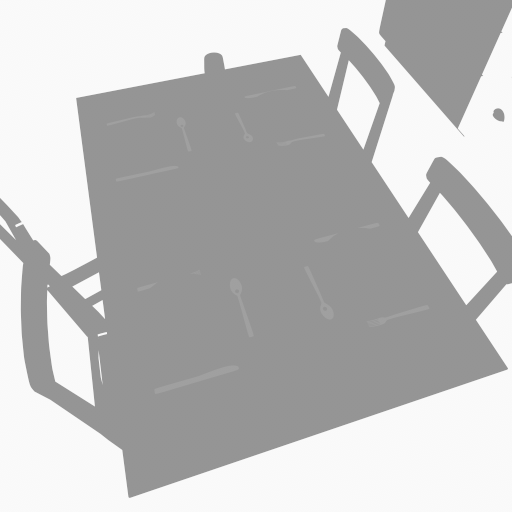}

\makebox[5pt]{\rotatebox{-90}{\hspace{-36pt}\footnotesize{Roughness}}}

 \\

{\footnotesize{13.29}}
 & 
{\footnotesize{20.56}}
 & 
{\footnotesize{18.37}}
 & 
{\footnotesize{\textbf{21.47}}}
 &  &  & 
{\footnotesize{13.85}}
 & 
{\footnotesize{\textbf{19.76}}}
 & 
{\footnotesize{14.70}}
 & 
{\footnotesize{19.05}}
 &  \\
\midrule
\end{tabular}
\end{subfigure}

\makebox[5pt]{\rotatebox{90}{\hspace{-10pt} \footnotesize{\Lego}}}
\begin{subfigure}[b]{0.98\textwidth}
\begin{tabular}{ccccccccccc}

\includegraphics[trim={50 50 50 50},clip,width=0.077\textwidth]{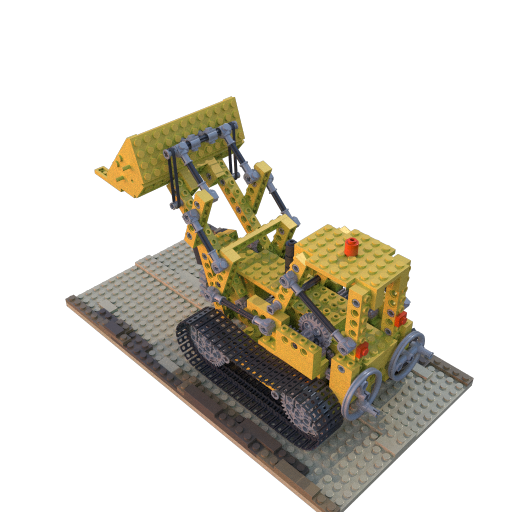}
 & 
\includegraphics[trim={50 50 50 50},clip,width=0.077\textwidth]{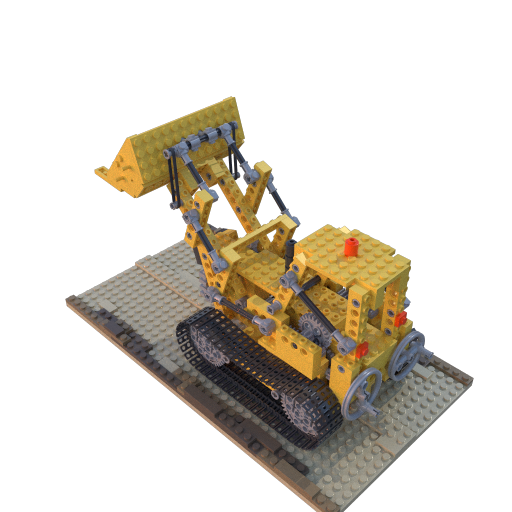}
 & 
\includegraphics[trim={50 50 50 50},clip,width=0.077\textwidth]{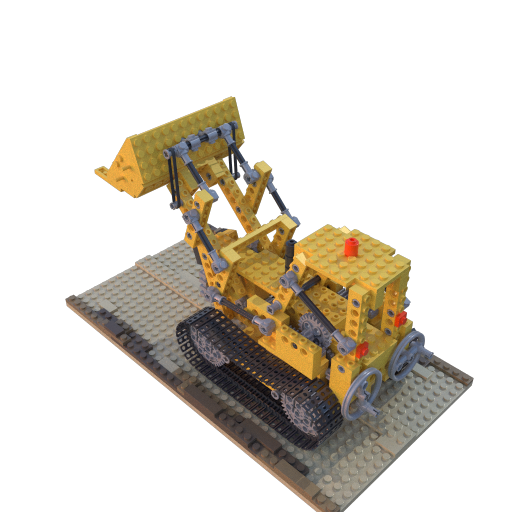}
 & 
\includegraphics[trim={50 50 50 50},clip,width=0.077\textwidth]{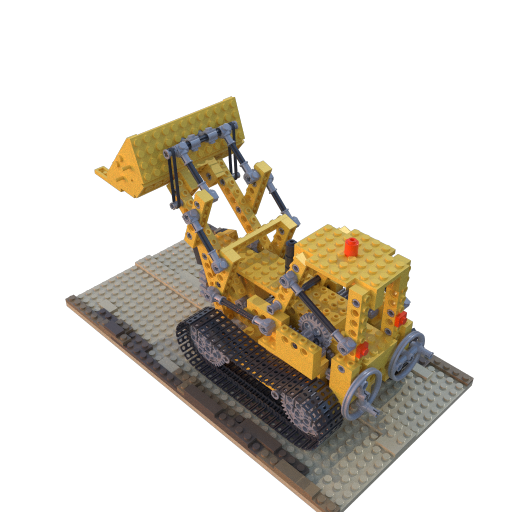}
 & 
\includegraphics[trim={50 50 50 50},clip,width=0.077\textwidth]{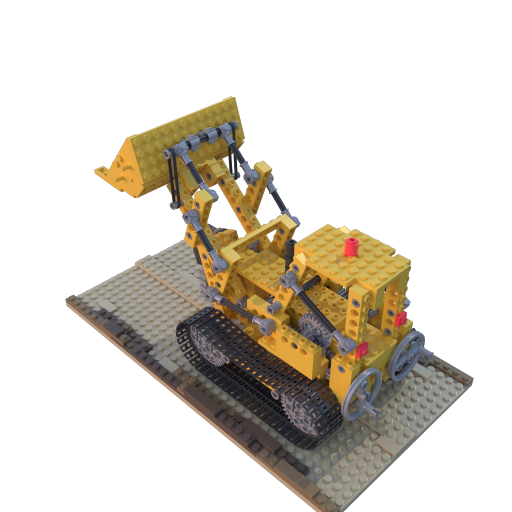}
 &  & 
\includegraphics[trim={50 50 50 50},clip,width=0.077\textwidth]{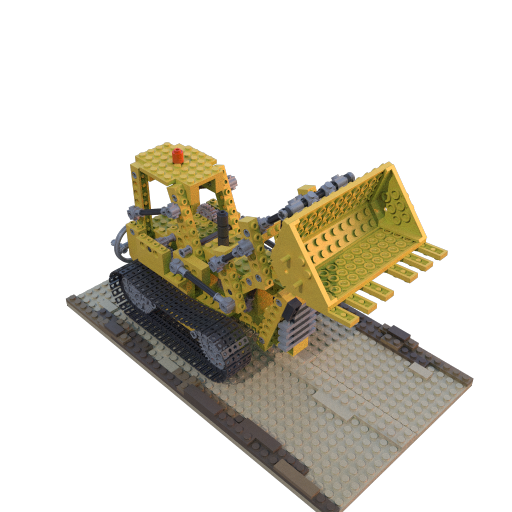}
 & 
\includegraphics[trim={50 50 50 50},clip,width=0.077\textwidth]{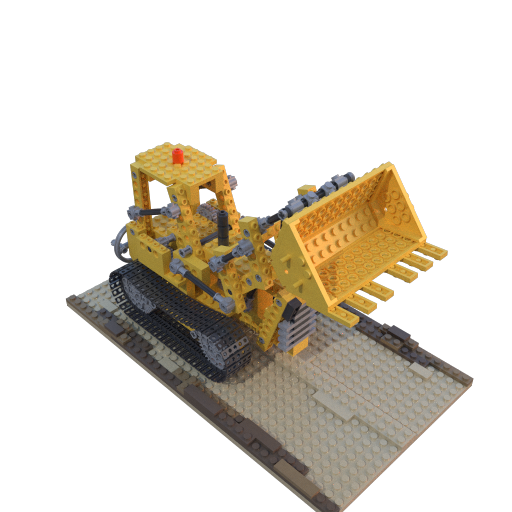}
 & 
\includegraphics[trim={50 50 50 50},clip,width=0.077\textwidth]{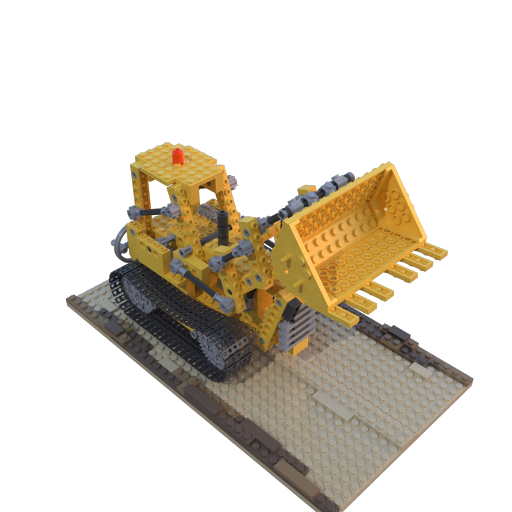}
 & 
\includegraphics[trim={50 50 50 50},clip,width=0.077\textwidth]{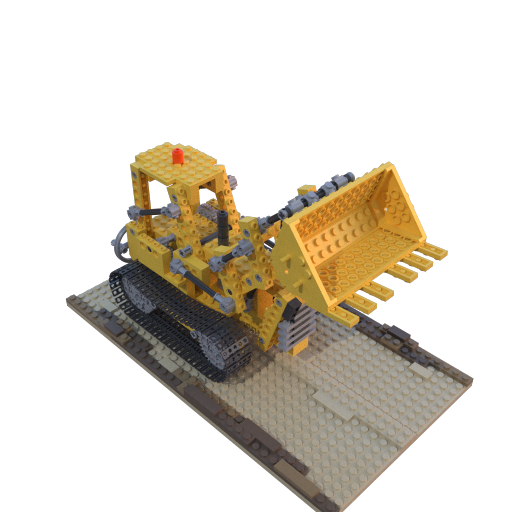}
 & 
\includegraphics[trim={50 50 50 50},clip,width=0.077\textwidth]{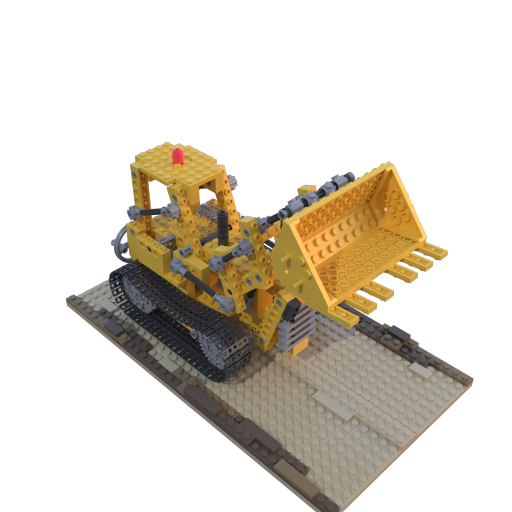}

\makebox[5pt]{\rotatebox{-90}{\hspace{-36pt}\footnotesize{Rendering}}}

 \\

{\footnotesize{33.78}}
 & 
{\footnotesize{41.40}}
 & 
{\footnotesize{42.01}}
 & 
{\footnotesize{\textbf{42.84}}}
 &  &  & 
{\footnotesize{30.43}}
 & 
{\footnotesize{41.93}}
 & 
{\footnotesize{42.49}}
 & 
{\footnotesize{\textbf{44.15}}}
 &  \\

\includegraphics[trim={50 50 50 50},clip,width=0.077\textwidth]{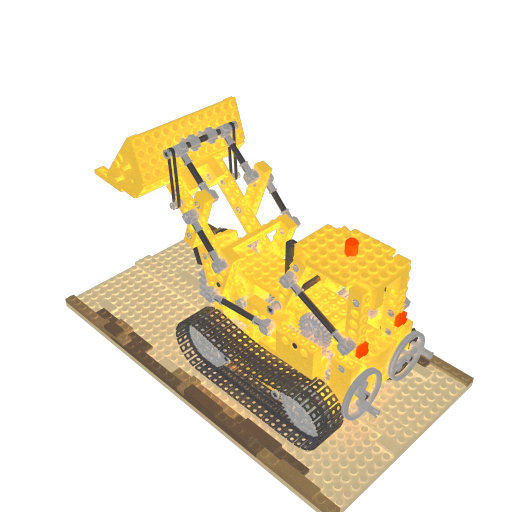}
 & 
\includegraphics[trim={50 50 50 50},clip,width=0.077\textwidth]{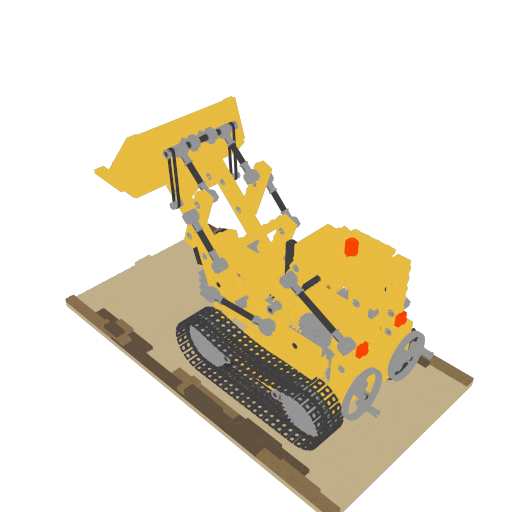}
 & 
\includegraphics[trim={50 50 50 50},clip,width=0.077\textwidth]{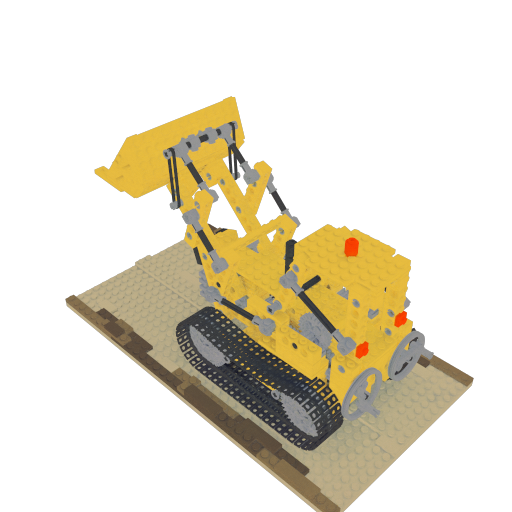}
 & 
\includegraphics[trim={50 50 50 50},clip,width=0.077\textwidth]{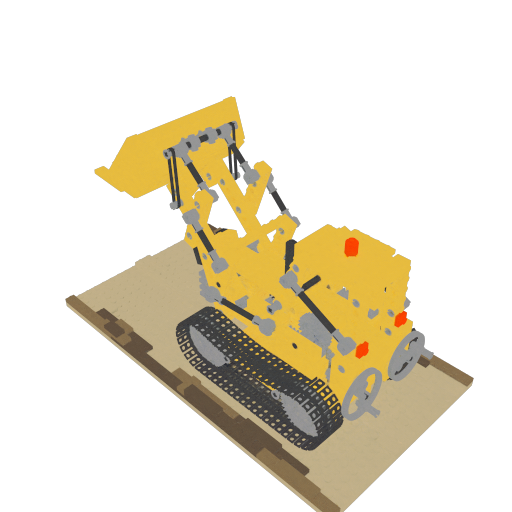}
 & 
\includegraphics[trim={50 50 50 50},clip,width=0.077\textwidth]{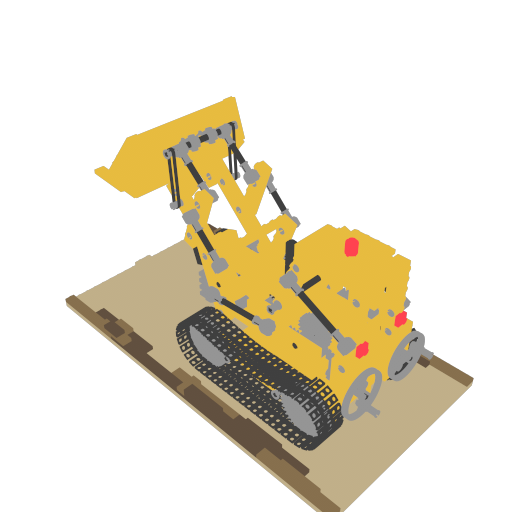}
 &  & 
\includegraphics[trim={50 50 50 50},clip,width=0.077\textwidth]{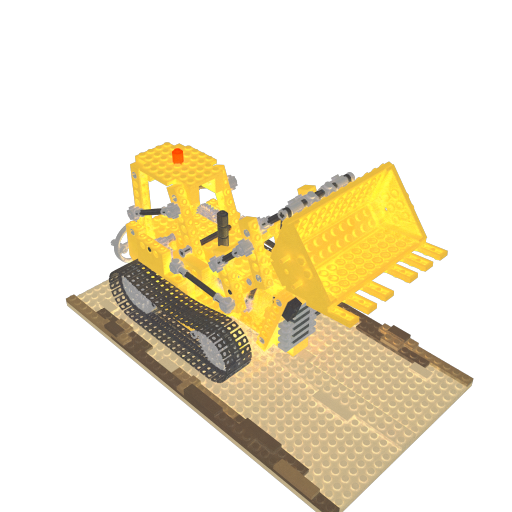}
 & 
\includegraphics[trim={50 50 50 50},clip,width=0.077\textwidth]{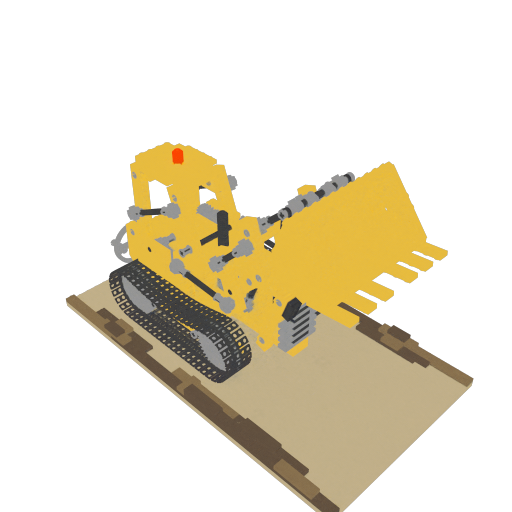}
 & 
\includegraphics[trim={50 50 50 50},clip,width=0.077\textwidth]{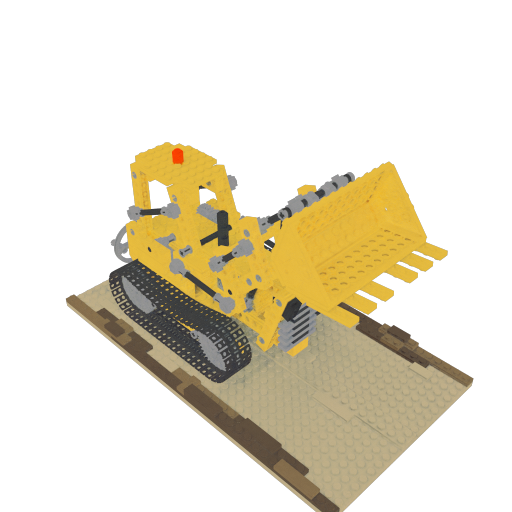}
 & 
\includegraphics[trim={50 50 50 50},clip,width=0.077\textwidth]{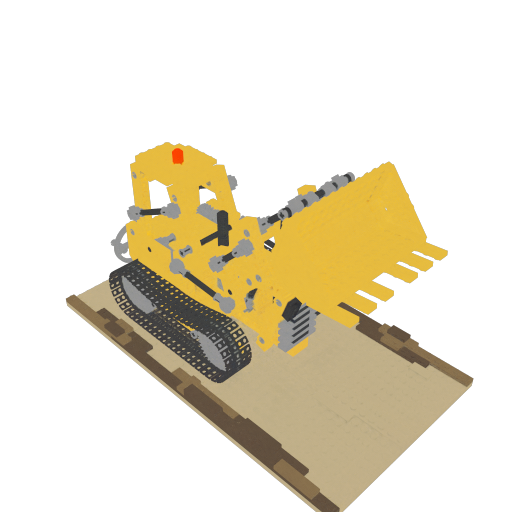}
 & 
\includegraphics[trim={50 50 50 50},clip,width=0.077\textwidth]{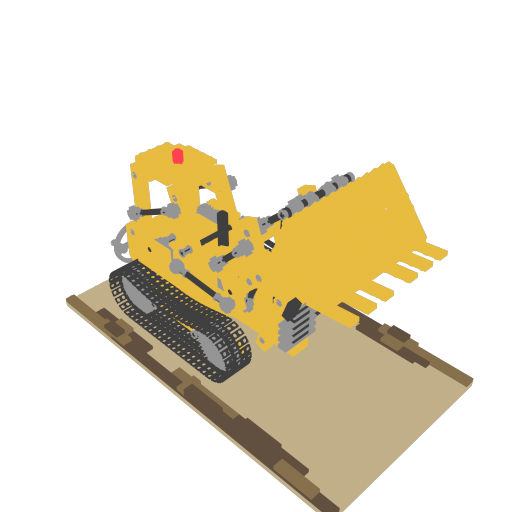}

\makebox[5pt]{\rotatebox{-90}{\hspace{-33pt}\footnotesize{Albedo}}}

 \\

{\footnotesize{23.10}}
 & 
{\footnotesize{\textbf{42.64}}}
 & 
{\footnotesize{31.48}}
 & 
{\footnotesize{41.97}}
 &  &  & 
{\footnotesize{22.60}}
 & 
{\footnotesize{\textbf{44.83}}}
 & 
{\footnotesize{32.31}}
 & 
{\footnotesize{43.79}}
 &  \\

\includegraphics[trim={50 50 50 50},clip,width=0.077\textwidth]{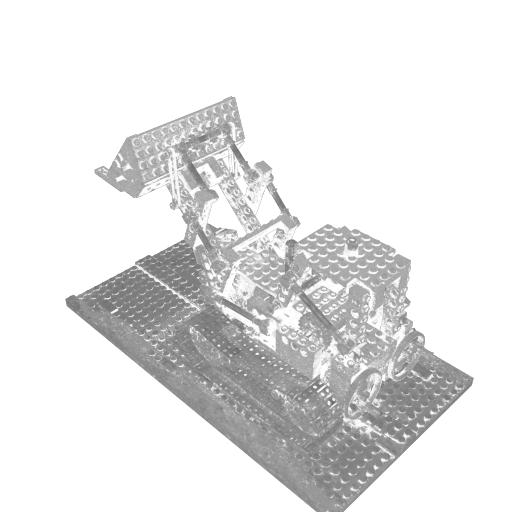}
 & 
\includegraphics[trim={50 50 50 50},clip,width=0.077\textwidth]{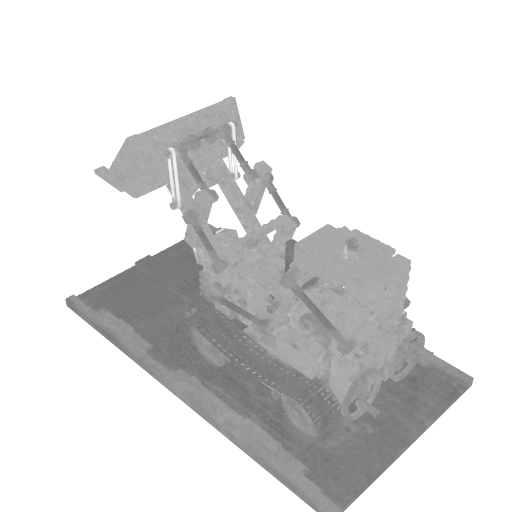}
 & 
\includegraphics[trim={50 50 50 50},clip,width=0.077\textwidth]{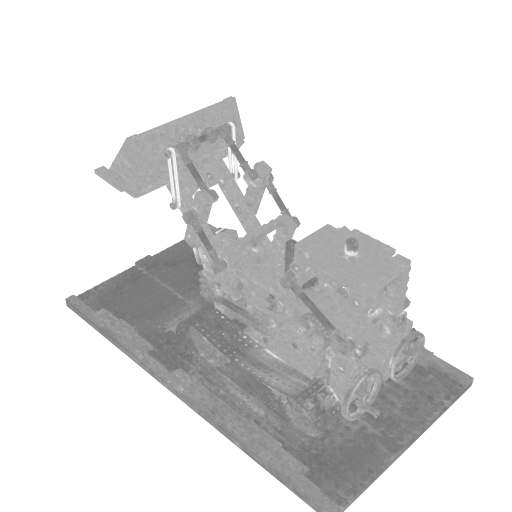}
 & 
\includegraphics[trim={50 50 50 50},clip,width=0.077\textwidth]{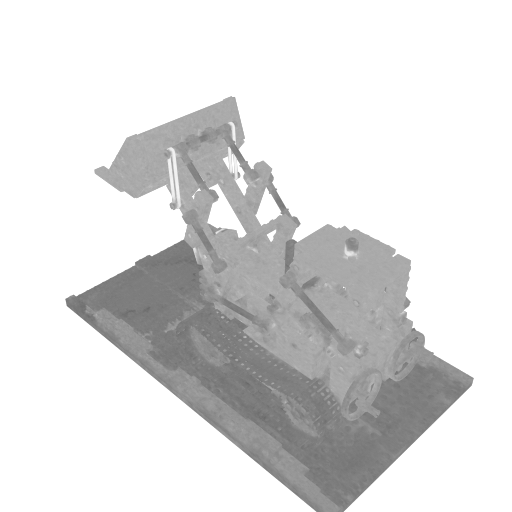}
 & 
\includegraphics[trim={50 50 50 50},clip,width=0.077\textwidth]{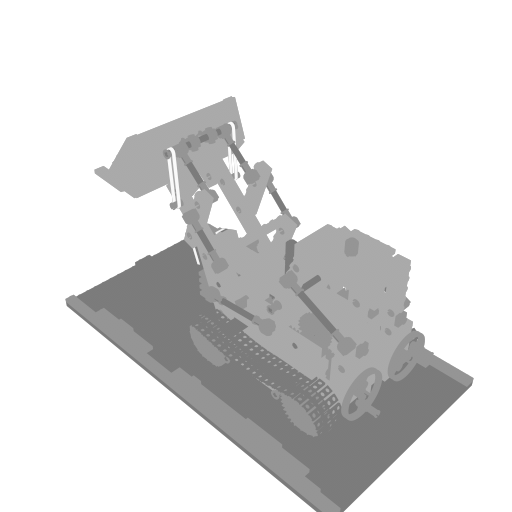}
 &  & 
\includegraphics[trim={50 50 50 50},clip,width=0.077\textwidth]{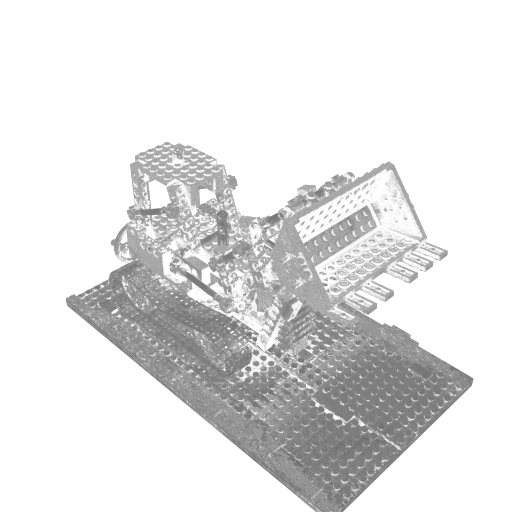}
 & 
\includegraphics[trim={50 50 50 50},clip,width=0.077\textwidth]{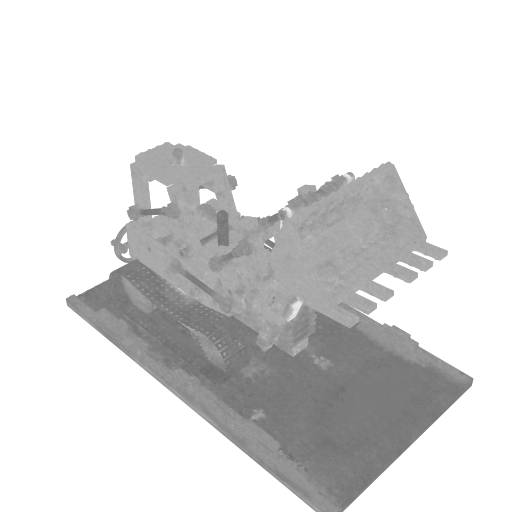}
 & 
\includegraphics[trim={50 50 50 50},clip,width=0.077\textwidth]{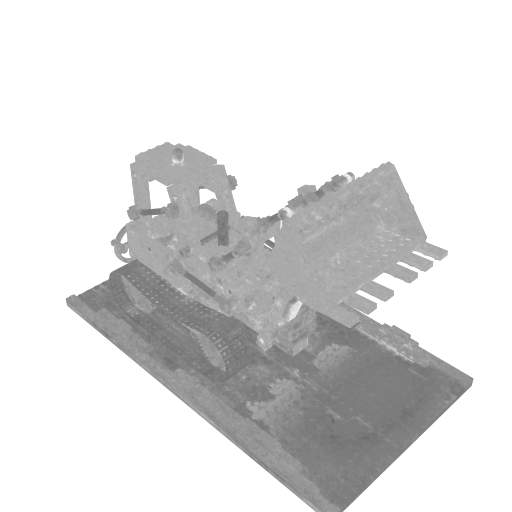}
 & 
\includegraphics[trim={50 50 50 50},clip,width=0.077\textwidth]{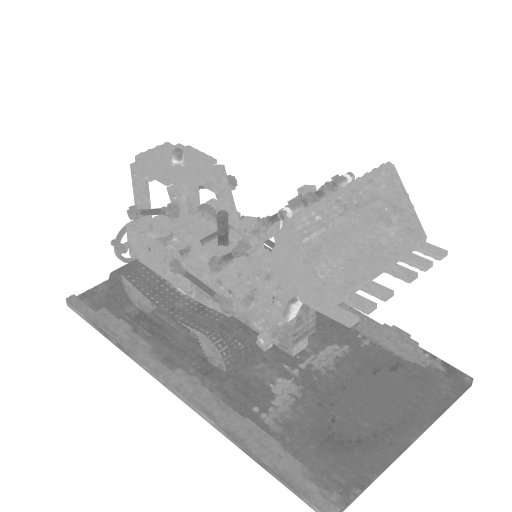}
 & 
\includegraphics[trim={50 50 50 50},clip,width=0.077\textwidth]{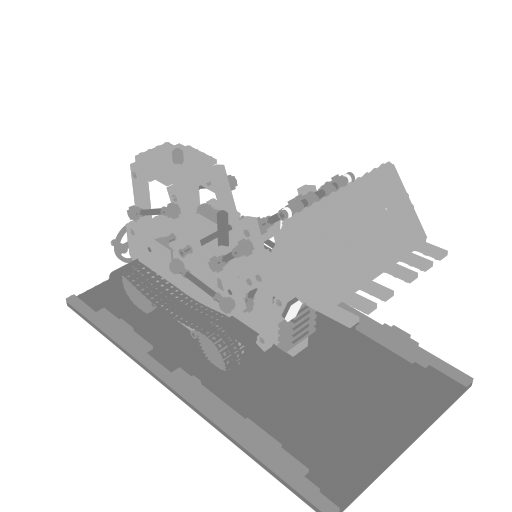}

\makebox[5pt]{\rotatebox{-90}{\hspace{-36pt}\footnotesize{Roughness}}}

 \\

{\footnotesize{18.43}}
 & 
{\footnotesize{\textbf{35.22}}}
 & 
{\footnotesize{31.12}}
 & 
{\footnotesize{34.61}}
 &  &  & 
{\footnotesize{17.69}}
 & 
{\footnotesize{\textbf{35.09}}}
 & 
{\footnotesize{32.19}}
 & 
{\footnotesize{33.83}}
 &  \\
\end{tabular}
\end{subfigure}

\endgroup

    \vspace{-0.3cm}
    \caption{
    \textbf{Main results.} \mycaption{For each scene, we compare the rendering, recovered albedo, and recovered roughness for direct illumination, PRB, and AD-Ours. We also compare to the case where the radiance cache is trained without the prior. We show two different views of each scene, and report PSNR to ground truth (GT). For \Staircase, \textit{Ours without prior} diverges at $70\%$ progress and we report the best results. Additional scenes are in supplementary.
    }
    }
    \label{fig:main_results}
\end{figure*}

We present results across all methods when reconstructing the albedo and roughness of the indoor and NeRF scenes in \autoref{fig:main_results} and supplementary.
In most cases, our method faithfully recovers the parameters thanks to its use of a radiometric prior that correctly accounts for global illumination.
Despite its simplicity, our method achieves comparable (and sometimes better) results to PRB. \newtext{Compared to the premise of Zhang et. al \shortcite{inverse_rendering_gi}, which uses only ground truth images to train the radiance network, our method converges to a better solution. This is because we optimize the radiance network to respect the radiometric prior in additional spatio-directional locations, rather than just fitting it to the pixels in the input images.}

{\em Performance and memory use.}
A key component of our method is the neural radiance field $L_\theta$.
We use it to compute the measurement integral (see \autoref{eq:RHS-measurement}), whereas AD-PT and PRB instead solve the path integral using an MC estimator.
The neural field brings two benefits: it bounds the memory usage since we only trace a single bounce, and reduces variance; the network prediction itself is noise-free.
This yields improved performance and robustness (see Figure \ref{fig:train_times} and the training curves in supplementary materials).


\begin{figure*}
\centering
\captionsetup[subfigure]{labelformat=empty}
\begingroup
\renewcommand{\arraystretch}{0.6}
\setlength{\tabcolsep}{0.1em}

\newlength{\biaswidth}
\setlength{\biaswidth}{1.57cm}
\newlength{\heatmapwidth}
\setlength{\heatmapwidth}{1.8cm}

\begin{subfigure}[b]{\textwidth}
\begin{tabular}{ccccccccccccc}

{\hspace{0.18cm}\footnotesize Direct Illum.} & & \multicolumn{5}{c}{\footnotesize PRB} & & \multicolumn{3}{c}{\footnotesize Ours} & & {\footnotesize Reference} \\
\cmidrule(l){1-1} \cmidrule{3-7} \cmidrule{9-11} \cmidrule{13-13}

{\hspace{0.18cm} \footnotesize 1} & &
{\footnotesize 2} &
{\footnotesize 4} &
{\footnotesize 10} &
{\footnotesize 15} &
{\footnotesize 30} & &
{\footnotesize 1} &
{\footnotesize 2} &
{\footnotesize 4} & &
{\footnotesize 128} \\

\hspace{0.18cm} \includegraphics[height=\biaswidth]{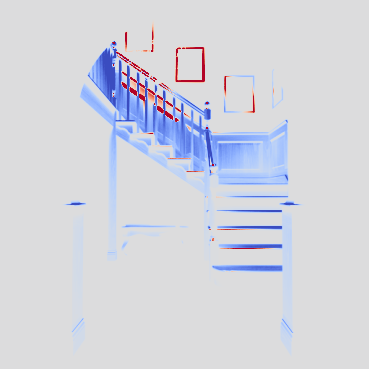} & &
\includegraphics[height=\biaswidth]{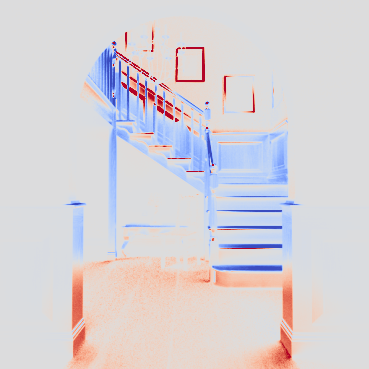} &
\includegraphics[height=\biaswidth]{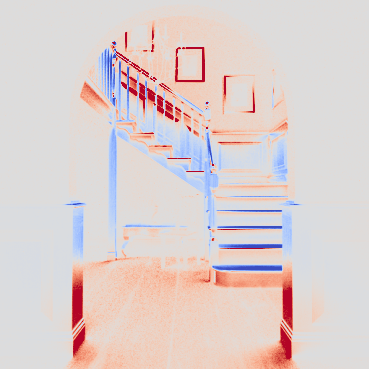} &
\includegraphics[height=\biaswidth]{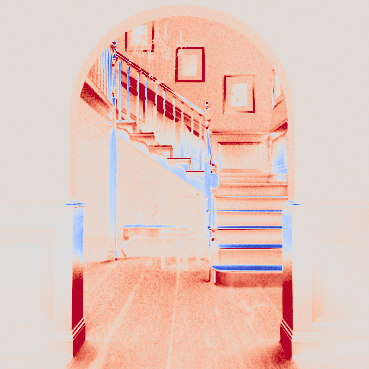} &
\includegraphics[height=\biaswidth]{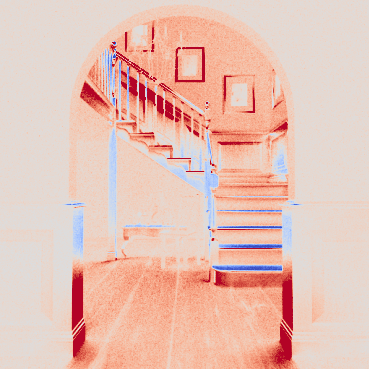} &
\includegraphics[height=\biaswidth]{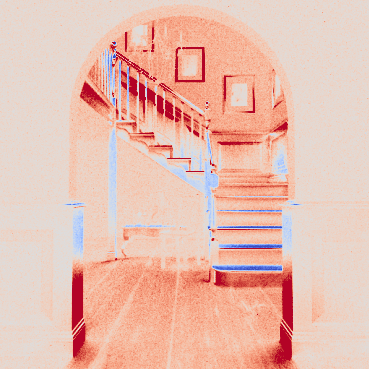} & &
\includegraphics[height=\biaswidth]{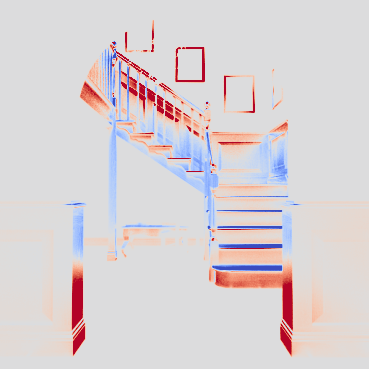} &
\includegraphics[height=\biaswidth]{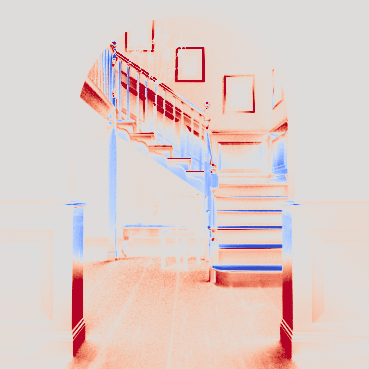} &
\includegraphics[height=\biaswidth]{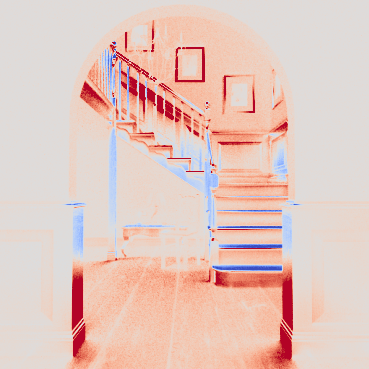} & &
\includegraphics[height=\biaswidth]{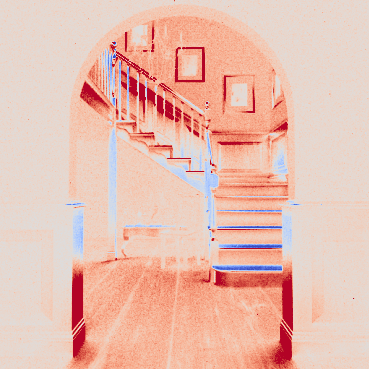} \\

\includegraphics[height=\heatmapwidth]{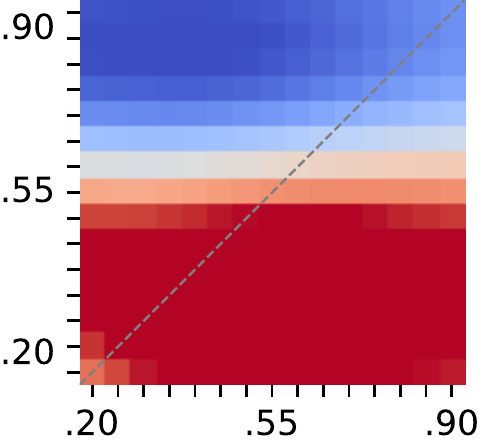} & &
\includegraphics[height=\heatmapwidth]{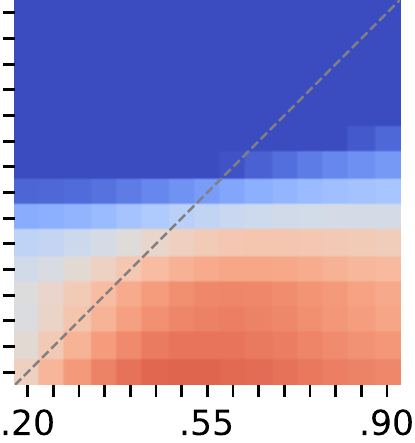} &
\includegraphics[height=\heatmapwidth]{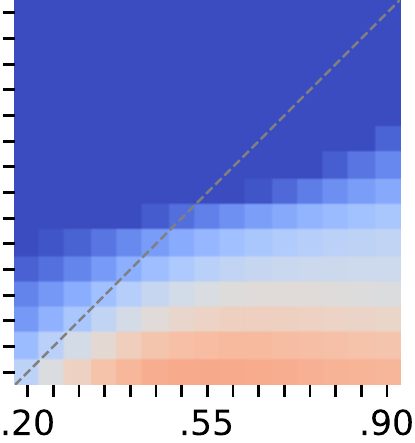} &
\includegraphics[height=\heatmapwidth]{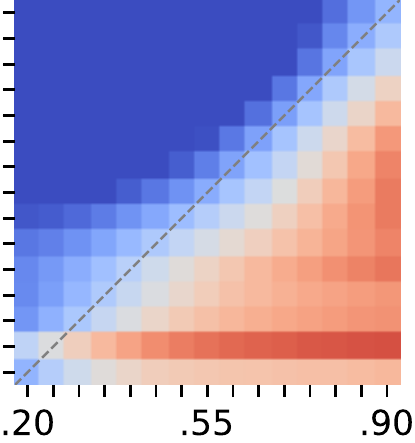} &
\includegraphics[height=\heatmapwidth]{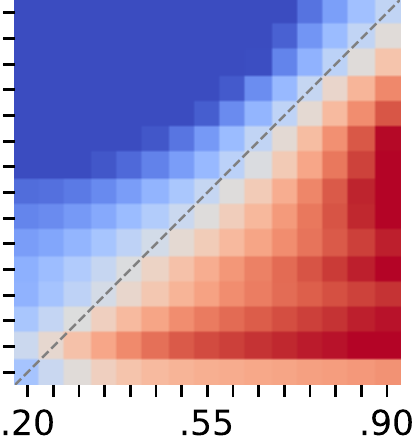} &
\includegraphics[height=\heatmapwidth]{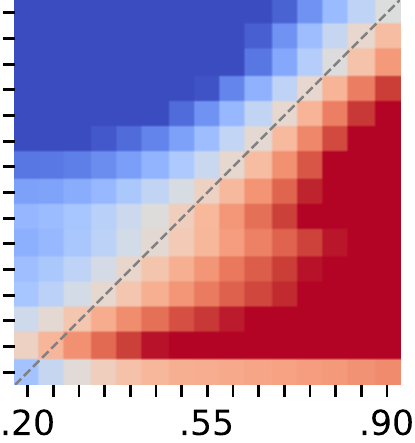} & &
\includegraphics[height=\heatmapwidth]{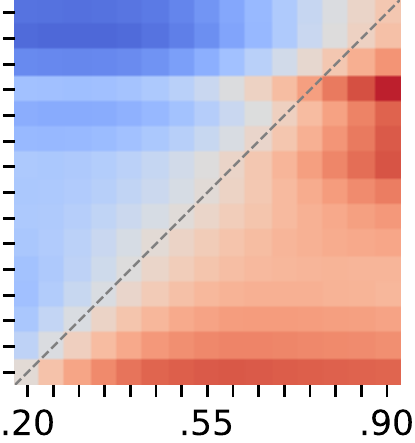} &
\includegraphics[height=\heatmapwidth]{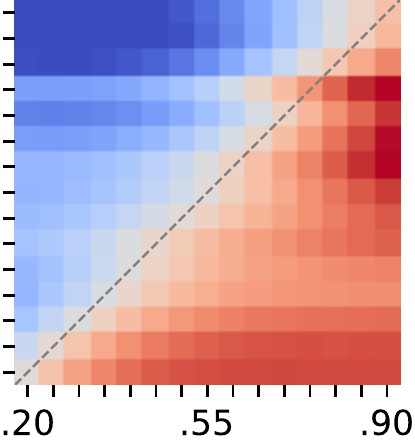} &
\includegraphics[height=\heatmapwidth]{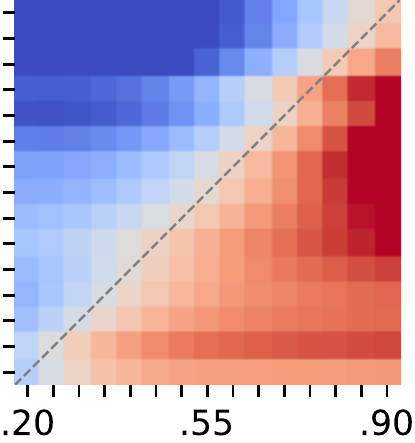} & &
\includegraphics[height=\heatmapwidth]{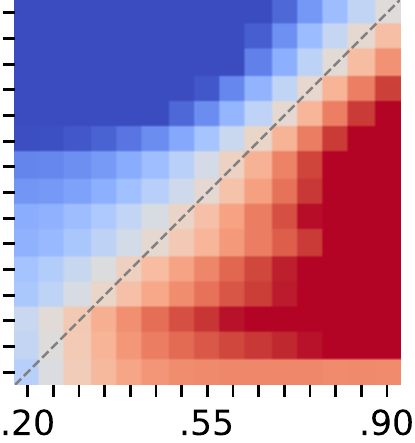} \\

\end{tabular}
\end{subfigure}

\endgroup
\vspace{-0.3cm}
\caption{
\newtext{
\textbf{Analysis of biased gradients.} \mycaption{We visualize gradients with respect to the roughness parameter of the wooden material on the staircase and the picture frames, using different methods. We cap the maximum path length for PRB at each specified number. For our approach, we perform a certain number $k$ of \emph{differentiable} bounces before querying our radiance cache. \textbf{The first row} visualizes per-pixel gradient magnitudes with respect to roughness ($\frac{\partial I}{\partial \phi}$ in Equation \eqref{eq:inverse-chain-rule}) at a value of 0.5. We can see that the roughness of the staircase leads to indirect gradients on pixels on the floor. These indirect gradients are missed by our method using only $k=1$ differentiable bounce, but accounted for when tracing more differentiable bounces ($k=2,4$). Note that at equal number of bounces, our method provides more accurate global information than PRB because our radiance field accurately captures global illumination, irrespective of the number of differentiable bounces $k$. \textbf{The second row} visualizes the magnitude of the gradient of the L2 loss with respect to the roughness parameters over the entire image, comparing the current state with the target state ($\frac{\partial \mathcal{L}}{\partial \phi}$ in Equation \eqref{eq:inverse-chain-rule}). For unbiased gradients, cells on the diagonal have gradients equal to zero, and non-zero values indicate bias. As shown, direct illumination, PRB with low numbers of bounces, and our method provide biased gradients. Increasing the number $k$ of differentiable bounces reduces bias in our gradients. All results in this paper use 'Ours' with $k=1$ differentiable bounce, but we sample the residual at an extra bounce. \textbf{In all visualizations}, red indicates negative and blue indicate positive values.} }
}
\label{fig:bias}
\end{figure*}

\begin{table}
    \centering
    \caption{\newtext{\textbf{\textit{Cube} scene path lengths.}} \mycaption{We report the average and maximum path length for path tracing in the {\textit Cube} scene depending on surface albedo.}}
    \setlength{\tabcolsep}{3pt}
    \begin{small}
    \begin{tabular}{lcccccc}
        \toprule
        Albedo & 0.3 & 0.5 & 0.7 & 0.9 & 0.95 & 0.97 \\
        \midrule
        Average path length & 1.43 & 1.98 & 3.30 & 9.89 & 19.76 & 32.89 \\
        Limit & 5 & 7 & 13 & 42 & 84 & 140 \\
        \bottomrule
    \end{tabular}
    \end{small}
    \label{tab:cube_pathlengths}
    \vspace{-1mm}
\end{table}
\begin{figure}
\vspace{-2mm}
    \centering
    \includegraphics[width=0.23\textwidth]{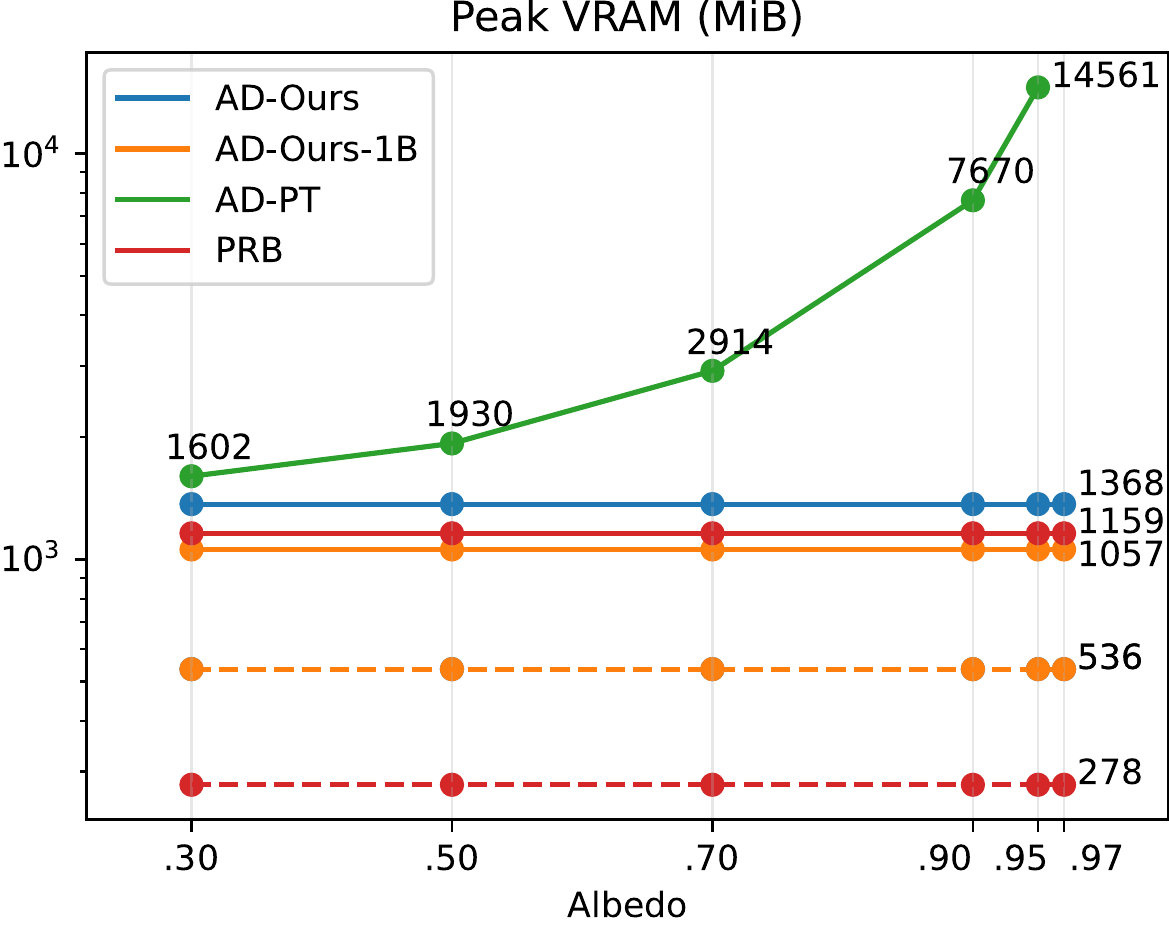}
    \includegraphics[width=0.23\textwidth]{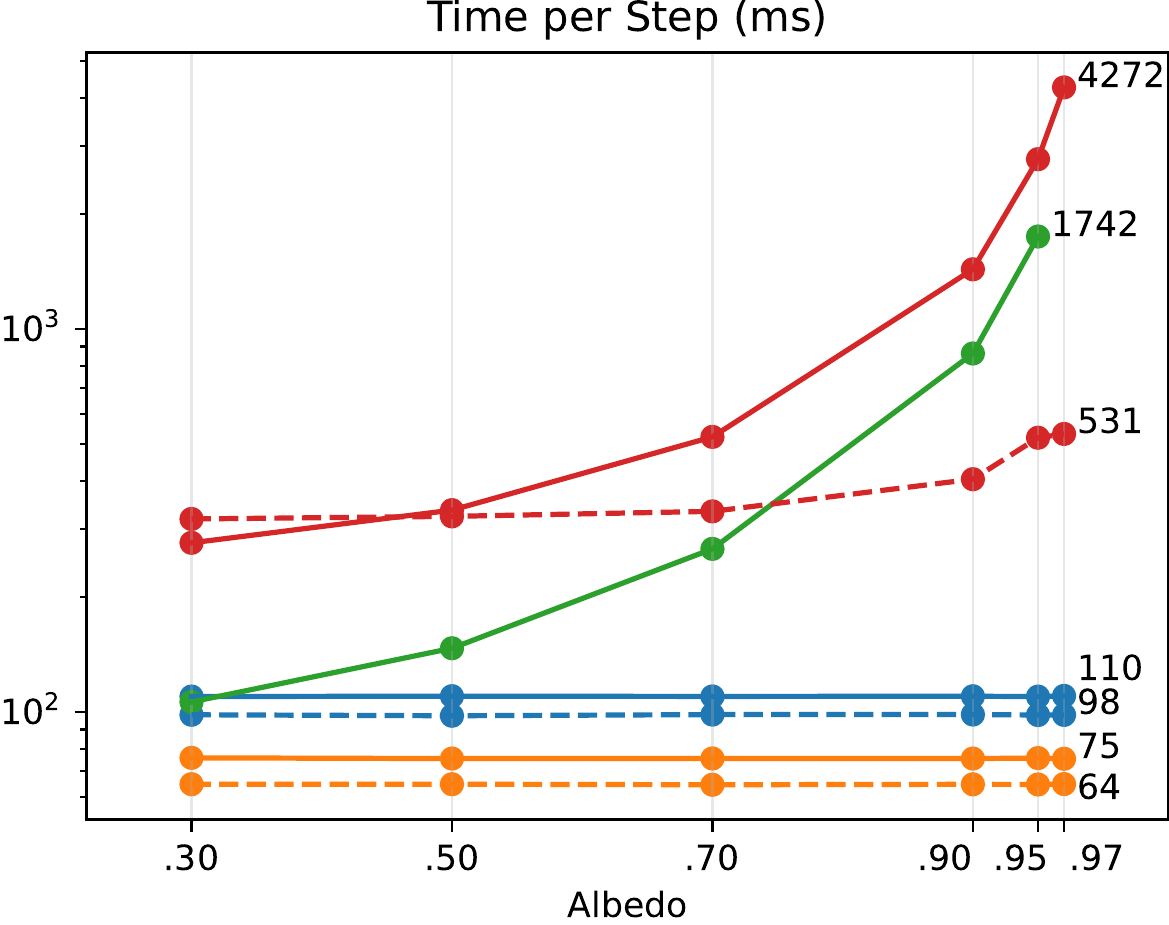}
    \caption{
    \textbf{\newtext{\textit{Cube} scene measurements.}} \mycaption{We compare the time and memory consumption of each training step, with all parameters represented in dense grids. Dotted lines have Mitsuba 3 mega kernels enabled. AD-PT runs out of VRAM (24GB) at albedo $0.97$. Our method uses a constant amount of VRAM and time.}
    }
    \label{fig:perf_mitsuba}
    \vspace{-3mm}
\end{figure}

\newtext{
{\em Time and memory consumption.}
A key benefit of our approach is that the memory and time usage of each training step is independent of the light path lengths necessary to represent global illumination.
We illustrate this on the \textit{Cube} scene where the cameras are placed at the corners inside a closed cube, the walls have a uniform, diffuse material, and the only light source is an area light on the ceiling. By varying the albedo of the walls, we can control the average path length in a path tracer as shown in Table \ref{tab:cube_pathlengths}; implementation details are provided in the supplementary document.
In Fig.~\ref{fig:perf_mitsuba} we report how time and memory consumption of all methods change as the albedo increases from $0.3$ to $0.97$. To avoid overhead between Mitsuba and PyTorch and to be able to use a mega-kernel, we conduct measurements using Mitsuba dense grids representing the scene parameters and radiance (results obtained with MLPs can be found in the supplementary document). We observe that both our method and PRB use a constant amount of memory, while memory requirements for AD-PT grow with increasing albedo, i.e., increasing average path length. Our method, however, runs faster than PRB and AD-PT, and in constant time independent of albedo.
}

\newtext{
\textit{Biased gradients.}
Unlike path integral methods with sufficiently many bounces, our method provides biased gradients due to two reasons. First, the neural radiance field is only an approximation of the true radiance field, which in itself causes bias. Furthermore, our method introduces bias because we do not differentiate the neural radiance field itself with respect to the scene parameters. More formally, in Equation \eqref{eq:differntial-rendering-equation} the term $\partial_{\phi}L$ is assumed to be zero.

We illustrate the effect of this approximation in Figure~\ref{fig:bias}. Note that we can easily generalize our method by replacing the term $E + \mathcal{T}(L_\theta)$ in Equation~\ref{eq:RHS-measurement} with a truncated series expansion $\sum_{i=0}^{k-1} \mathcal{T}^i(E) +\mathcal{T}^k(L_\theta)$. For a small $k>1$, we can obtain additional differentiable light bounces with standard AD, and increasing $k$ reduces bias in the gradients. Figure~\ref{fig:bias} shows how setting $k=2, 4$ reduces bias in our gradients. In practice, we use $k=1$ as in Equation~\ref{eq:RHS-measurement} for all results in this paper since this is sufficient to obtain good results.
}

\subsection{Ablation Study}
\begin{figure*}
    \centering
    \captionsetup[subfigure]{labelformat=empty}
    \begingroup
\renewcommand{\arraystretch}{0.6}
\setlength{\tabcolsep}{0.1em}

\makebox[5pt]{\rotatebox{90}{\hspace{-10pt} \footnotesize{\Kitchen}}}
\begin{subfigure}[b]{0.98\textwidth}
\begin{tabular}{cccccccc}

{\footnotesize{AD-Direct}}
 & 
{\footnotesize{\begin{tabular}{@{}c@{}}w/ Radiometric Prior \\ (Eq. (10)) \end{tabular}}}
 & 
{\footnotesize{\begin{tabular}{@{}c@{}}w/ Stop gradient \\ prior (Sec. 4.2)\end{tabular}}}
 & 
{\footnotesize{\begin{tabular}{@{}c@{}}w/ Second-bounce \\ prior (Sec. 4.2)\end{tabular}}}
 & 
{\footnotesize{\begin{tabular}{@{}c@{}}w/ LHS loss \\ (AD-Ours, Sec 4.3)\end{tabular}}}
 & 
{\footnotesize{\begin{tabular}{@{}c@{}}AD-Ours \\ w/o Prior\end{tabular}}}
 & 
{\footnotesize{GT}}
 \\

\includegraphics[width=0.135\textwidth]{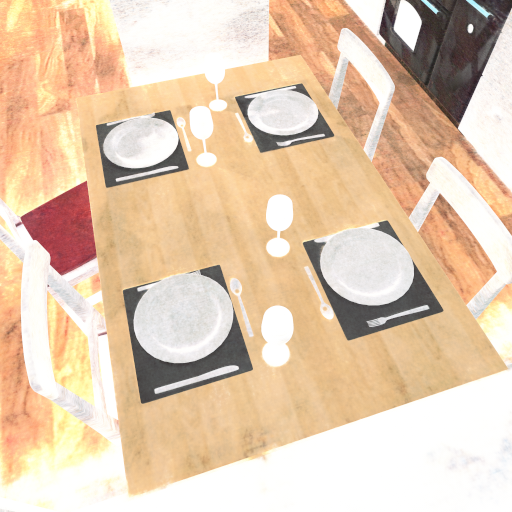}
 & 
\includegraphics[width=0.135\textwidth]{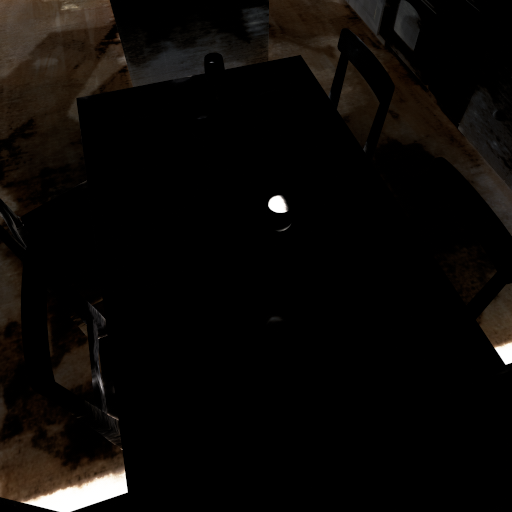}
 & 
\includegraphics[width=0.135\textwidth]{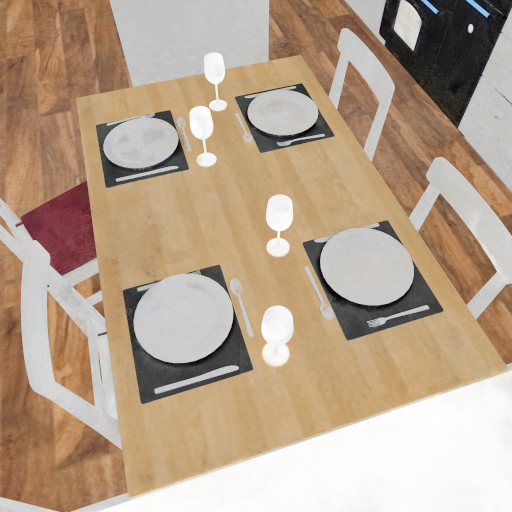}
 & 
\includegraphics[width=0.135\textwidth]{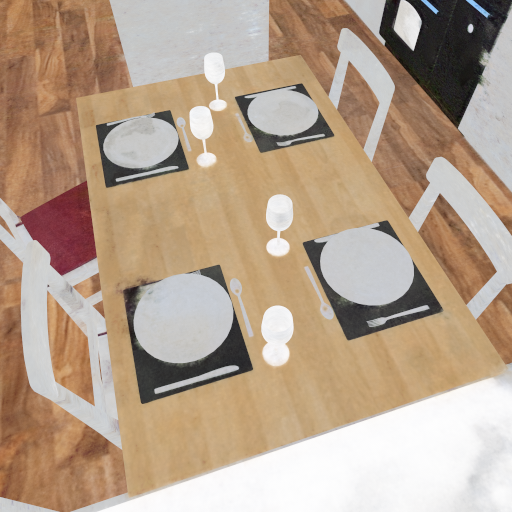}
 & 
\includegraphics[width=0.135\textwidth]{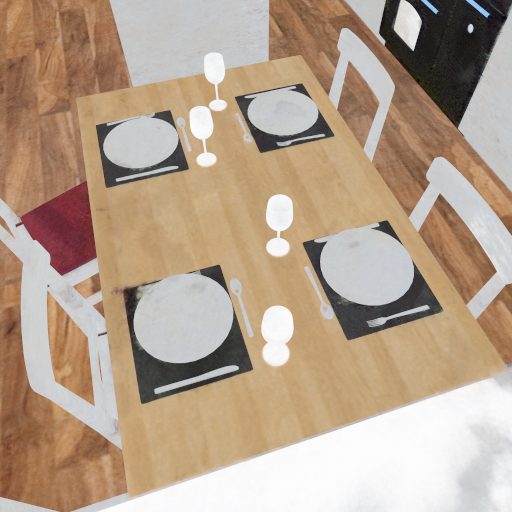}
 & 
\includegraphics[width=0.135\textwidth]{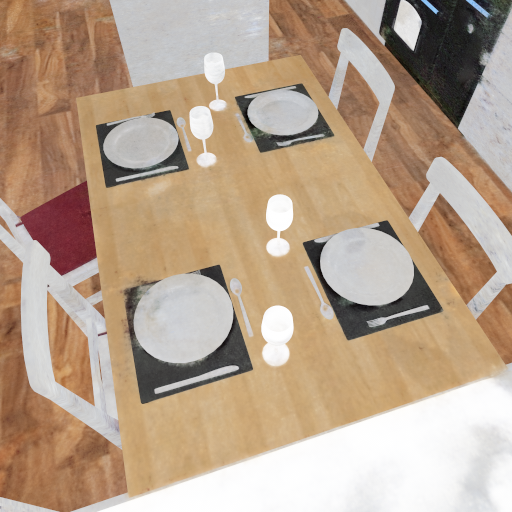}
 & 
\includegraphics[width=0.135\textwidth]{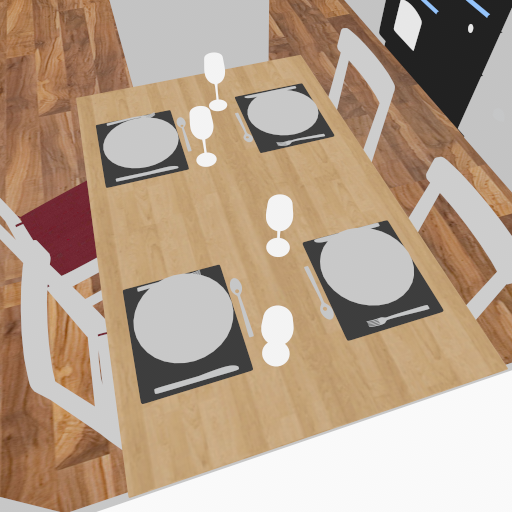}

\makebox[5pt]{\rotatebox{-90}{\hspace{-42pt}\footnotesize{Albedo}}}

 \\

{\footnotesize{MAPE: 0.9503}}
 & 
{\footnotesize{0.9630}}
 & 
{\footnotesize{0.0833}}
 & 
{\footnotesize{0.0895}}
 & 
{\footnotesize{\textbf{0.0777}}}
 & 
{\footnotesize{0.1107}}
 &  \\

\includegraphics[width=0.135\textwidth]{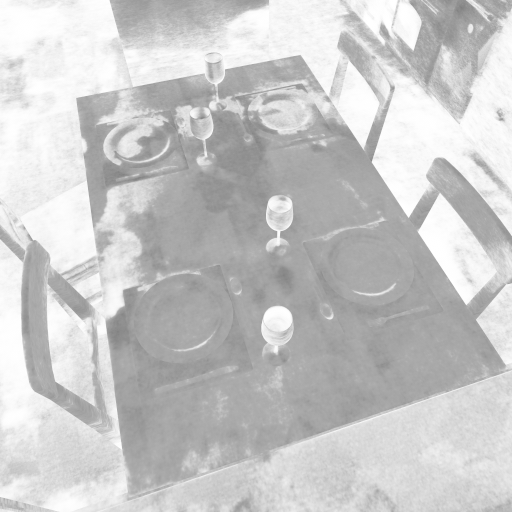}
 & 
\includegraphics[width=0.135\textwidth]{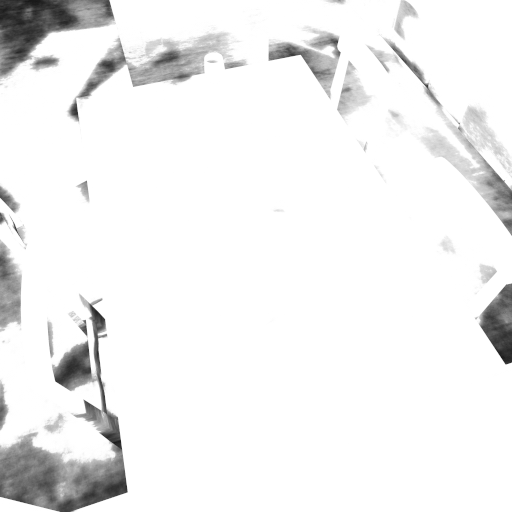}
 & 
\includegraphics[width=0.135\textwidth]{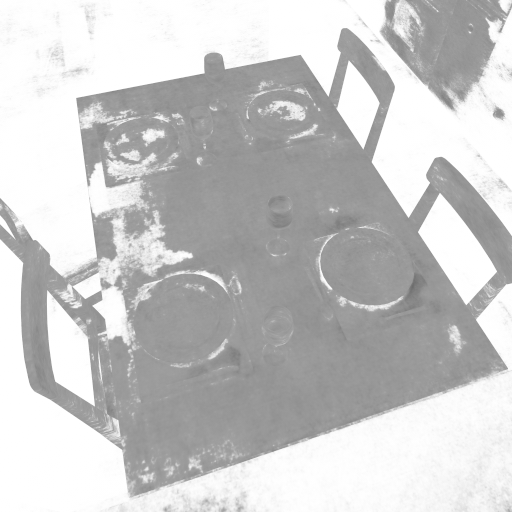}
 & 
\includegraphics[width=0.135\textwidth]{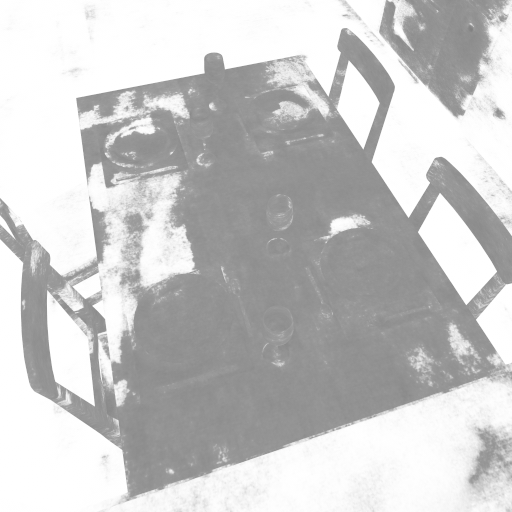}
 & 
\includegraphics[width=0.135\textwidth]{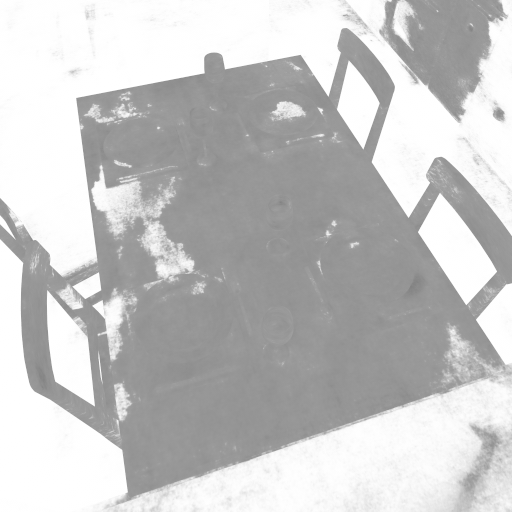}
 & 
\includegraphics[width=0.135\textwidth]{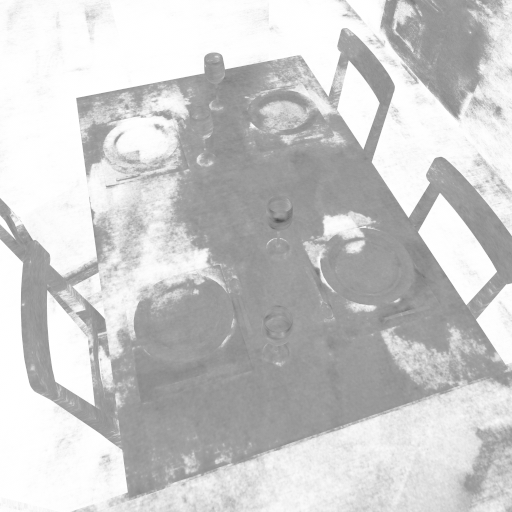}
 & 
\includegraphics[width=0.135\textwidth]{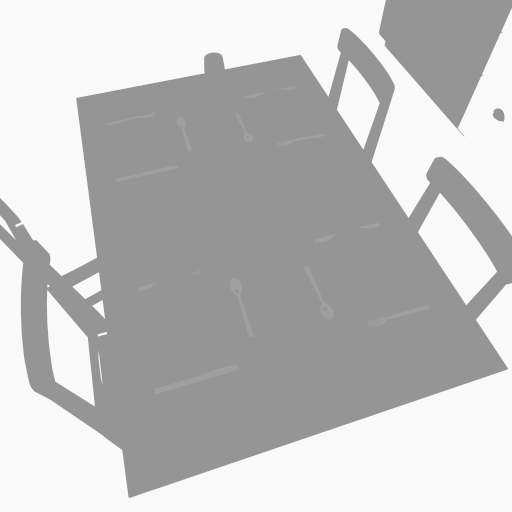}

\makebox[5pt]{\rotatebox{-90}{\hspace{-47pt}\footnotesize{Roughness}}}

 \\

{\footnotesize{0.3286}}
 & 
{\footnotesize{1.3354}}
 & 
{\footnotesize{0.1448}}
 & 
{\footnotesize{0.1762}}
 & 
{\footnotesize{\textbf{0.1092}}}
 & 
{\footnotesize{0.2612}}
 &  \\
\midrule
\end{tabular}
\end{subfigure}

\makebox[5pt]{\rotatebox{90}{\hspace{-10pt} \footnotesize{\Lego}}}
\begin{subfigure}[b]{0.98\textwidth}
\begin{tabular}{cccccccc}

\includegraphics[trim={50 50 50 50},clip,width=0.135\textwidth]{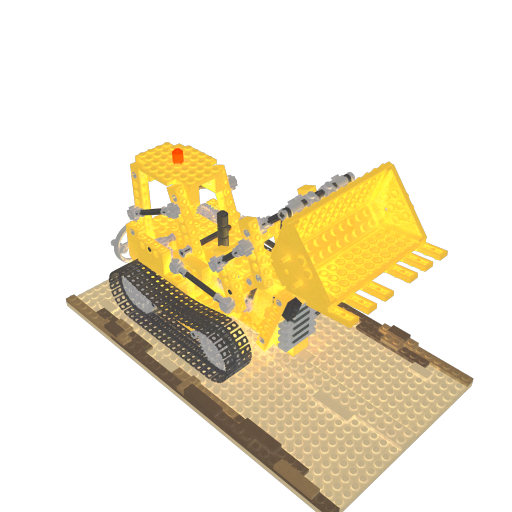}
 & 
\includegraphics[trim={50 50 50 50},clip,width=0.135\textwidth]{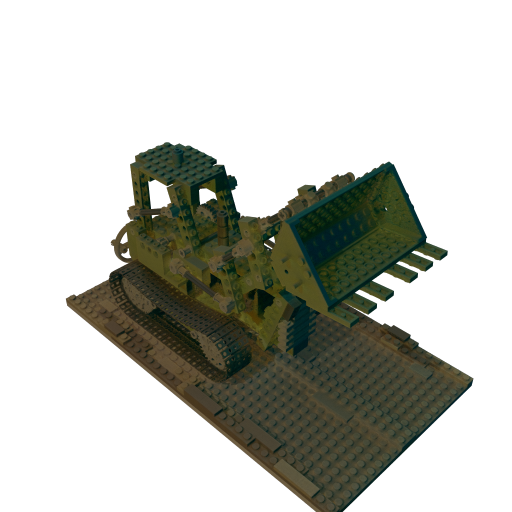}
 & 
\includegraphics[trim={50 50 50 50},clip,width=0.135\textwidth]{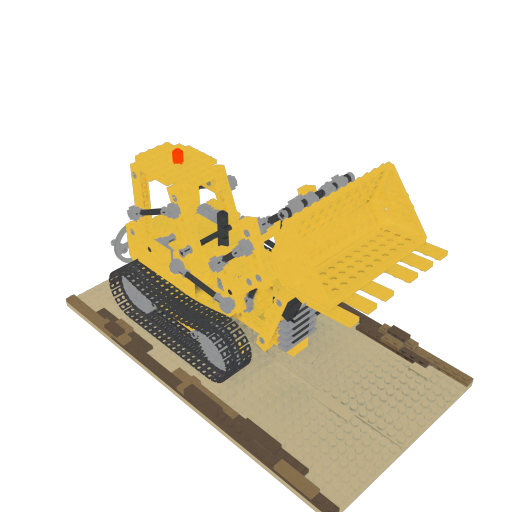}
 & 
\includegraphics[trim={50 50 50 50},clip,width=0.135\textwidth]{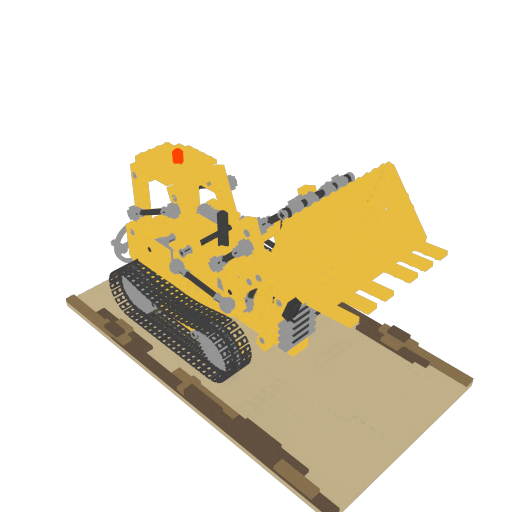}
 & 
\includegraphics[trim={50 50 50 50},clip,width=0.135\textwidth]{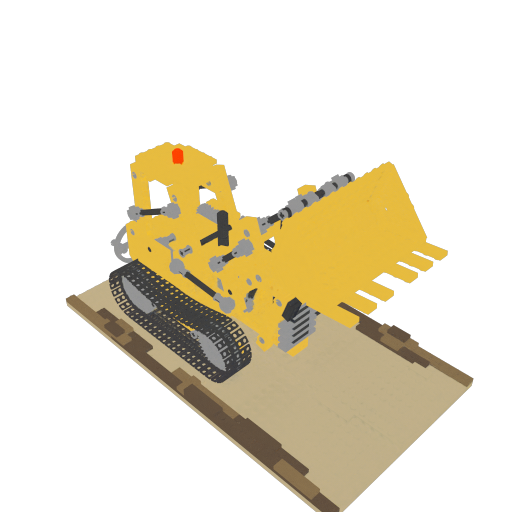}
 & 
\includegraphics[trim={50 50 50 50},clip,width=0.135\textwidth]{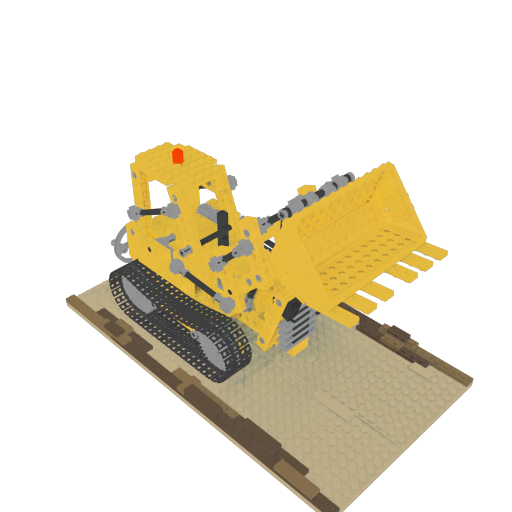}
 & 
\includegraphics[trim={50 50 50 50},clip,width=0.135\textwidth]{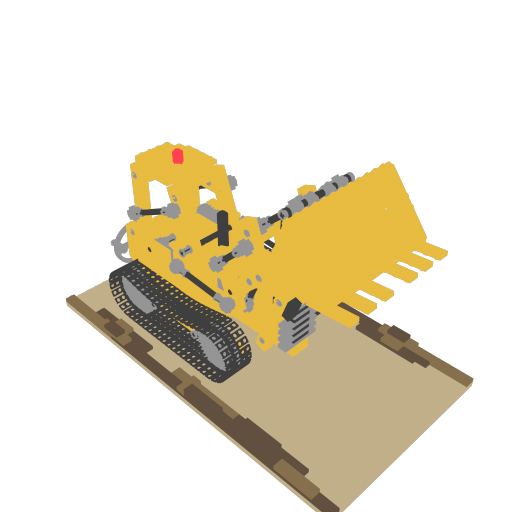}

\makebox[5pt]{\rotatebox{-90}{\hspace{-42pt}\footnotesize{Albedo}}}

 \\

{\footnotesize{MAPE: 0.0731}}
 & 
{\footnotesize{0.2429}}
 & 
{\footnotesize{0.0166}}
 & 
{\footnotesize{\textbf{0.0077}}}
 & 
{\footnotesize{0.0082}}
 & 
{\footnotesize{0.0229}}
 &  \\

\includegraphics[trim={50 50 50 50},clip,width=0.135\textwidth]{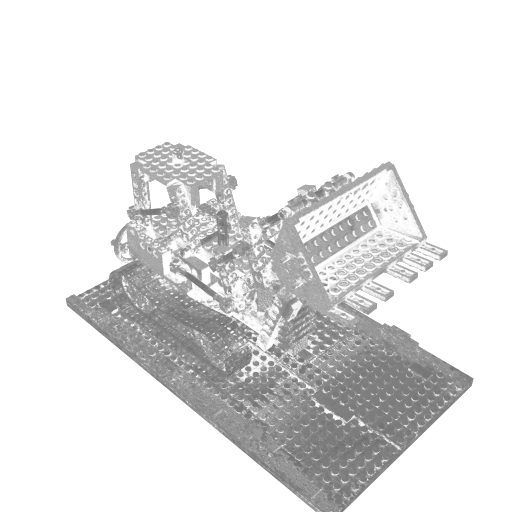}
 & 
\includegraphics[trim={50 50 50 50},clip,width=0.135\textwidth]{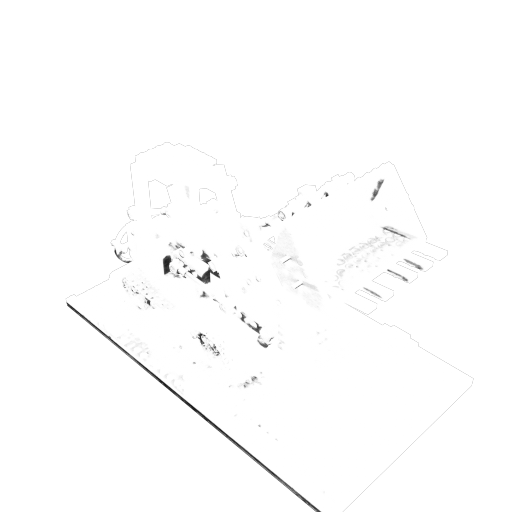}
 & 
\includegraphics[trim={50 50 50 50},clip,width=0.135\textwidth]{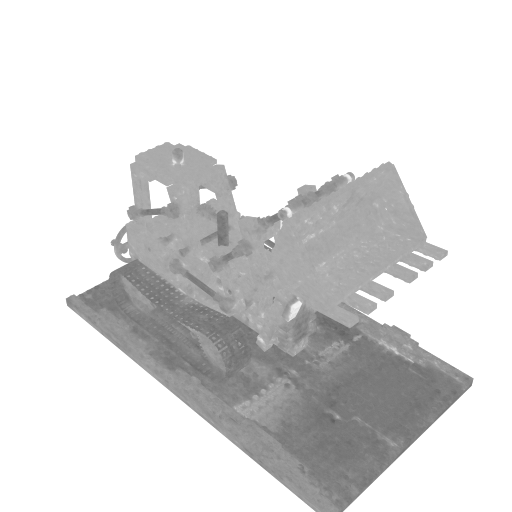}
 & 
\includegraphics[trim={50 50 50 50},clip,width=0.135\textwidth]{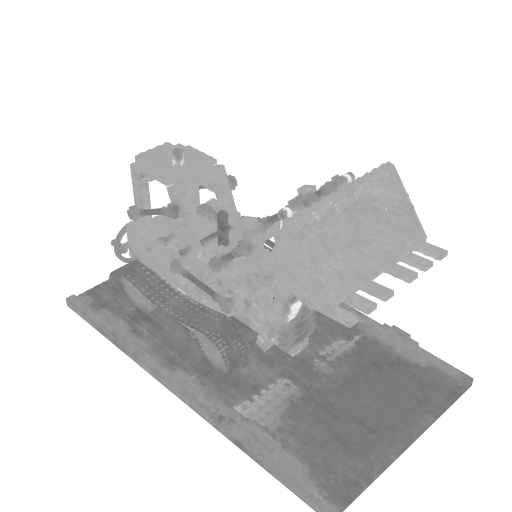}
 & 
\includegraphics[trim={50 50 50 50},clip,width=0.135\textwidth]{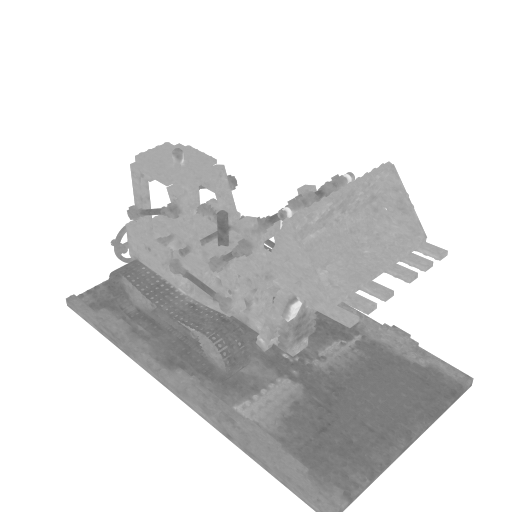}
 & 
\includegraphics[trim={50 50 50 50},clip,width=0.135\textwidth]{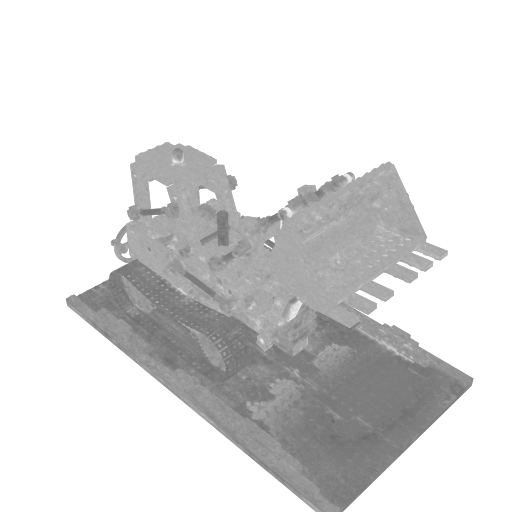}
 & 
\includegraphics[trim={50 50 50 50},clip,width=0.135\textwidth]{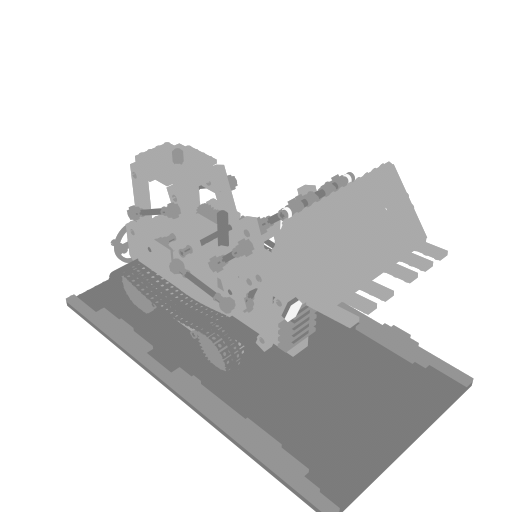}

\makebox[5pt]{\rotatebox{-90}{\hspace{-47pt}\footnotesize{Roughness}}}

 \\

{\footnotesize{0.1443}}
 & 
{\footnotesize{0.8113}}
 & 
{\footnotesize{0.0292}}
 & 
{\footnotesize{\textbf{0.0246}}}
 & 
{\footnotesize{0.0269}}
 & 
{\footnotesize{0.0313}}
 &  \\
\end{tabular}
\end{subfigure}

\endgroup

    \vspace{-0.3cm}
    \caption{\newtext{\textbf{Ablation study of individual components and design decisions.}} \mycaption{We start with the direct illumination integrator (left), and add the radiometric prior to it. The results significantly improve when we ignore the gradients of the prior w.r.t scene parameters. Adding the prior to the second bounce better accounts for additional global illumination effects for areas unseen by the input cameras. Using ground truth data to improve the radiance field further improves the quality. Finally, the second column from the right shows our full method, except that we omit the prior.}}
\label{fig:ablation-materials}
\end{figure*}

\begin{figure*}
    \centering
    \captionsetup[subfigure]{labelformat=empty}
    \begingroup
\renewcommand{\arraystretch}{0.6}
\setlength{\tabcolsep}{0.1em}

\makebox[5pt]{\rotatebox{90}{\hspace{-10pt} \footnotesize{Dragon}}}
\begin{subfigure}[b]{0.98\textwidth}
\begin{tabular}{cccccccc}

{\footnotesize{AD-Direct}}
 & 
{\footnotesize{\begin{tabular}{@{}c@{}}w/ Radiometric Prior \\ (Eq. (10)) \end{tabular}}}
 & 
{\footnotesize{\begin{tabular}{@{}c@{}}w/ Stop gradient \\ prior (Sec. 4.2)\end{tabular}}}
 & 
{\footnotesize{\begin{tabular}{@{}c@{}}w/ Second-bounce \\ prior (Sec. 4.2)\end{tabular}}}
 & 
{\footnotesize{\begin{tabular}{@{}c@{}}w/ LHS loss \\ (AD-Ours, Sec 4.3)\end{tabular}}}
 & 
{\footnotesize{GT}}
 \\

\includegraphics[width=0.16\textwidth]{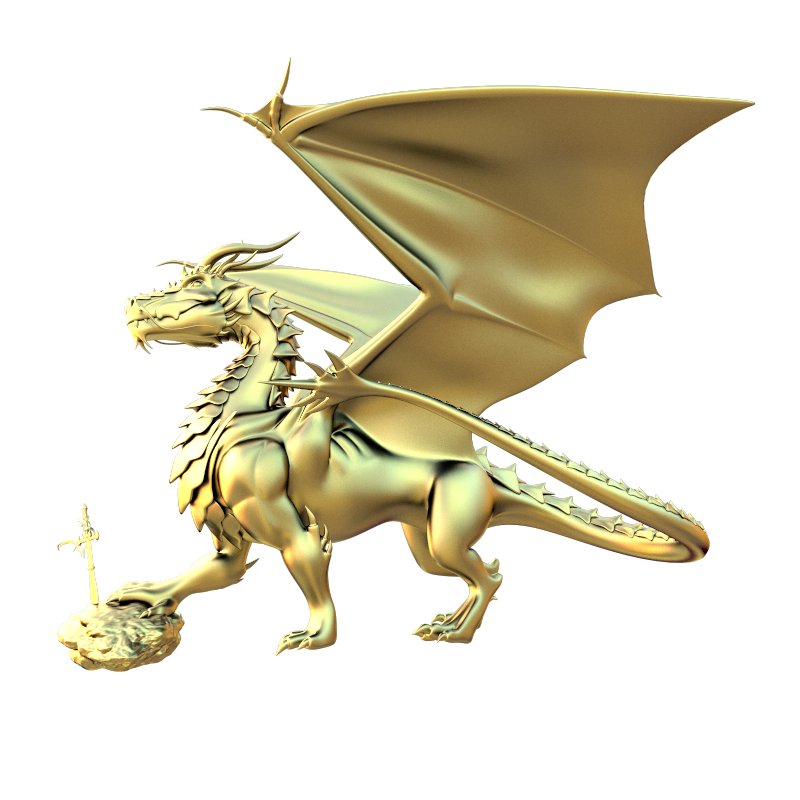}
 & 
\includegraphics[width=0.16\textwidth]{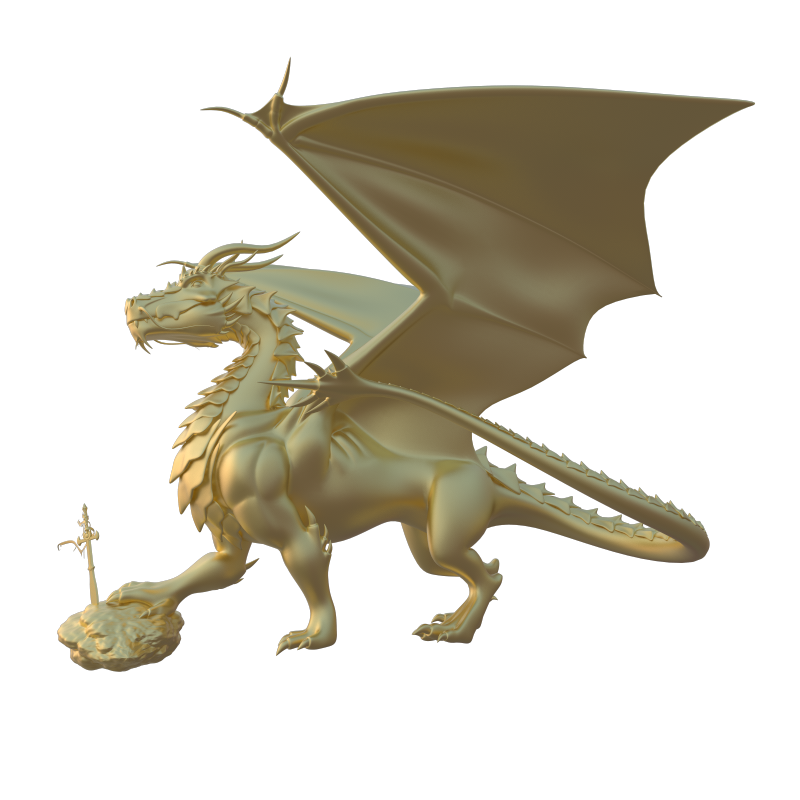}
 & 
\includegraphics[width=0.16\textwidth]{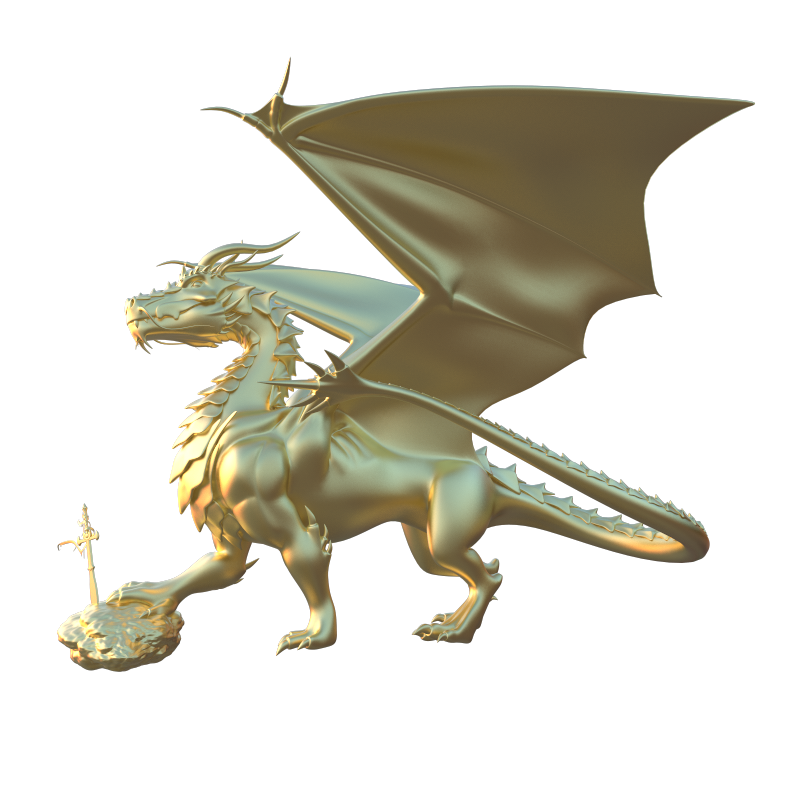}
 & 
\includegraphics[width=0.16\textwidth]{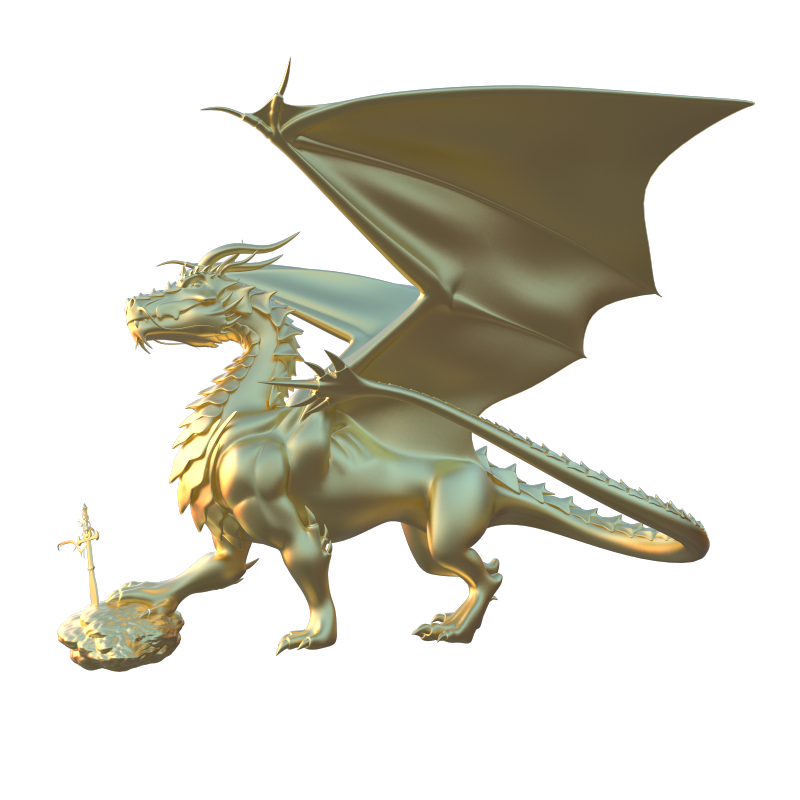}
 & 
\includegraphics[width=0.16\textwidth]{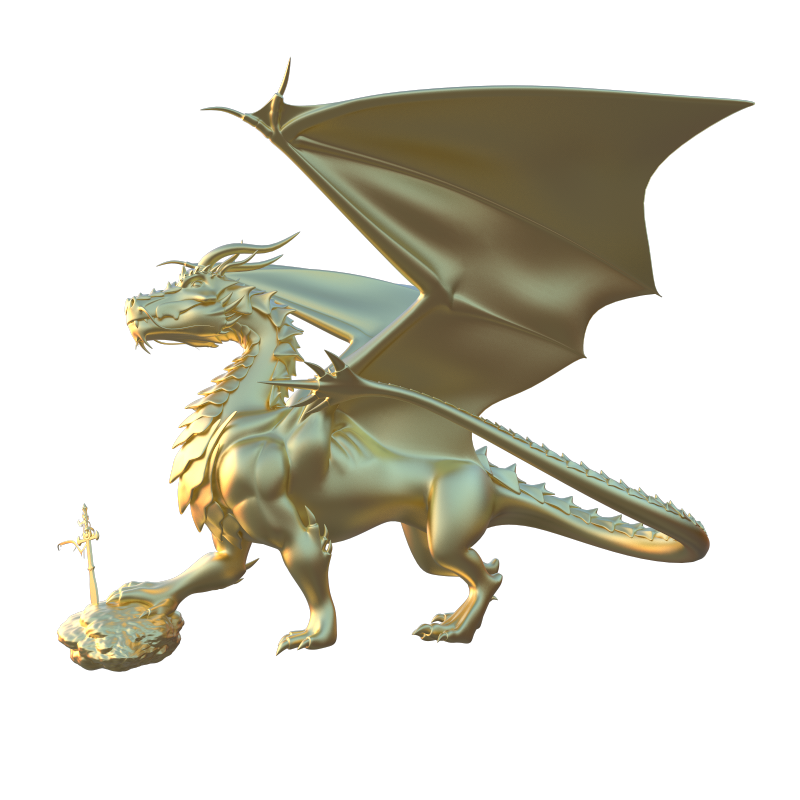}
 & 
\includegraphics[width=0.16\textwidth]{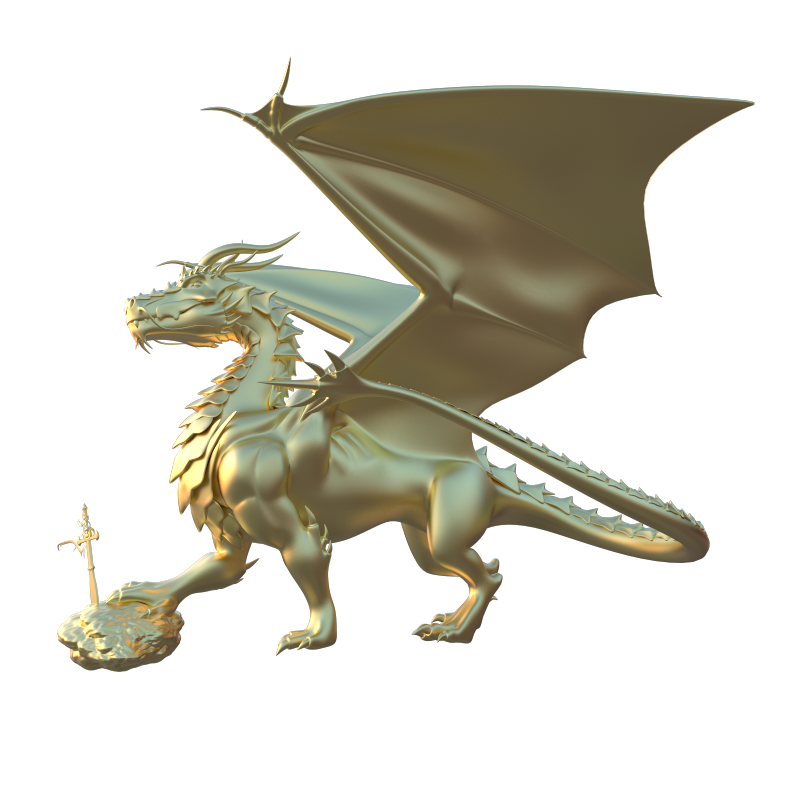}

\makebox[5pt]{\rotatebox{-90}{\hspace{-52pt}\footnotesize{Rendering}}}

 \\

{\footnotesize{MAPE: 0.2200}}
 & 
{\footnotesize{0.0515}}
 & 
{\footnotesize{0.0349}}
 & 
{\footnotesize{0.0263}}
 & 
{\footnotesize{\textbf{0.0232}}}
 &  \\

\includegraphics[width=0.16\textwidth]{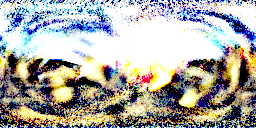}
 & 
\includegraphics[width=0.16\textwidth]{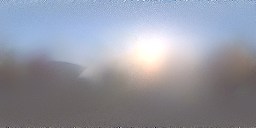}
 & 
\includegraphics[width=0.16\textwidth]{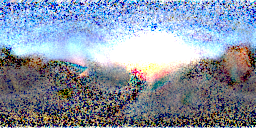}
 & 
\includegraphics[width=0.16\textwidth]{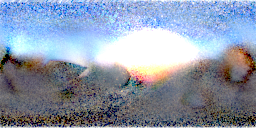}
 & 
\includegraphics[width=0.16\textwidth]{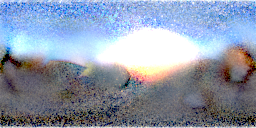}
 & 
\includegraphics[width=0.16\textwidth]{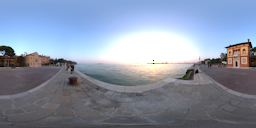}

\makebox[5pt]{\rotatebox{-90}{\hspace{-33pt}\footnotesize{Env. Map}}}

 \\

{\footnotesize{2.7400}}
 & 
{\footnotesize{\textbf{0.6320}}}
 & 
{\footnotesize{0.8250}}
 & 
{\footnotesize{0.6510}}
 & 
{\footnotesize{0.6490}}
 &  \\
\end{tabular}
\end{subfigure}

\endgroup

    \vspace{-0.3cm}
    \caption{\textbf{Ablation study} \mycaption{for environment map optimization reveals the effectiveness of each component and agrees with the ablation study for  material optimization (Figure \ref{fig:ablation-materials}).
    We start from the direct illumination integrator on the left, add the radiometric prior to it, stop propagating its gradient to the environment map, use the prior on the second bounce, and finally, enable the use of LHS reconstruction loss. Please note how omission of indirect effects in direct illumination causes material colors being baked into the environment map.}}
    \label{fig:ablation-envmap}
\end{figure*}

We now study how each component of our method contributes to the final result when optimizing material properties (\autoref{fig:ablation-materials}) and lighting (\autoref{fig:ablation-envmap}). We start from a direct illumination integrator, which recovers albedo/roughness and environment maps with baked-in artifacts that are due to ignoring indirect effects. Next, we add the radiometric prior and prevent its gradient from updating the scene parameters.
The next column demonstrates the improvement due to leveraging the radiometric prior also on the second bounce. Finally, adding the LHS reconstruction loss to train the radiance network yields further improved results.

\section{Limitations and Future Work}




{\em Specular materials.}
We do not attempt to model outgoing radiance on specular surfaces with the radiance MLP.
Instead, we keep tracing the path until it hits a non-specular surface. Our technique thus reverts to differentiating the path integral for specular chains and therefore suffers from the same limitations as AD-PT---large operation graph---if many surfaces in the scene are specular.

{\em Joint reconstruction of parameters.}
In this paper, we either optimized MLP representations of BRDF parameters or an environment map.
Extending our method to joint optimization of materials, geometry, and lighting would be an interesting avenue for future research.

\section{Conclusion}
In this paper, we proposed an inverse rendering method that leverages a radiometric prior---as a physical condition on the optimization---to account for global illumination. 
We evaluate the prior using a neural radiance field, which is trained during the optimization of scene parameters by minimizing the norm of the residual of the rendering equation.
Our method is theoretically grounded yet simple and practical.
We achieved the best results using neural representations, but the approach is compatible with standard data structures.
In contrast to prior work that requires tracing complete path samples, our approach recovers scene parameters with comparable (and sometimes better) quality. 
Introducing additional physics constraints on the optimization may improve performance further; we believe this is a direction worth pursuing in the future.



\begin{acks}
This material is based upon work supported by the National Science Foundation under Grant No. IIS2126407. We would also like to thank Aaron Lefohn for his support, and NVIDIA for funding the work with an NVIDIA academic partnership.
\end{acks}

\bibliographystyle{ACM-Reference-Format}
\bibliography{references}

\if 0\mode
    \appendix
    \section{Using MLP\lowercase{s} for scene parameters}

We show in Figure \ref{fig:grid_results} that representing SVBRDF parameters with MLPs yields reconstructions of higher quality than using dense grids. For this reason, although it requires the use of additional libraries to Mitsuba and is not compatible with mega kernels at this time, we still favor using MLPs.

\begin{figure}
    \centering
    \captionsetup[subfigure]{labelformat=empty}
    \begingroup
\renewcommand{\arraystretch}{0.6}
\setlength{\tabcolsep}{0.3em}

\begin{subfigure}{0.98\textwidth}
\begin{tabular}{cccccc}
& \multicolumn{2}{c}{\footnotesize{Albedo}} & & \multicolumn{2}{c}{\footnotesize{Roughness}}\\
& 
{\footnotesize{MLP}}
 & 
{\footnotesize{Grid}}
 & & 
{\footnotesize{MLP}}
 & 
{\footnotesize{Grid}}
\\

{\makebox[3pt]{\rotatebox{90}{\hspace{10pt} \footnotesize Ours}}}
 & 
\includegraphics[width=0.1\textwidth]{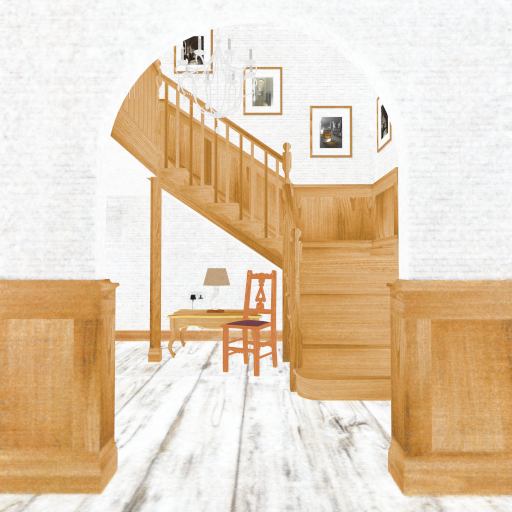}
 & 
\includegraphics[width=0.1\textwidth]{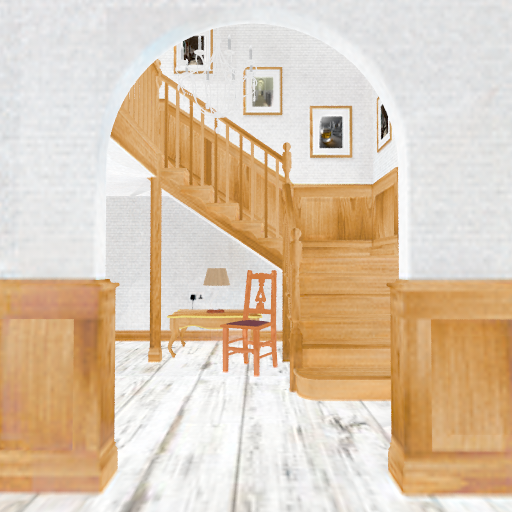}
 &  & 
\includegraphics[width=0.1\textwidth]{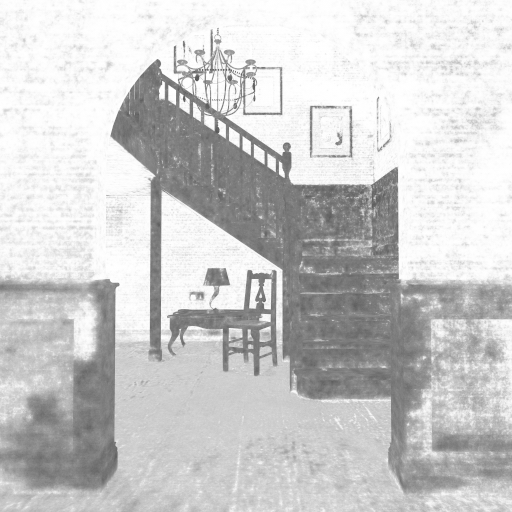}
 & 
\includegraphics[width=0.1\textwidth]{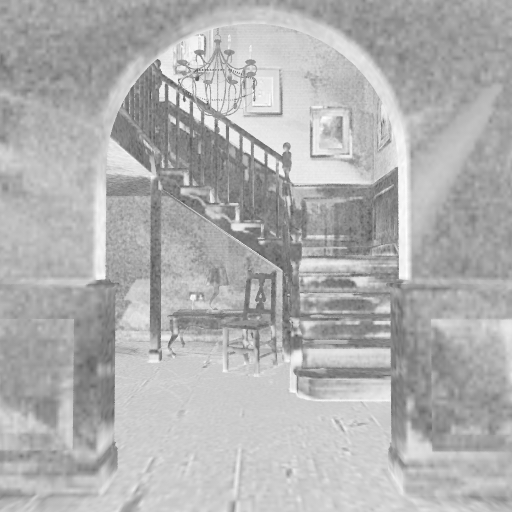}
 \\
  & 
 {\footnotesize{27.03}}
  & 
 {\footnotesize{19.19}}
  &  & 
 {\footnotesize{16.38}}
  & 
 {\footnotesize{8.92}}

\end{tabular}
\end{subfigure}

\begin{subfigure}{0.98\textwidth}
\begin{tabular}{cccccc}

{\makebox[3pt]{\rotatebox{90}{\hspace{10pt} \footnotesize PRB}}}
 & 
\includegraphics[width=0.1\textwidth]{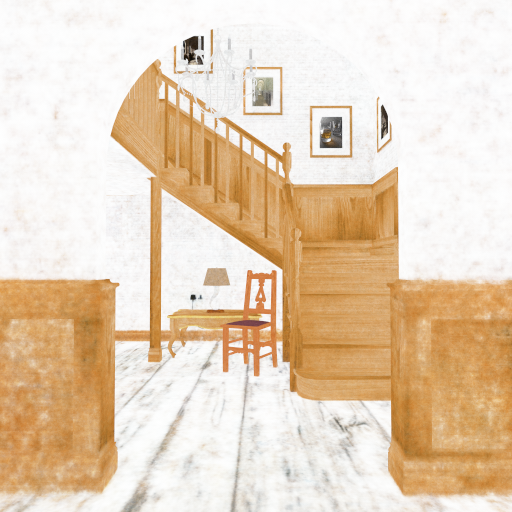}
 & 
\includegraphics[width=0.1\textwidth]{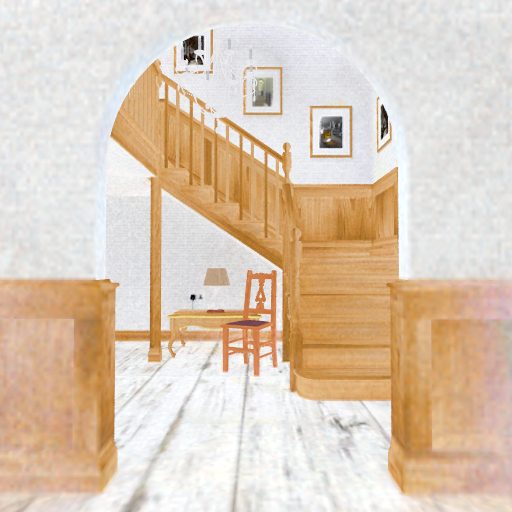}
 &  & 
\includegraphics[width=0.1\textwidth]{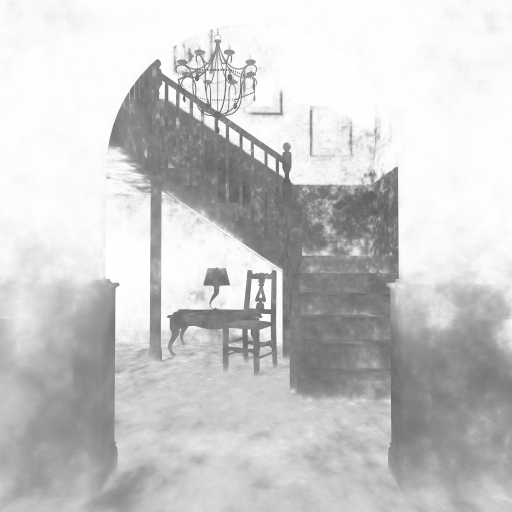}
 & 
\includegraphics[width=0.1\textwidth]{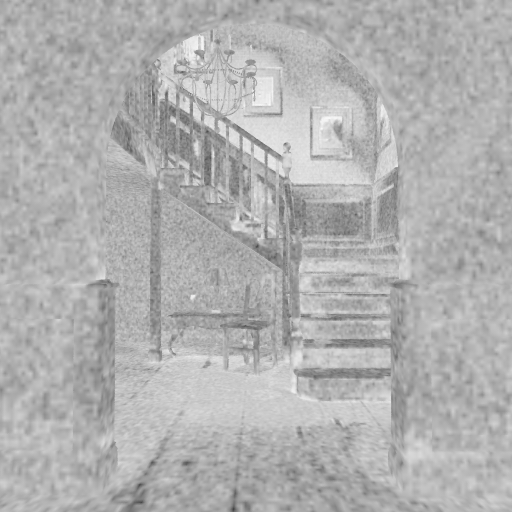}
 \\
  & 
 {\footnotesize{24.93}}
  & 
 {\footnotesize{18.40}}
  &  & 
 {\footnotesize{17.25}}
  & 
 {\footnotesize{8.66}}

\end{tabular}
\end{subfigure}

\endgroup

    \caption{
    \textbf{Grid SVBRDF.} \mycaption{We compare the results of using a dense grid with resolution $256^3$ to store the scene parameters versus using an MLP. MLP results are superior regardless of rendering method. PSNR is reported.}
    }
    \label{fig:grid_results}
\end{figure}

\begin{figure}
    \centering
    \includegraphics[width=0.23\textwidth]{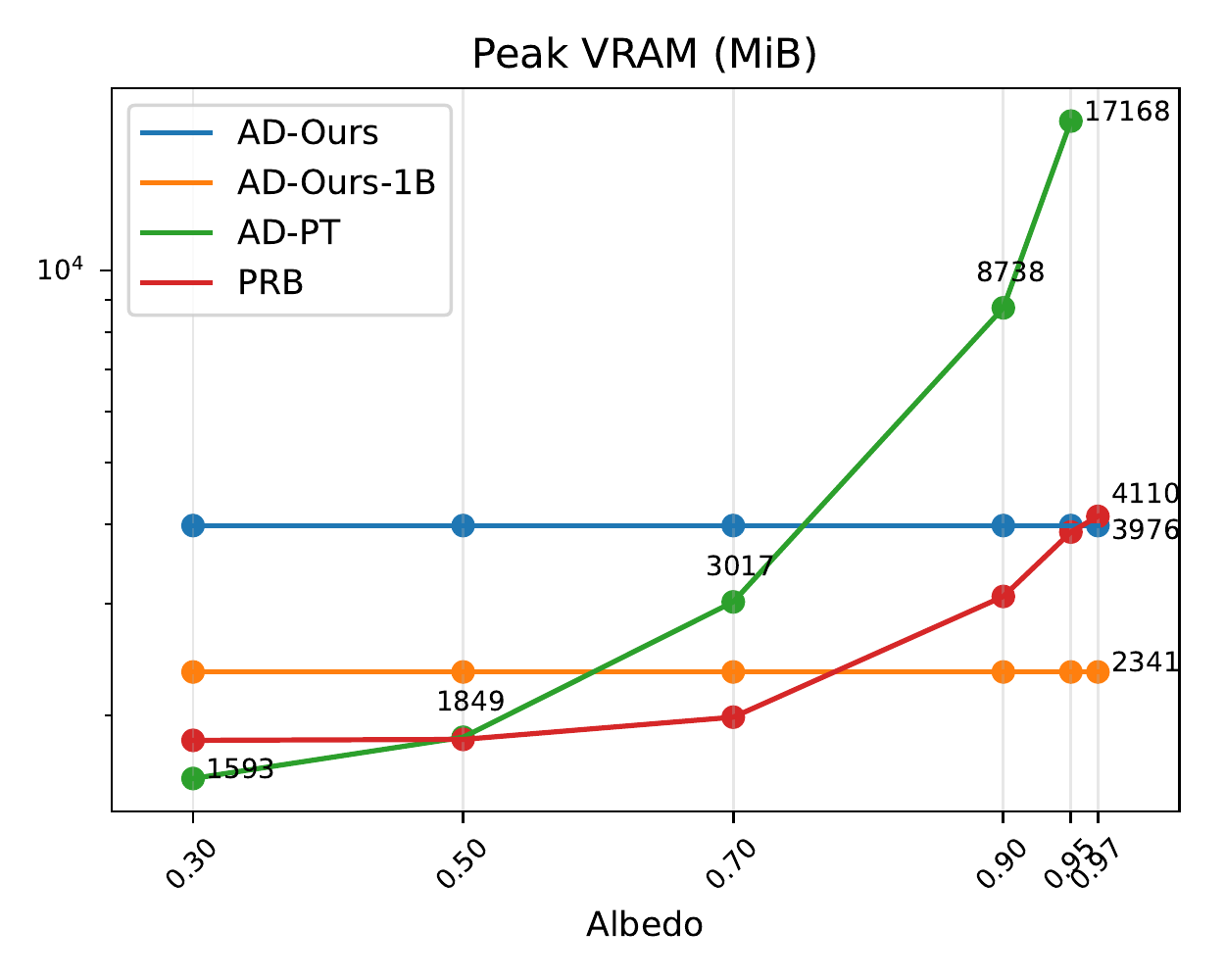}
    \includegraphics[width=0.23\textwidth]{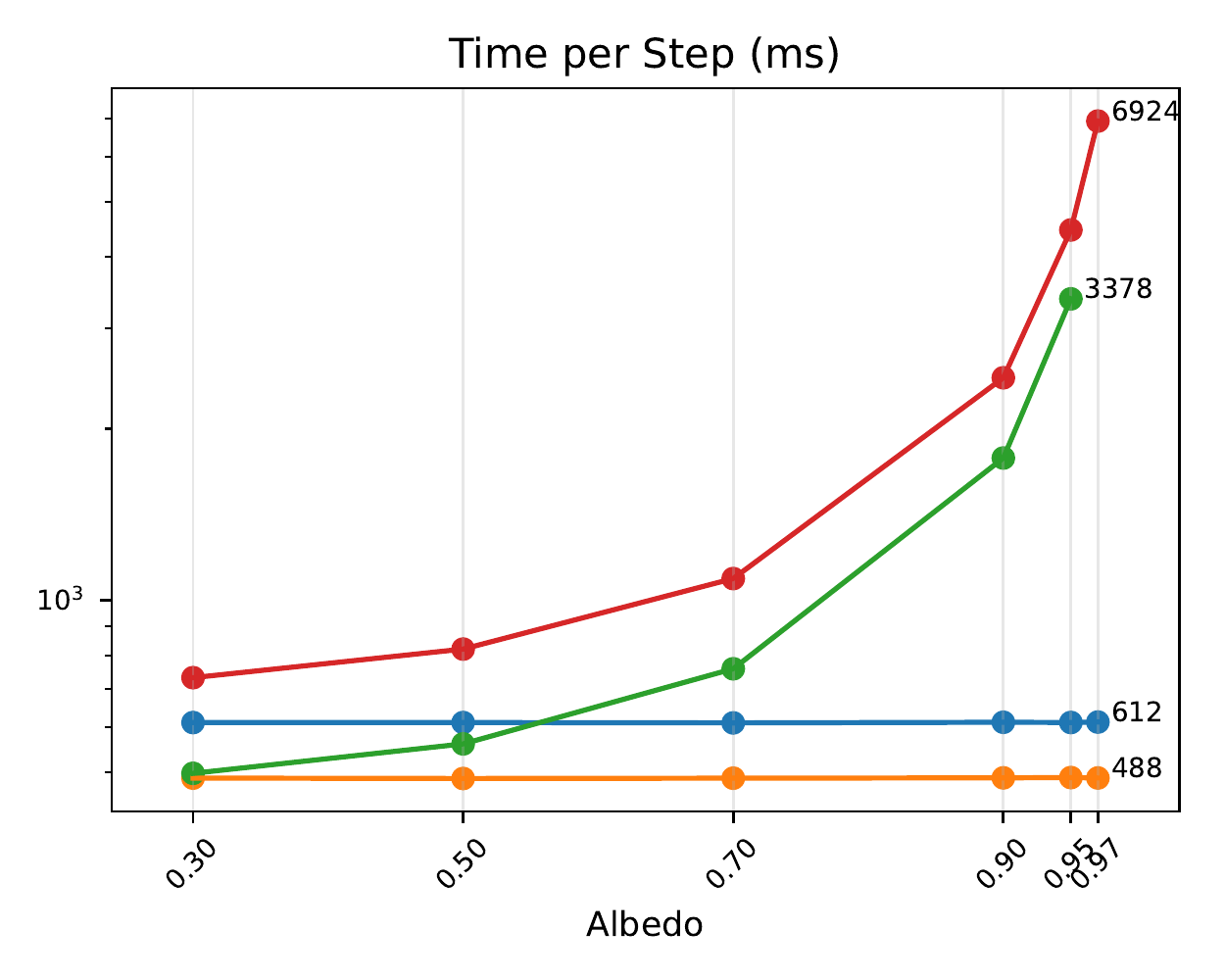}
    \caption{
        {\textbf{\textit{Cube} scene measurements.}} \mycaption{We compare the time and memory consumption of each training step, with all parameters represented as MLPs. Note that the Y-axis is in log scale. AD-PT runs out of VRAM (24GB) at albedo 0.97. Our method uses constant amounts of VRAM and time.}    
    }
    \label{fig:perf_mlp}
\end{figure}

\section{Additional Time and Memory Analysis}

As described in Section 6 of the paper, we measure the time and memory consumption of different methods as the path lengths change in a scene. To enable the comparison with mega kernels, as shown in the main paper, we used dense grids in Mitsuba as radiance and scene parameter representation. In contrast, here we present the results using MLPs in PyTorch in Figure \ref{fig:perf_mlp}. The VRAM consumption is the sum of the peak allocated memory reported by both Dr.JIT and PyTorch. While our method has a larger VRAM overhead due to additional radiance MLP queries, its time and memory usage remains constant as path length increases, while the costs of other methods grow rapidly.

In all experiments, we initialize the radiance grid values to the albedo of the walls and never update them during the measurement, i.e., the back-propagation and gradients are computed as usual but not applied to the grid values. This ensures that the measurements are from fixed albedos. We obtain the peak VRAM numbers from the Dr.JIT memory allocator. For methods that solve path integrals (AD-PT and PRB), we enable Russian-Roulette with a minimum termination probability of 0.05, and cap the maximum path length to the 99.9 percentile when the scene is rendered with path tracing.

\section{Additional Results}

We present results of additional scenes in Figure \ref{fig:results_nerf}.

\begin{figure*}
    \centering
    \captionsetup[subfigure]{labelformat=empty}
    \begingroup
\renewcommand{\arraystretch}{0.6}

\makebox[5pt]{\rotatebox{90}{\hspace{-10pt} \footnotesize{\VeachDoor}}}
\begin{subfigure}[b]{0.98\textwidth}
\begin{tabular}{ccccccccccc}

{\footnotesize{AD-Direct}}
 & 
{\footnotesize{PRB}}
 & 
{\footnotesize{\begin{tabular}{@{}c@{}}AD-Ours \\ w/o Prior\end{tabular}}}
 & 
{\footnotesize{AD-Ours}}
 & 
{\footnotesize{GT}}
 &  & 
{\footnotesize{AD-Direct}}
 & 
{\footnotesize{PRB}}
 & 
{\footnotesize{\begin{tabular}{@{}c@{}}AD-Ours \\ w/o Prior\end{tabular}}}
 & 
{\footnotesize{AD-Ours}}
 & 
{\footnotesize{GT}}
 \\

\includegraphics[width=0.077\textwidth]{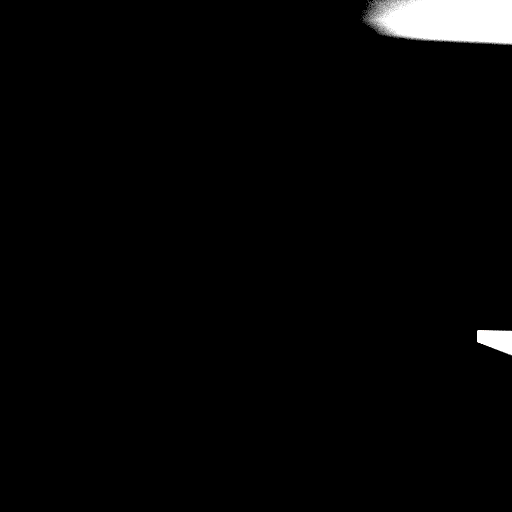}
 & 
\includegraphics[width=0.077\textwidth]{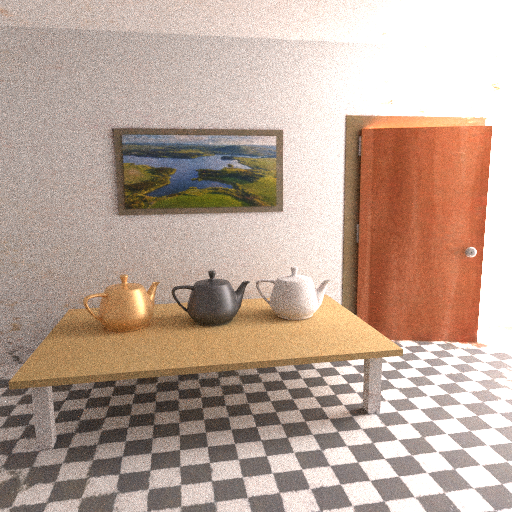}
 & 
\includegraphics[width=0.077\textwidth]{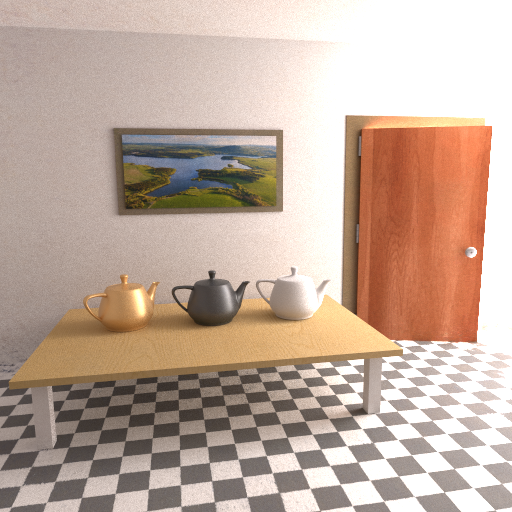}
 & 
\includegraphics[width=0.077\textwidth]{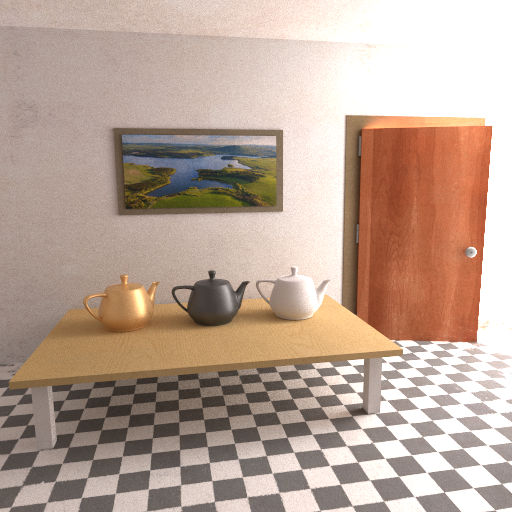}
 & 
\includegraphics[width=0.077\textwidth]{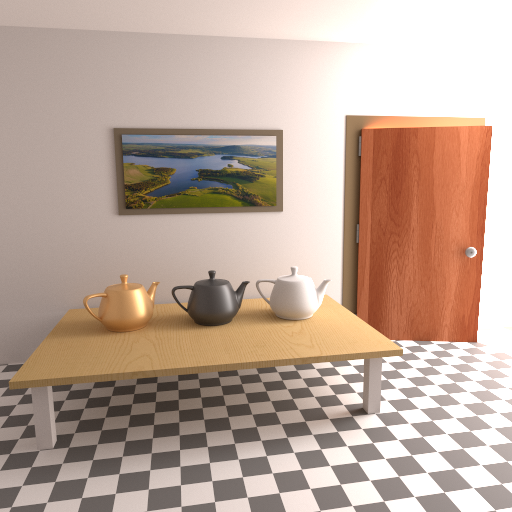}
 &  & 
\includegraphics[width=0.077\textwidth]{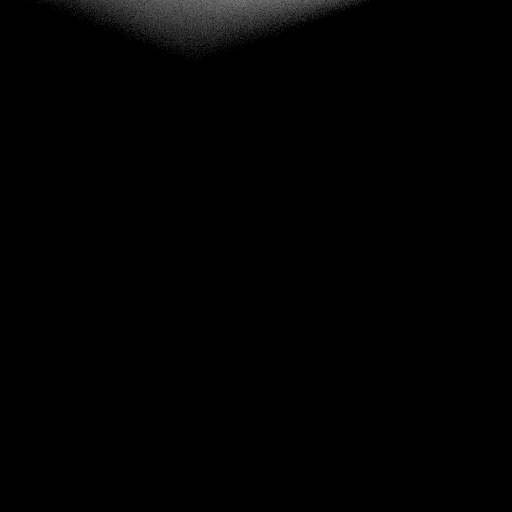}
 & 
\includegraphics[width=0.077\textwidth]{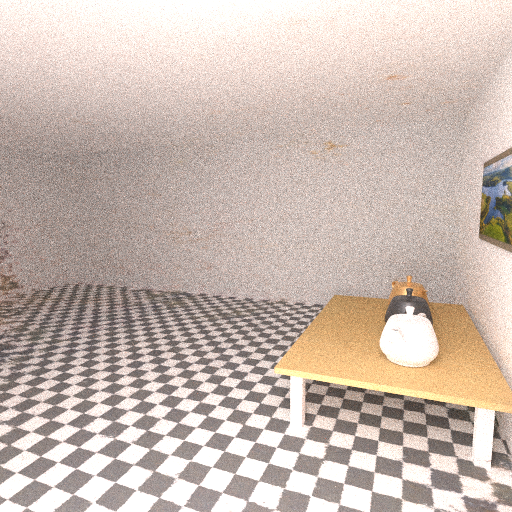}
 & 
\includegraphics[width=0.077\textwidth]{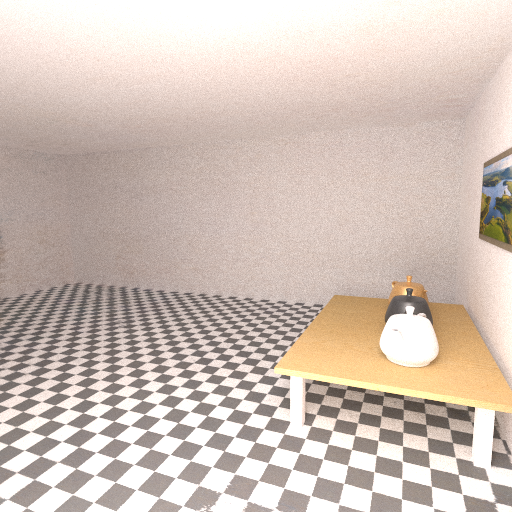}
 & 
\includegraphics[width=0.077\textwidth]{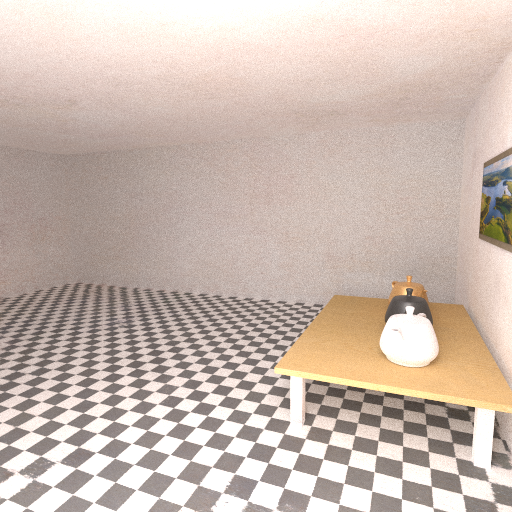}
 & 
\includegraphics[width=0.077\textwidth]{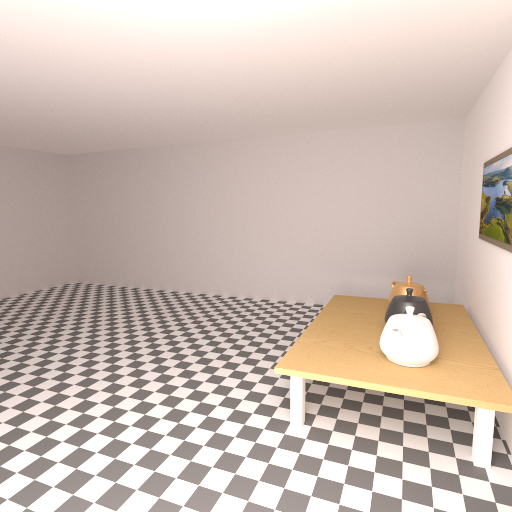}

\makebox[5pt]{\rotatebox{-90}{\hspace{-36pt}\footnotesize{Rendering}}}

 \\

{\footnotesize{-1.27}}
 & 
{\footnotesize{4.86}}
 & 
{\footnotesize{10.53}}
 & 
{\footnotesize{\textbf{12.01}}}
 &  &  & 
{\footnotesize{3.30}}
 & 
{\footnotesize{19.35}}
 & 
{\footnotesize{21.78}}
 & 
{\footnotesize{\textbf{24.81}}}
 &  \\

\includegraphics[width=0.077\textwidth]{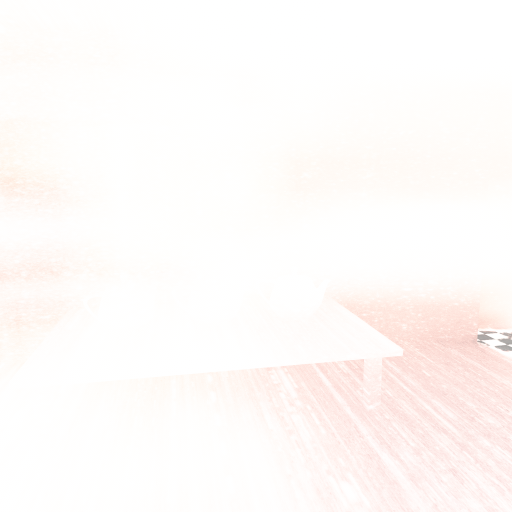}
 & 
\includegraphics[width=0.077\textwidth]{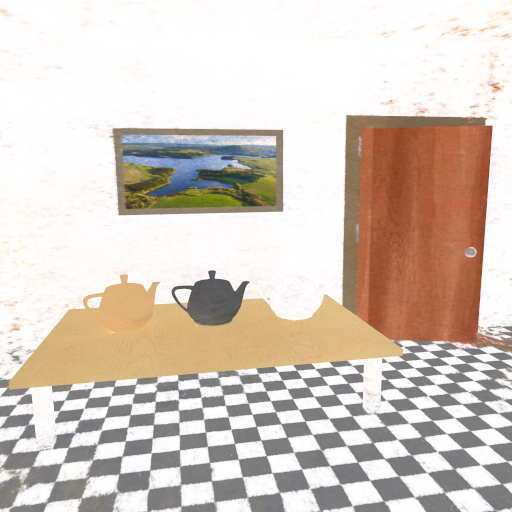}
 & 
\includegraphics[width=0.077\textwidth]{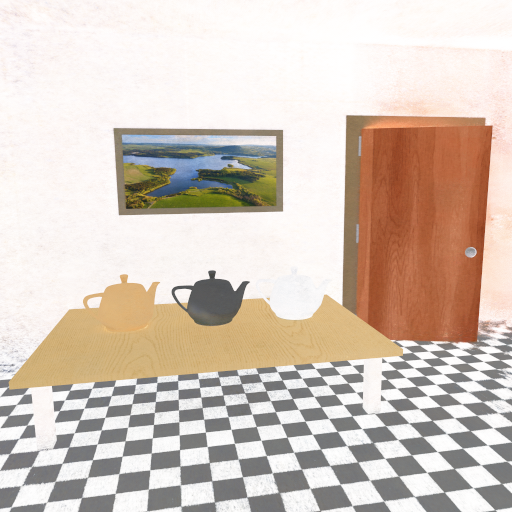}
 & 
\includegraphics[width=0.077\textwidth]{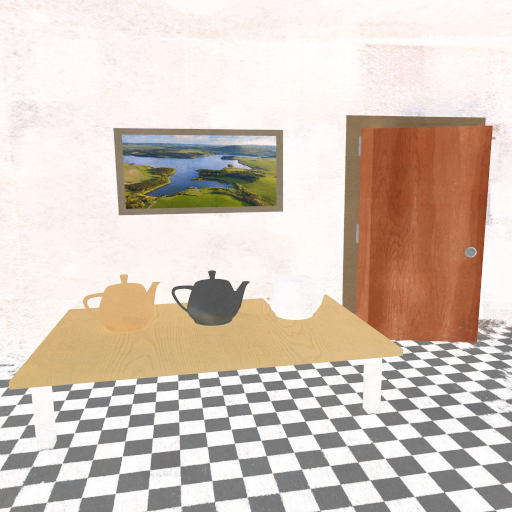}
 & 
\includegraphics[width=0.077\textwidth]{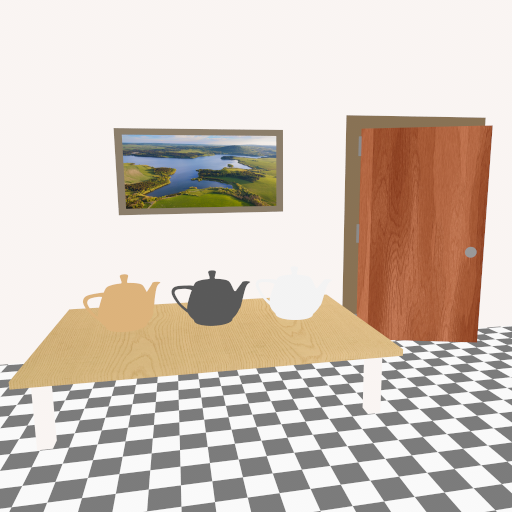}
 &  & 
\includegraphics[width=0.077\textwidth]{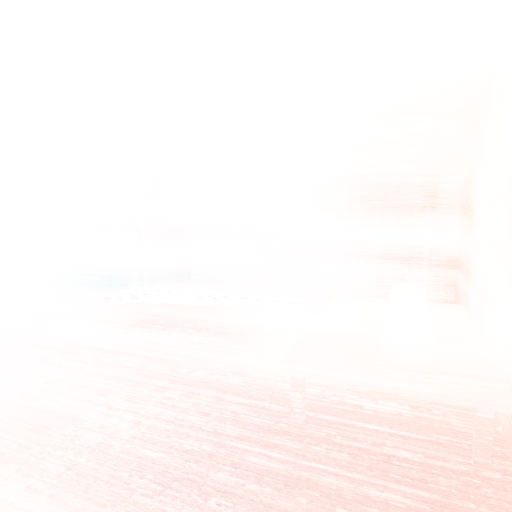}
 & 
\includegraphics[width=0.077\textwidth]{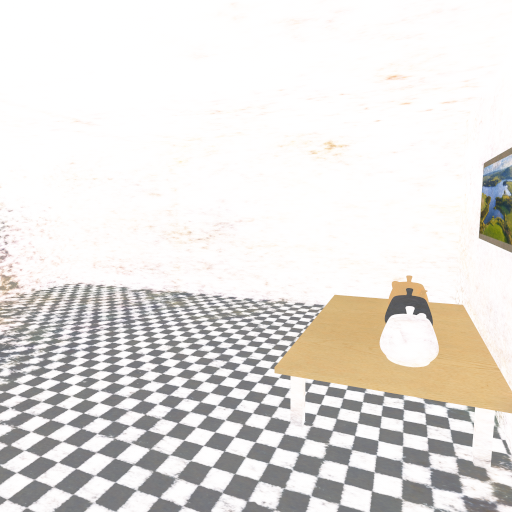}
 & 
\includegraphics[width=0.077\textwidth]{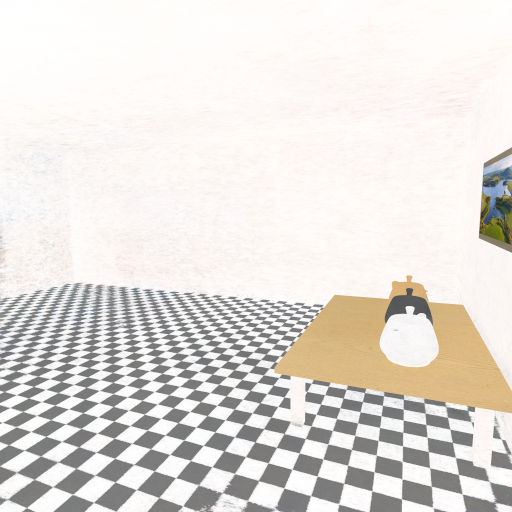}
 & 
\includegraphics[width=0.077\textwidth]{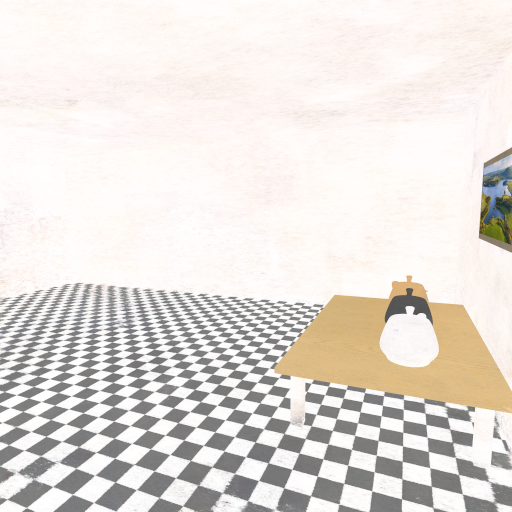}
 & 
\includegraphics[width=0.077\textwidth]{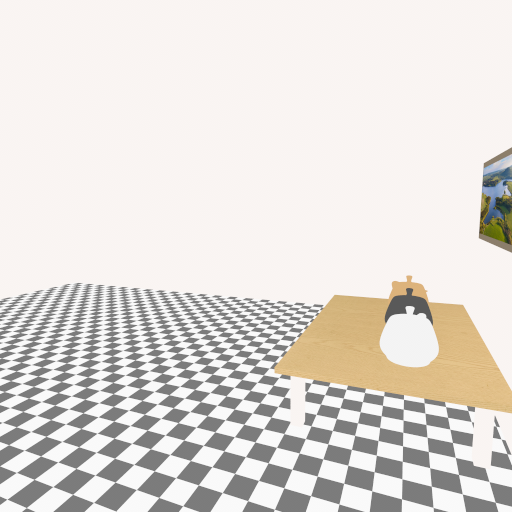}

\makebox[5pt]{\rotatebox{-90}{\hspace{-33pt}\footnotesize{Albedo}}}

 \\

{\footnotesize{6.59}}
 & 
{\footnotesize{22.54}}
 & 
{\footnotesize{25.13}}
 & 
{\footnotesize{\textbf{27.13}}}
 &  &  & 
{\footnotesize{9.42}}
 & 
{\footnotesize{21.25}}
 & 
{\footnotesize{26.60}}
 & 
{\footnotesize{\textbf{26.71}}}
 &  \\

\includegraphics[width=0.077\textwidth]{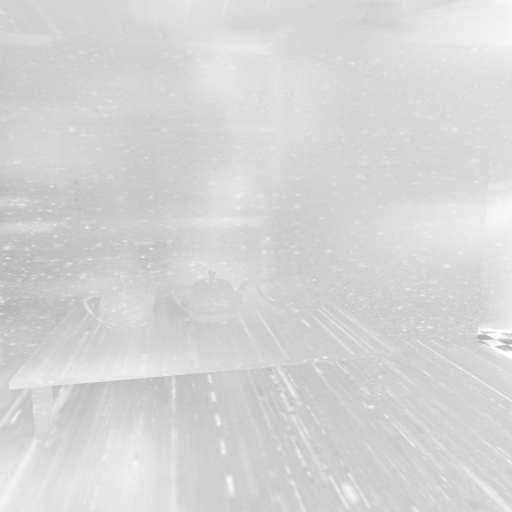}
 & 
\includegraphics[width=0.077\textwidth]{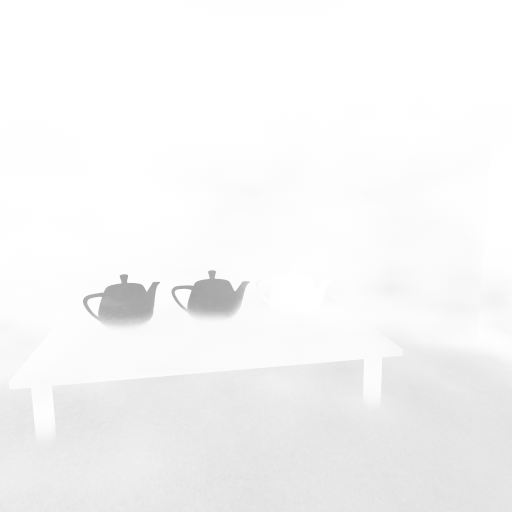}
 & 
\includegraphics[width=0.077\textwidth]{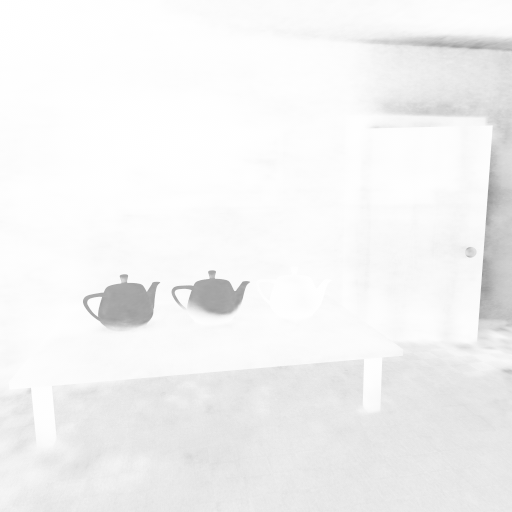}
 & 
\includegraphics[width=0.077\textwidth]{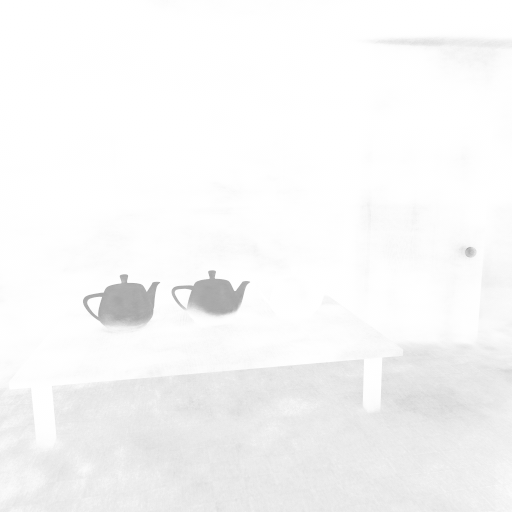}
 & 
\includegraphics[width=0.077\textwidth]{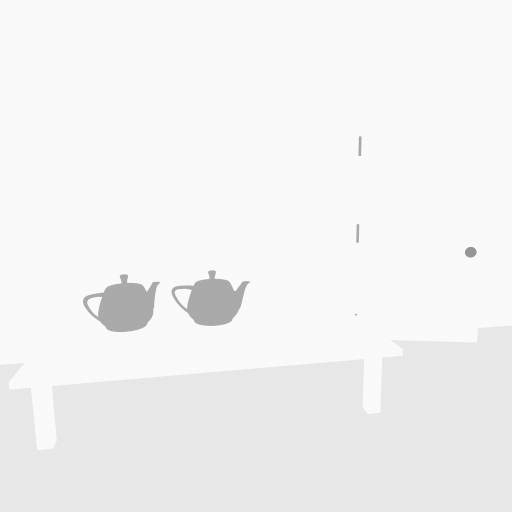}
 &  & 
\includegraphics[width=0.077\textwidth]{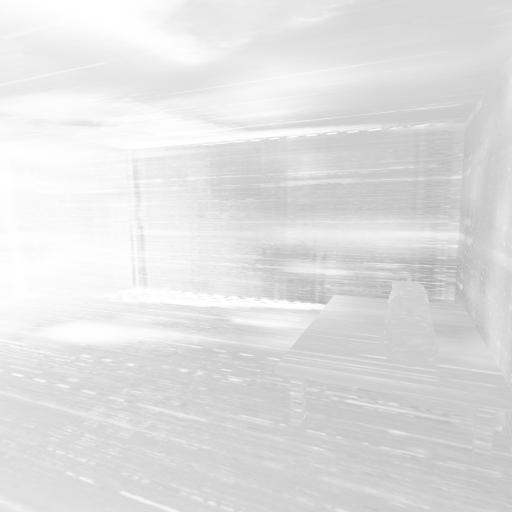}
 & 
\includegraphics[width=0.077\textwidth]{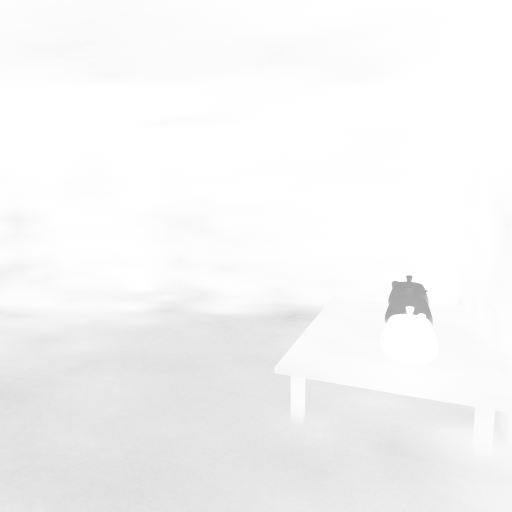}
 & 
\includegraphics[width=0.077\textwidth]{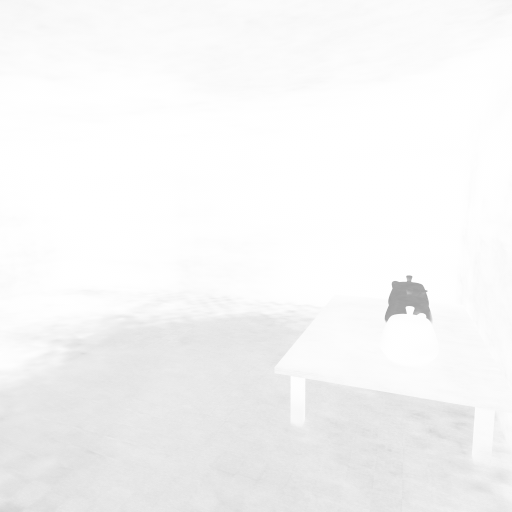}
 & 
\includegraphics[width=0.077\textwidth]{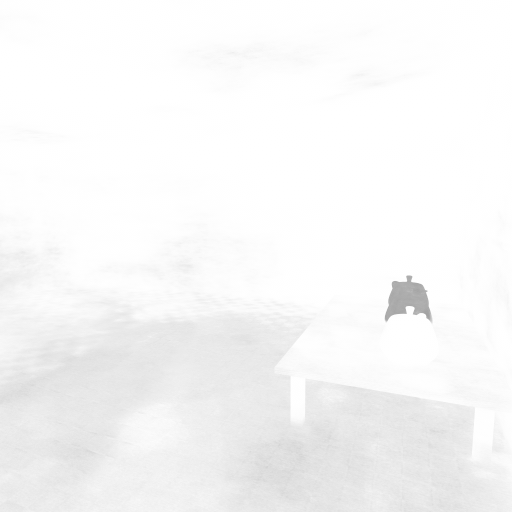}
 & 
\includegraphics[width=0.077\textwidth]{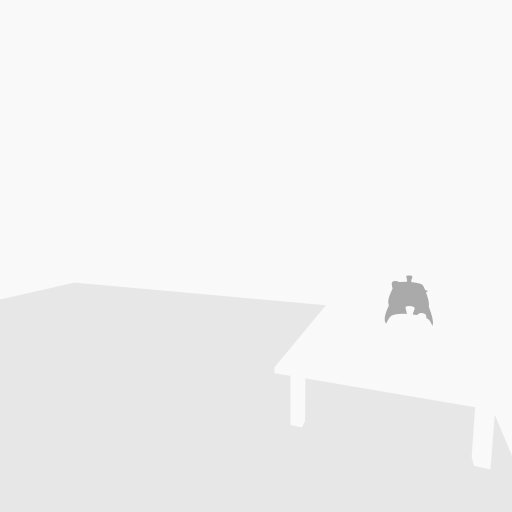}

\makebox[5pt]{\rotatebox{-90}{\hspace{-36pt}\footnotesize{Roughness}}}

 \\

{\footnotesize{12.02}}
 & 
{\footnotesize{\textbf{26.30}}}
 & 
{\footnotesize{21.57}}
 & 
{\footnotesize{25.48}}
 &  &  & 
{\footnotesize{14.06}}
 & 
{\footnotesize{\textbf{27.43}}}
 & 
{\footnotesize{26.70}}
 & 
{\footnotesize{25.90}}
 &  \\
\midrule
\end{tabular}
\end{subfigure}

\makebox[5pt]{\rotatebox{90}{\hspace{-10pt} \footnotesize{\Hotdog}}}
\begin{subfigure}[b]{0.98\textwidth}
\begin{tabular}{ccccccccccc}

\includegraphics[trim={50 50 50 50},clip,width=0.077\textwidth]{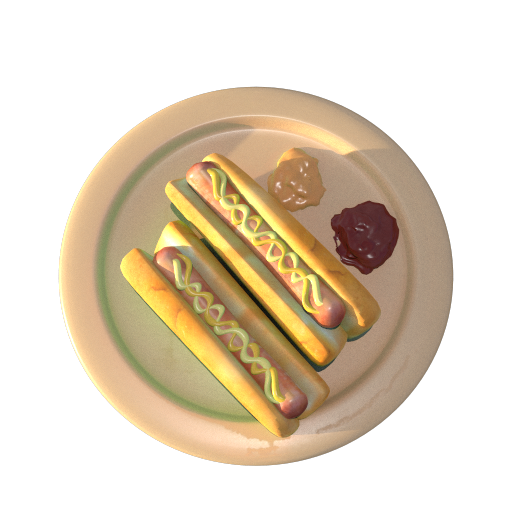}
 & 
\includegraphics[trim={50 50 50 50},clip,width=0.077\textwidth]{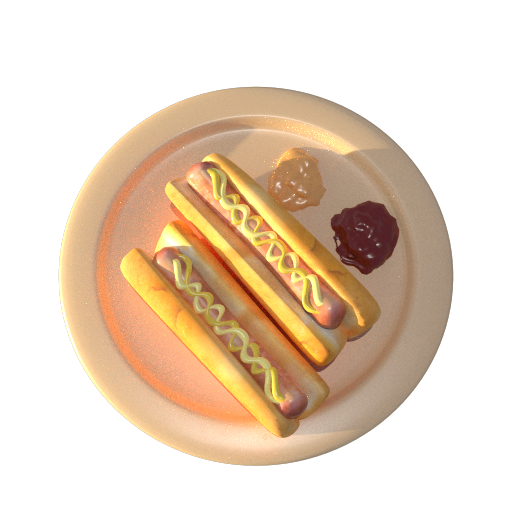}
 & 
\includegraphics[trim={50 50 50 50},clip,width=0.077\textwidth]{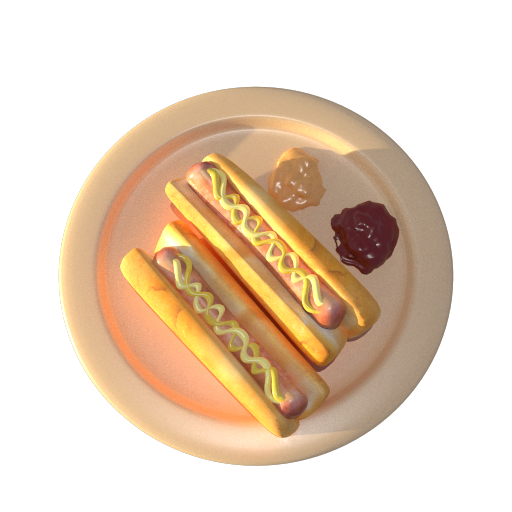}
 & 
\includegraphics[trim={50 50 50 50},clip,width=0.077\textwidth]{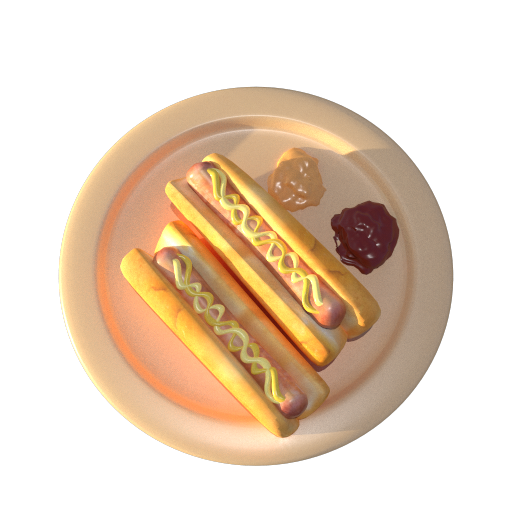}
 & 
\includegraphics[trim={50 50 50 50},clip,width=0.077\textwidth]{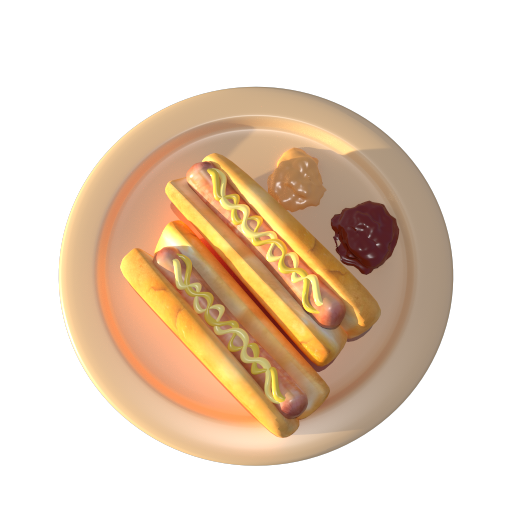}
 &  & 
\includegraphics[trim={50 50 50 50},clip,width=0.077\textwidth]{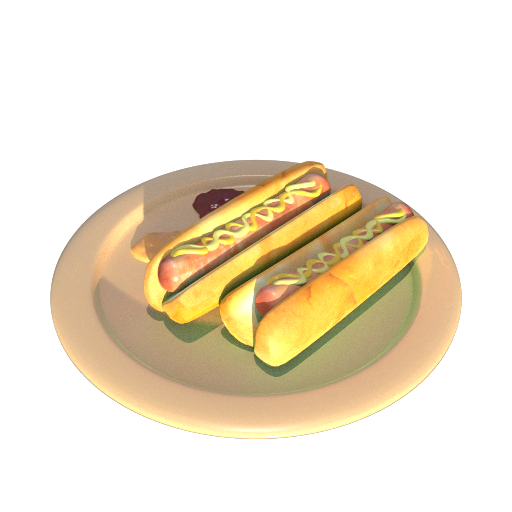}
 & 
\includegraphics[trim={50 50 50 50},clip,width=0.077\textwidth]{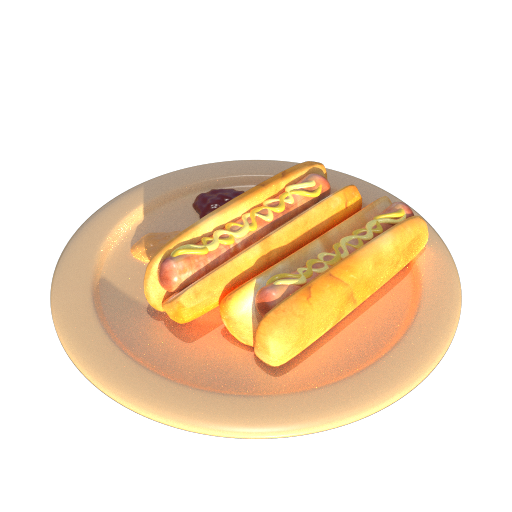}
 & 
\includegraphics[trim={50 50 50 50},clip,width=0.077\textwidth]{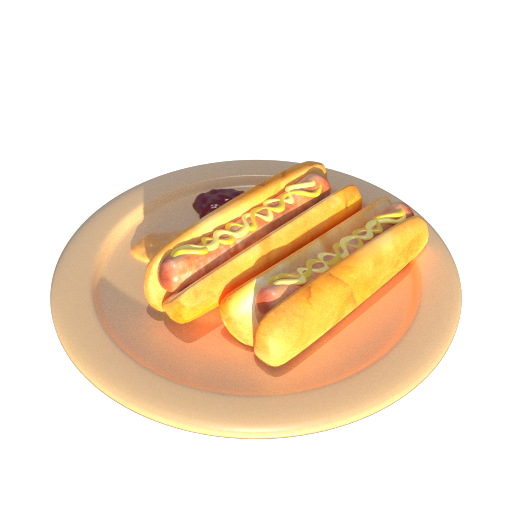}
 & 
\includegraphics[trim={50 50 50 50},clip,width=0.077\textwidth]{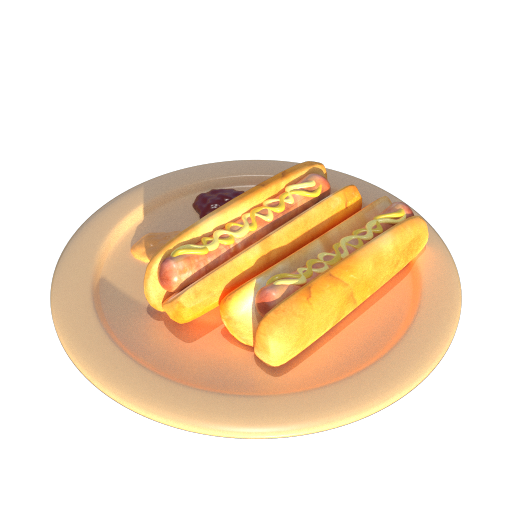}
 & 
\includegraphics[trim={50 50 50 50},clip,width=0.077\textwidth]{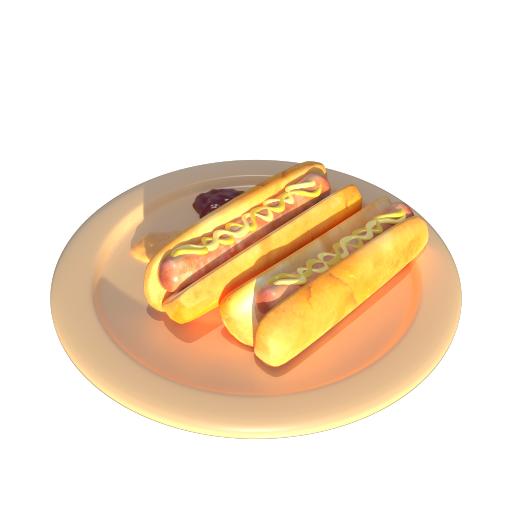}

\makebox[5pt]{\rotatebox{-90}{\hspace{-36pt}\footnotesize{Rendering}}}

 \\

{\footnotesize{22.08}}
 & 
{\footnotesize{31.71}}
 & 
{\footnotesize{36.25}}
 & 
{\footnotesize{\textbf{36.37}}}
 &  &  & 
{\footnotesize{21.11}}
 & 
{\footnotesize{29.56}}
 & 
{\footnotesize{33.20}}
 & 
{\footnotesize{\textbf{34.03}}}
 &  \\

\includegraphics[trim={50 50 50 50},clip,width=0.077\textwidth]{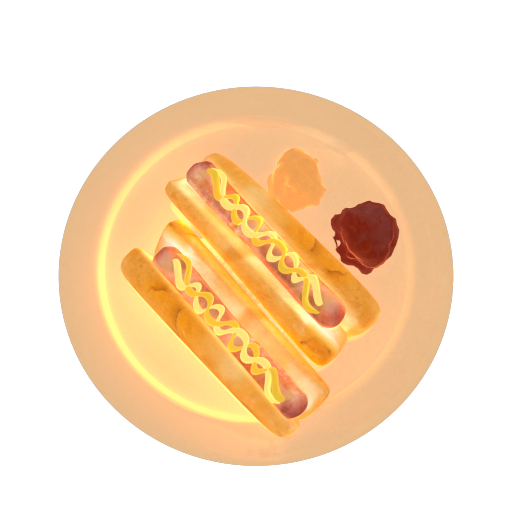}
 & 
\includegraphics[trim={50 50 50 50},clip,width=0.077\textwidth]{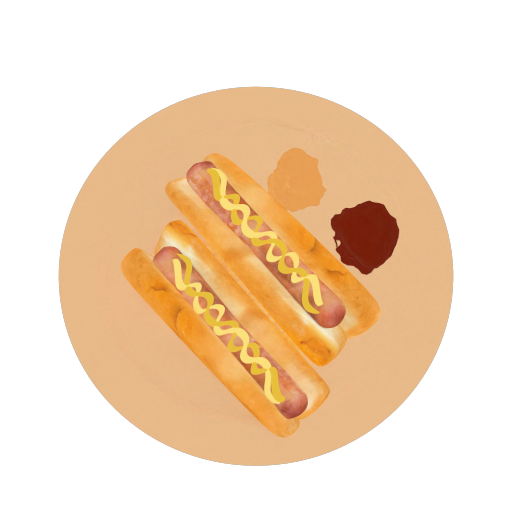}
 & 
\includegraphics[trim={50 50 50 50},clip,width=0.077\textwidth]{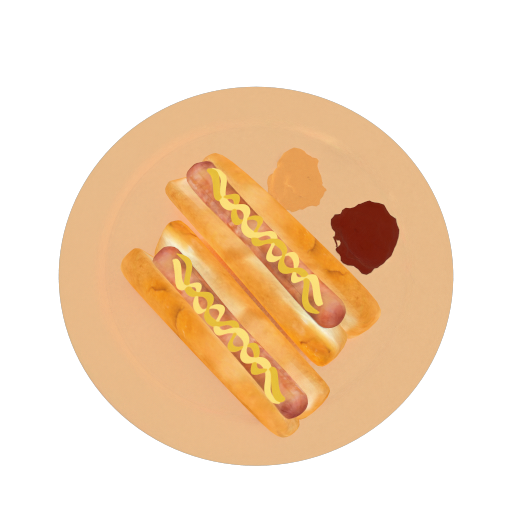}
 & 
\includegraphics[trim={50 50 50 50},clip,width=0.077\textwidth]{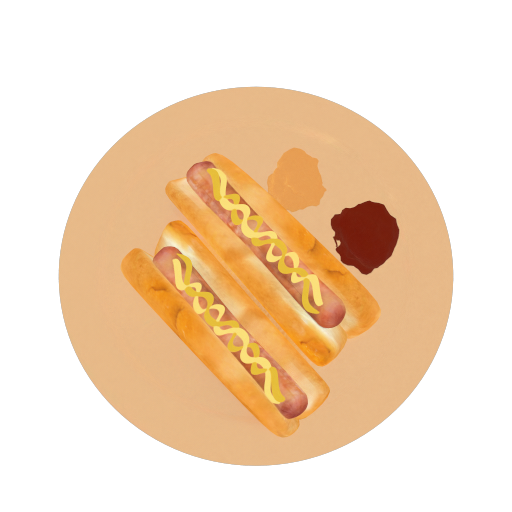}
 & 
\includegraphics[trim={50 50 50 50},clip,width=0.077\textwidth]{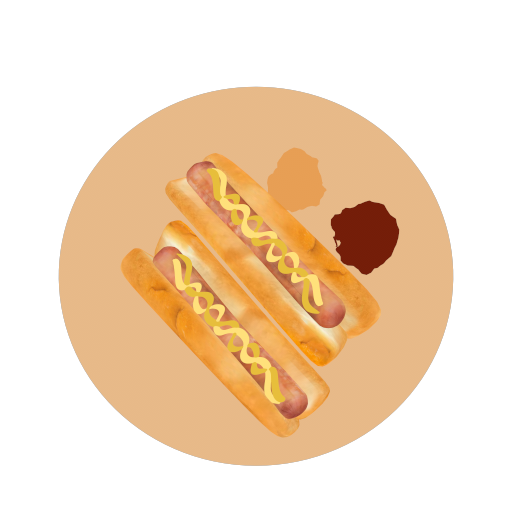}
 &  & 
\includegraphics[trim={50 50 50 50},clip,width=0.077\textwidth]{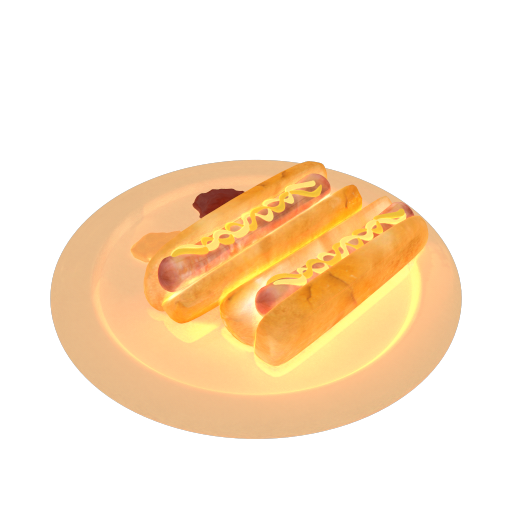}
 & 
\includegraphics[trim={50 50 50 50},clip,width=0.077\textwidth]{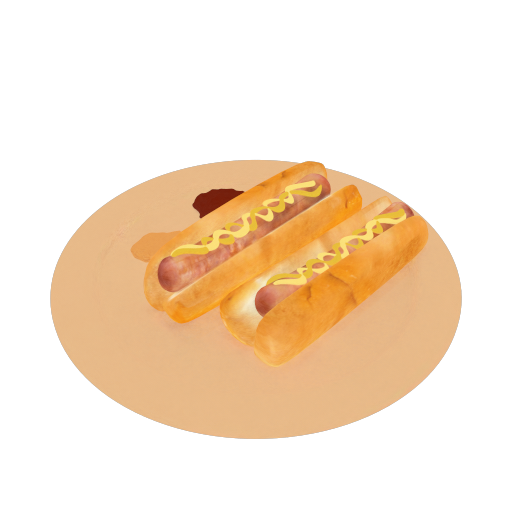}
 & 
\includegraphics[trim={50 50 50 50},clip,width=0.077\textwidth]{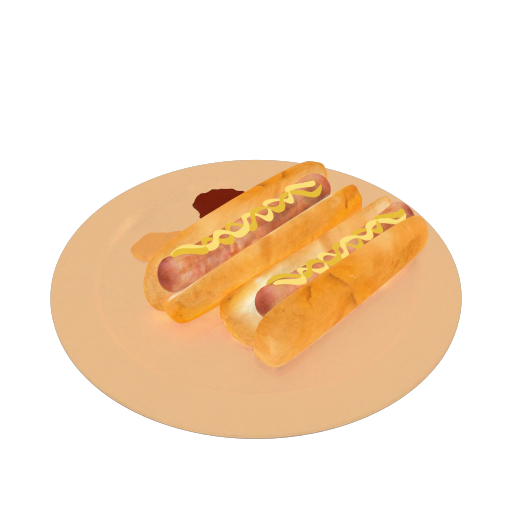}
 & 
\includegraphics[trim={50 50 50 50},clip,width=0.077\textwidth]{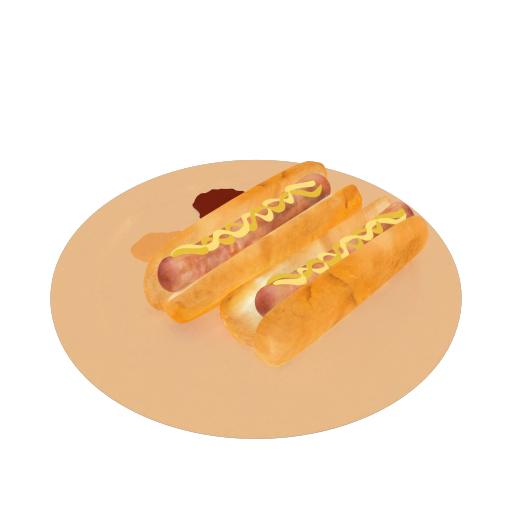}
 & 
\includegraphics[trim={50 50 50 50},clip,width=0.077\textwidth]{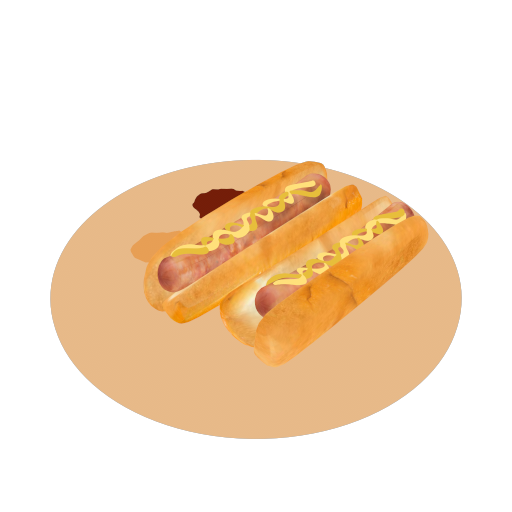}

\makebox[5pt]{\rotatebox{-90}{\hspace{-33pt}\footnotesize{Albedo}}}

 \\

{\footnotesize{22.81}}
 & 
{\footnotesize{41.23}}
 & 
{\footnotesize{37.22}}
 & 
{\footnotesize{\textbf{41.71}}}
 &  &  & 
{\footnotesize{23.59}}
 & 
{\footnotesize{41.14}}
 & 
{\footnotesize{37.41}}
 & 
{\footnotesize{\textbf{41.35}}}
 &  \\

\includegraphics[trim={50 50 50 50},clip,width=0.077\textwidth]{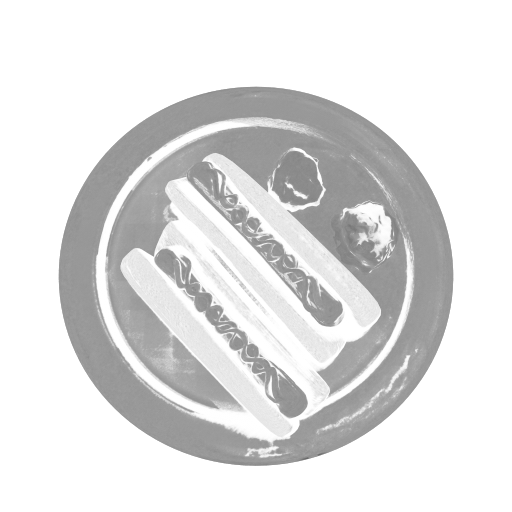}
 & 
\includegraphics[trim={50 50 50 50},clip,width=0.077\textwidth]{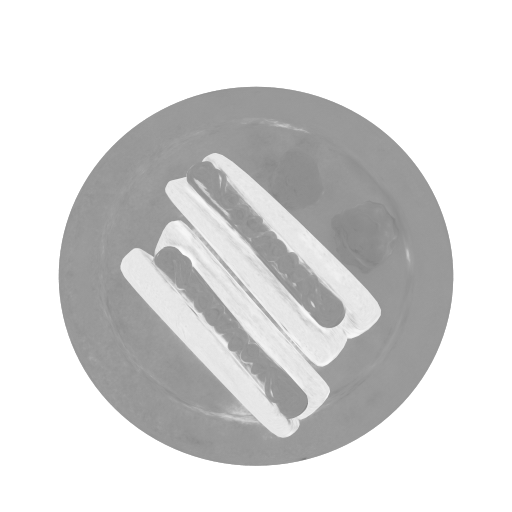}
 & 
\includegraphics[trim={50 50 50 50},clip,width=0.077\textwidth]{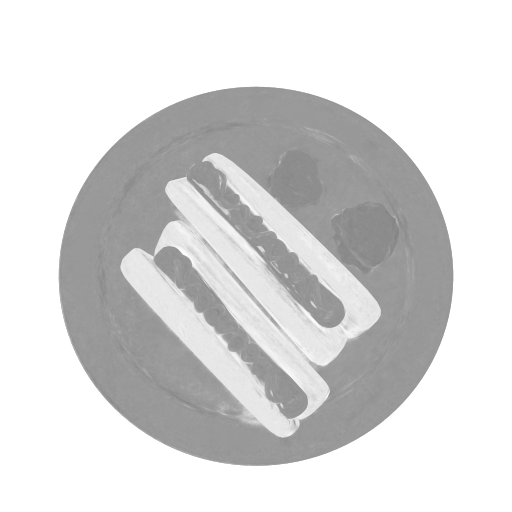}
 & 
\includegraphics[trim={50 50 50 50},clip,width=0.077\textwidth]{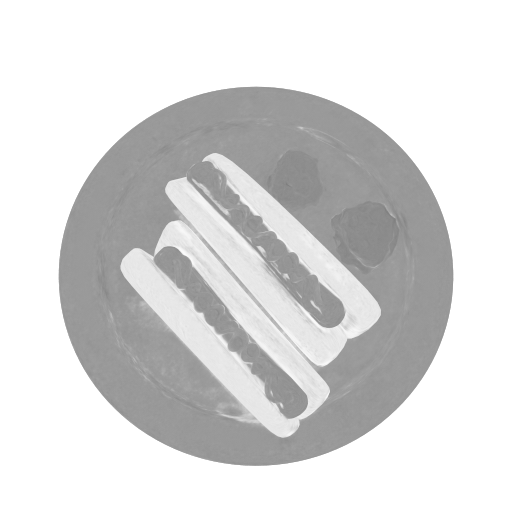}
 & 
\includegraphics[trim={50 50 50 50},clip,width=0.077\textwidth]{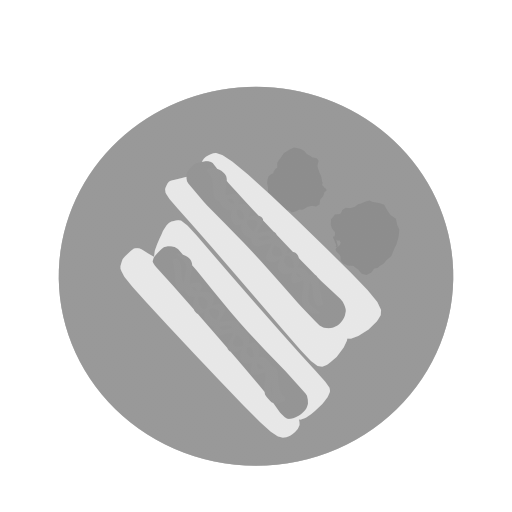}
 &  & 
\includegraphics[trim={50 50 50 50},clip,width=0.077\textwidth]{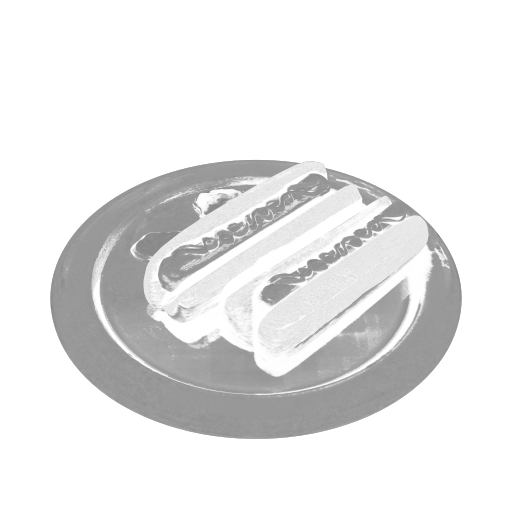}
 & 
\includegraphics[trim={50 50 50 50},clip,width=0.077\textwidth]{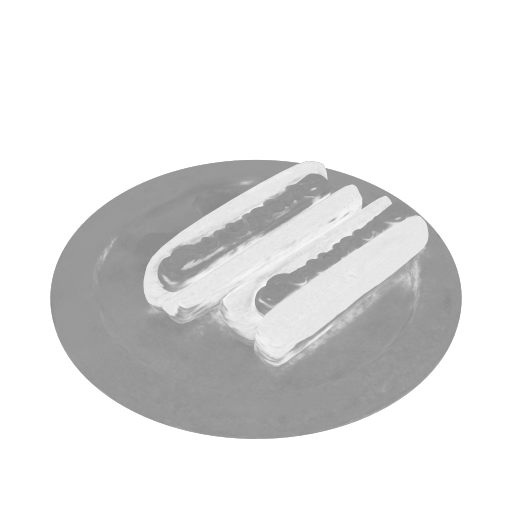}
 & 
\includegraphics[trim={50 50 50 50},clip,width=0.077\textwidth]{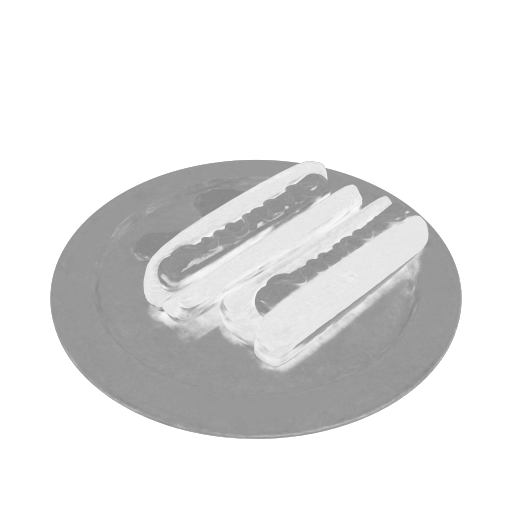}
 & 
\includegraphics[trim={50 50 50 50},clip,width=0.077\textwidth]{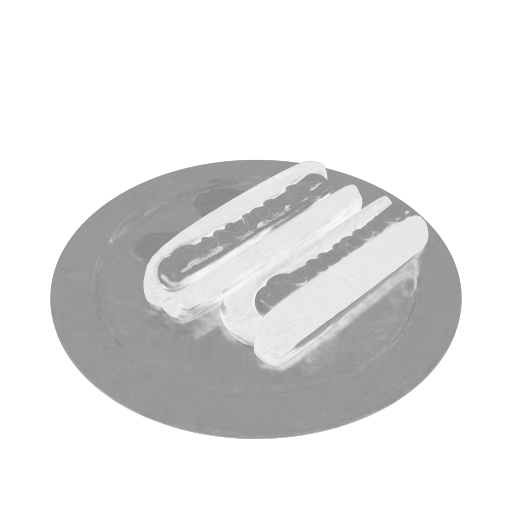}
 & 
\includegraphics[trim={50 50 50 50},clip,width=0.077\textwidth]{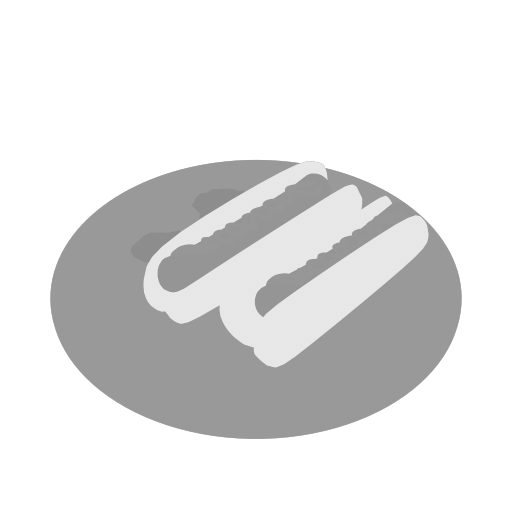}

\makebox[5pt]{\rotatebox{-90}{\hspace{-36pt}\footnotesize{Roughness}}}

 \\

{\footnotesize{16.47}}
 & 
{\footnotesize{\textbf{30.03}}}
 & 
{\footnotesize{28.15}}
 & 
{\footnotesize{28.13}}
 &  &  & 
{\footnotesize{18.00}}
 & 
{\footnotesize{\textbf{27.31}}}
 & 
{\footnotesize{25.43}}
 & 
{\footnotesize{26.28}}
 &  \\
\midrule
\end{tabular}
\end{subfigure}

\makebox[5pt]{\rotatebox{90}{\hspace{-10pt} \footnotesize{\Ficus}}}
\begin{subfigure}[b]{0.98\textwidth}
\begin{tabular}{ccccccccccc}

\includegraphics[trim={50 50 50 50},clip,width=0.077\textwidth]{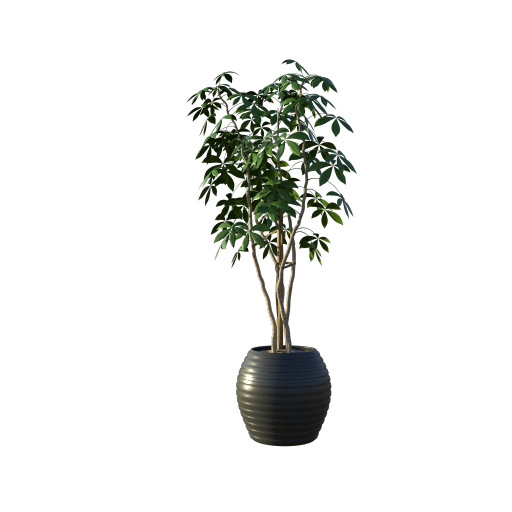}
 & 
\includegraphics[trim={50 50 50 50},clip,width=0.077\textwidth]{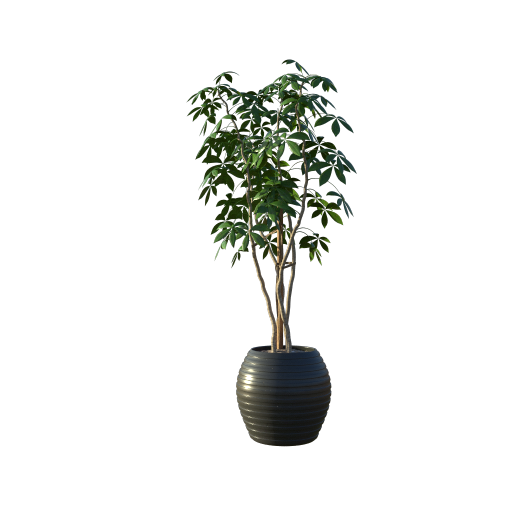}
 & 
\includegraphics[trim={50 50 50 50},clip,width=0.077\textwidth]{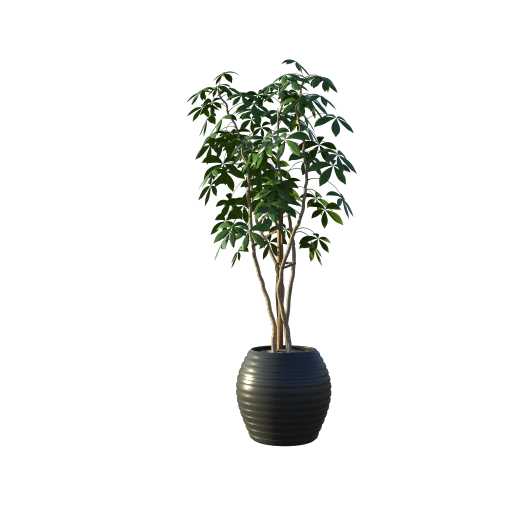}
 & 
\includegraphics[trim={50 50 50 50},clip,width=0.077\textwidth]{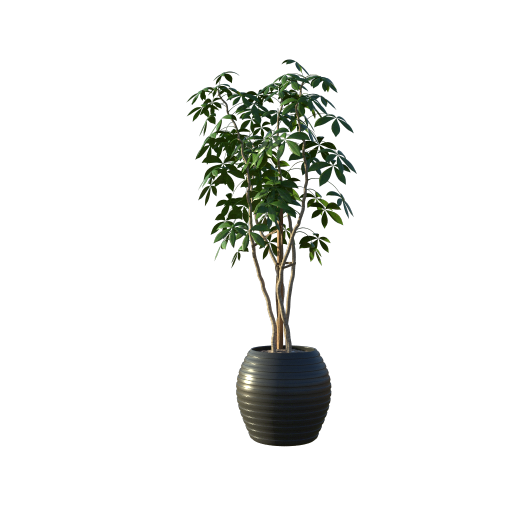}
 & 
\includegraphics[trim={50 50 50 50},clip,width=0.077\textwidth]{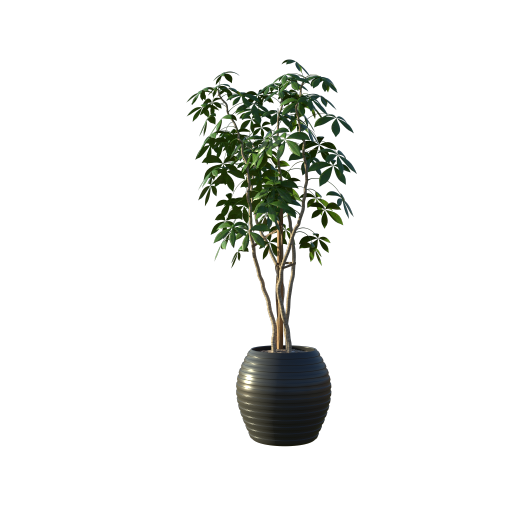}
 &  & 
\includegraphics[trim={50 50 50 50},clip,width=0.077\textwidth]{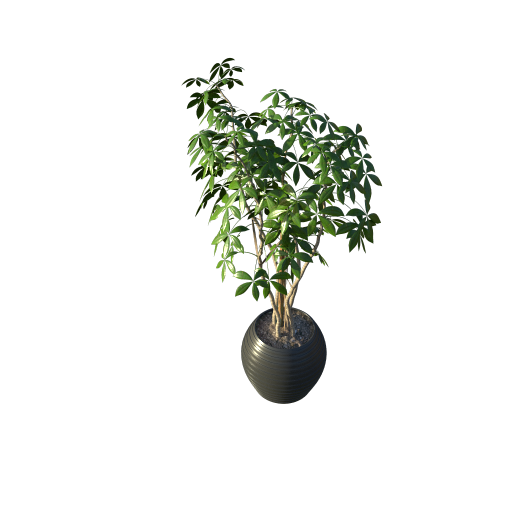}
 & 
\includegraphics[trim={50 50 50 50},clip,width=0.077\textwidth]{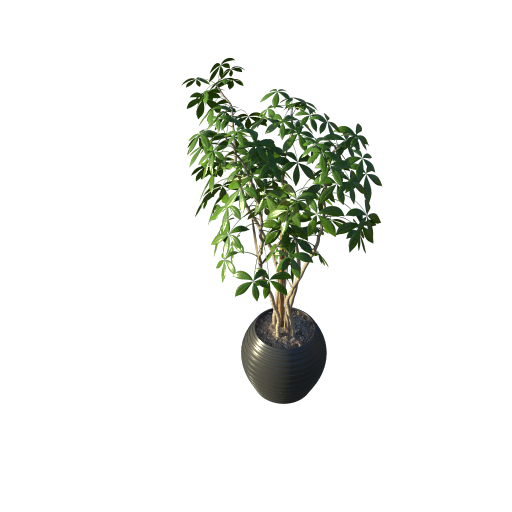}
 & 
\includegraphics[trim={50 50 50 50},clip,width=0.077\textwidth]{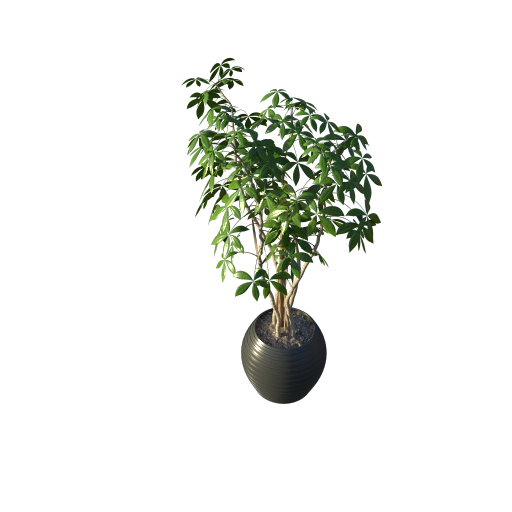}
 & 
\includegraphics[trim={50 50 50 50},clip,width=0.077\textwidth]{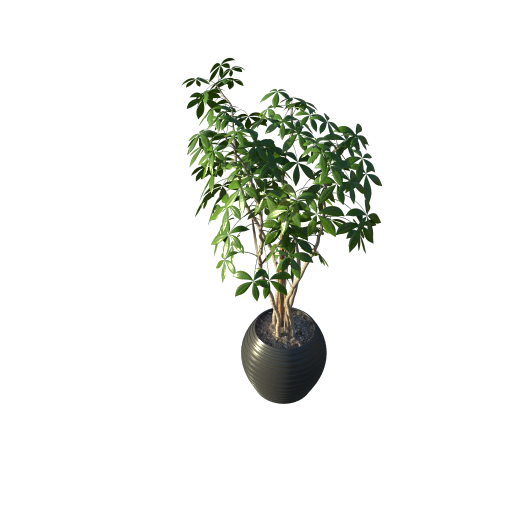}
 & 
\includegraphics[trim={50 50 50 50},clip,width=0.077\textwidth]{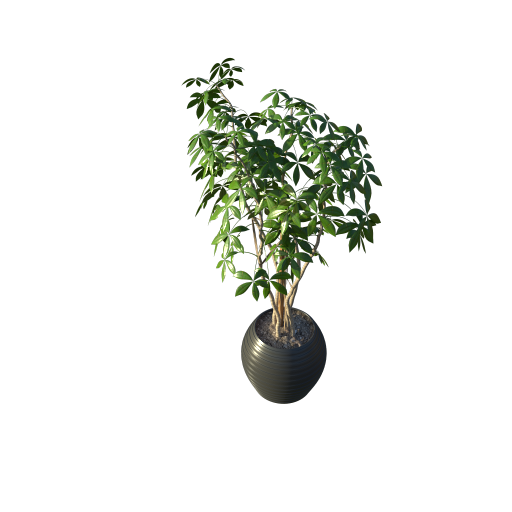}

\makebox[5pt]{\rotatebox{-90}{\hspace{-36pt}\footnotesize{Rendering}}}

 \\

{\footnotesize{50.59}}
 & 
{\footnotesize{51.84}}
 & 
{\footnotesize{\textbf{52.01}}}
 & 
{\footnotesize{50.52}}
 &  &  & 
{\footnotesize{47.77}}
 & 
{\footnotesize{\textbf{49.74}}}
 & 
{\footnotesize{49.66}}
 & 
{\footnotesize{48.34}}
 &  \\

\includegraphics[trim={50 50 50 50},clip,width=0.077\textwidth]{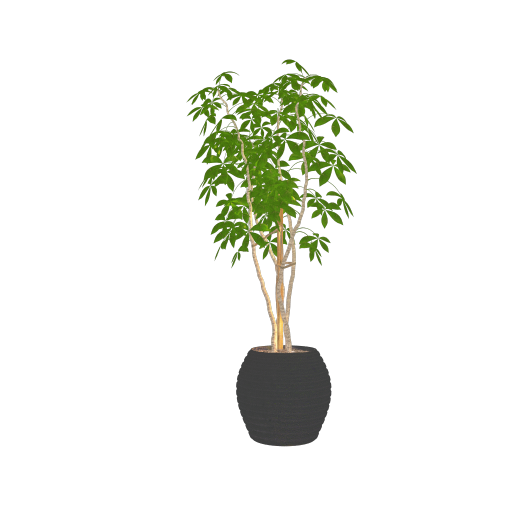}
 & 
\includegraphics[trim={50 50 50 50},clip,width=0.077\textwidth]{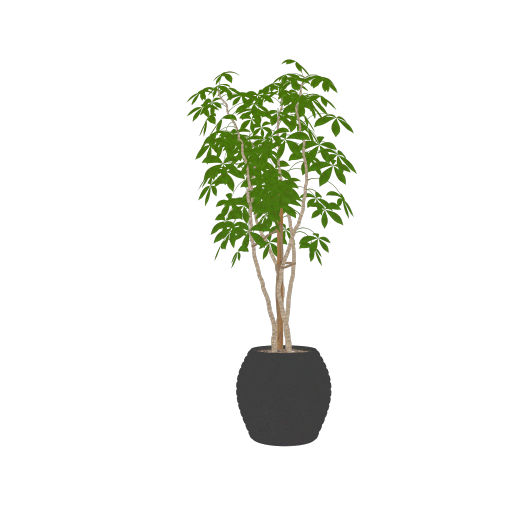}
 & 
\includegraphics[trim={50 50 50 50},clip,width=0.077\textwidth]{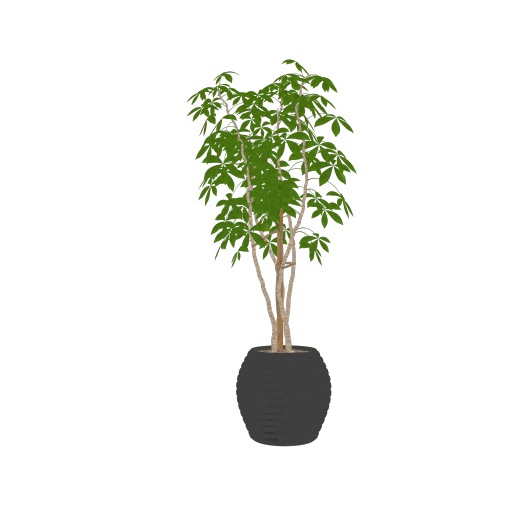}
 & 
\includegraphics[trim={50 50 50 50},clip,width=0.077\textwidth]{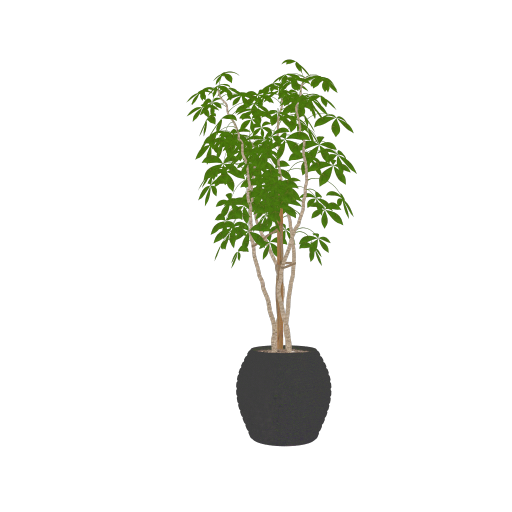}
 & 
\includegraphics[trim={50 50 50 50},clip,width=0.077\textwidth]{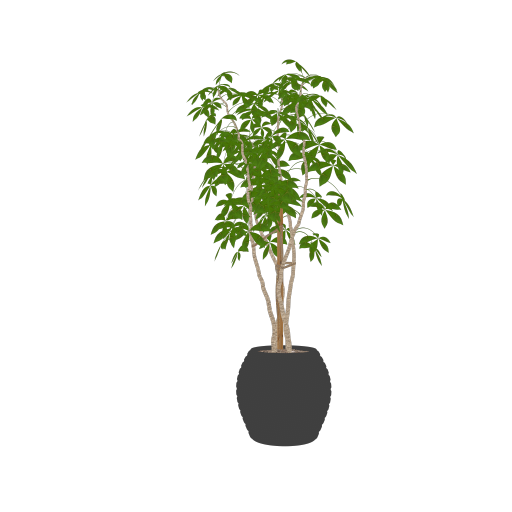}
 &  & 
\includegraphics[trim={50 50 50 50},clip,width=0.077\textwidth]{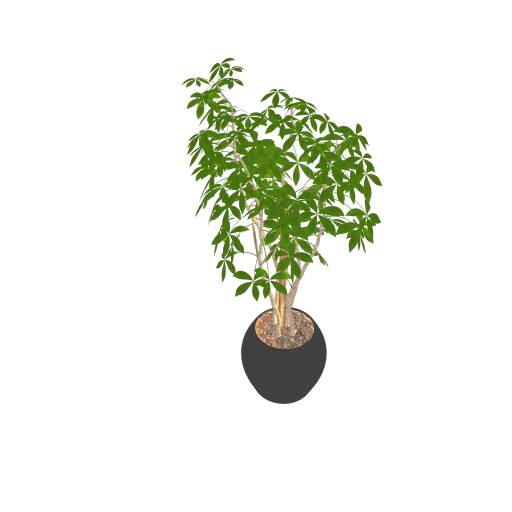}
 & 
\includegraphics[trim={50 50 50 50},clip,width=0.077\textwidth]{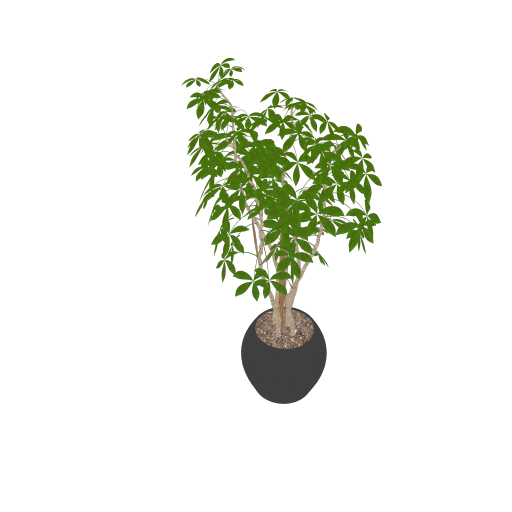}
 & 
\includegraphics[trim={50 50 50 50},clip,width=0.077\textwidth]{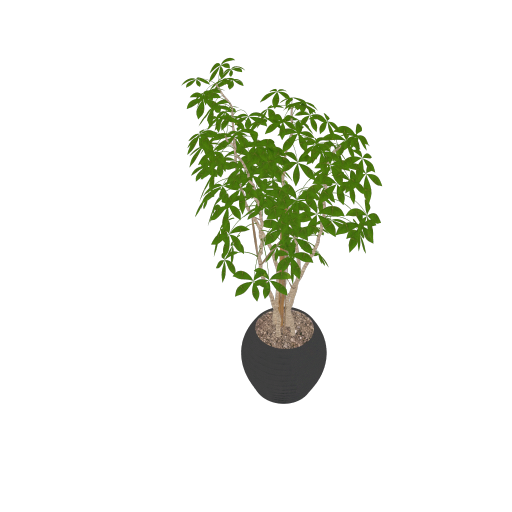}
 & 
\includegraphics[trim={50 50 50 50},clip,width=0.077\textwidth]{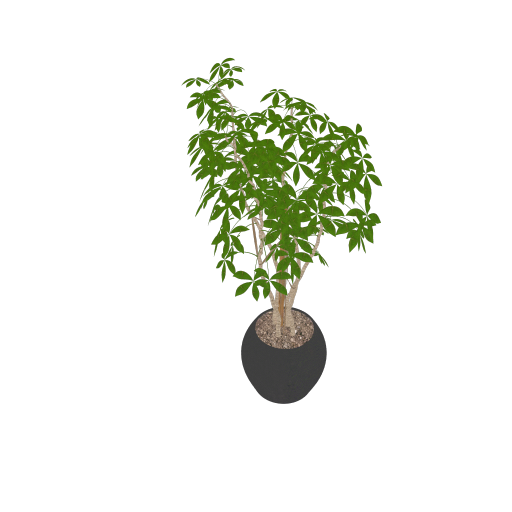}
 & 
\includegraphics[trim={50 50 50 50},clip,width=0.077\textwidth]{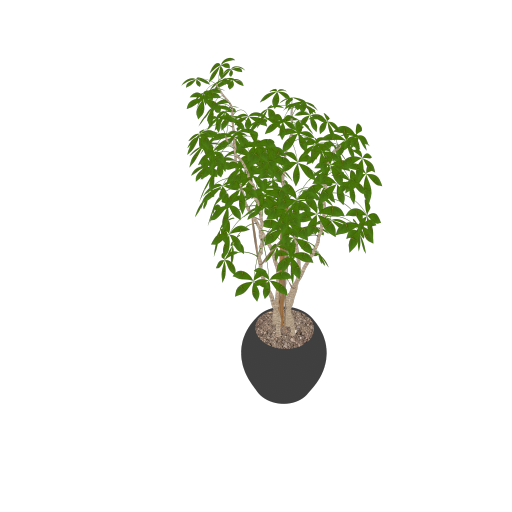}

\makebox[5pt]{\rotatebox{-90}{\hspace{-33pt}\footnotesize{Albedo}}}

 \\

{\footnotesize{35.22}}
 & 
{\footnotesize{52.48}}
 & 
{\footnotesize{51.24}}
 & 
{\footnotesize{\textbf{52.51}}}
 &  &  & 
{\footnotesize{36.15}}
 & 
{\footnotesize{\textbf{53.19}}}
 & 
{\footnotesize{51.29}}
 & 
{\footnotesize{53.10}}
 &  \\

\includegraphics[trim={50 50 50 50},clip,width=0.077\textwidth]{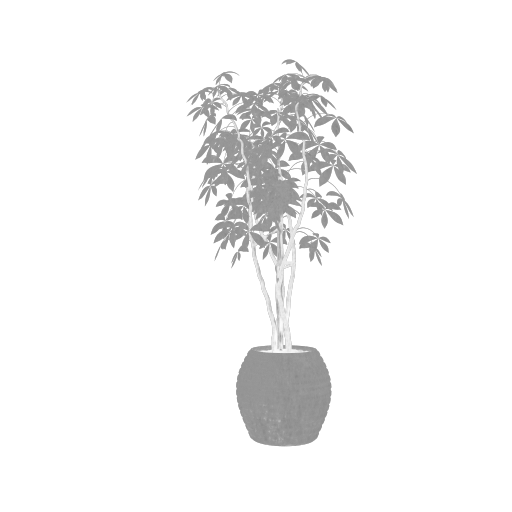}
 & 
\includegraphics[trim={50 50 50 50},clip,width=0.077\textwidth]{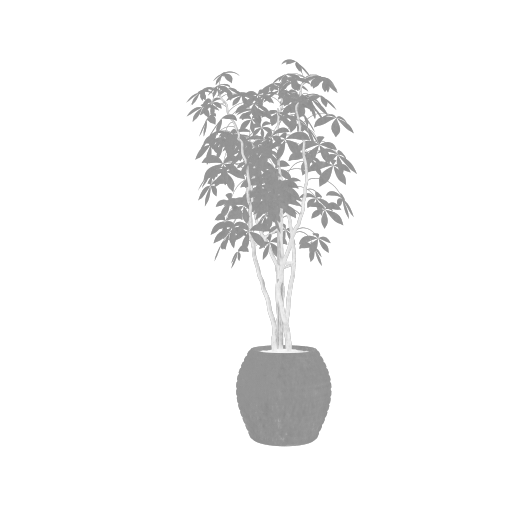}
 & 
\includegraphics[trim={50 50 50 50},clip,width=0.077\textwidth]{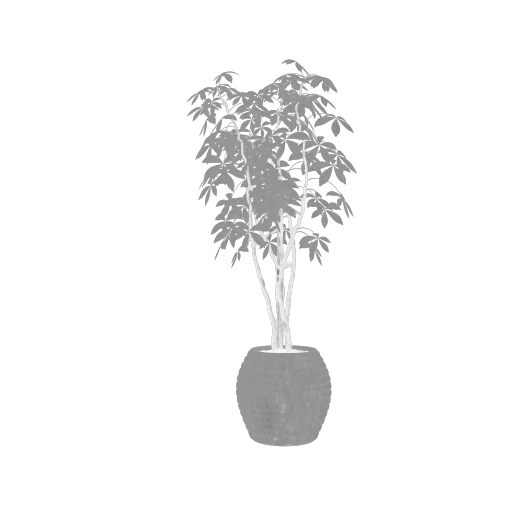}
 & 
\includegraphics[trim={50 50 50 50},clip,width=0.077\textwidth]{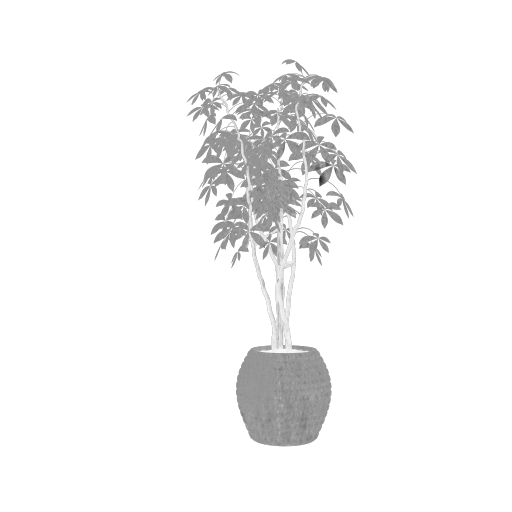}
 & 
\includegraphics[trim={50 50 50 50},clip,width=0.077\textwidth]{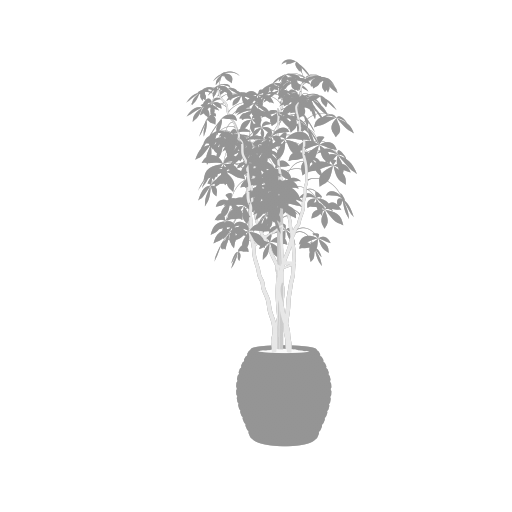}
 &  & 
\includegraphics[trim={50 50 50 50},clip,width=0.077\textwidth]{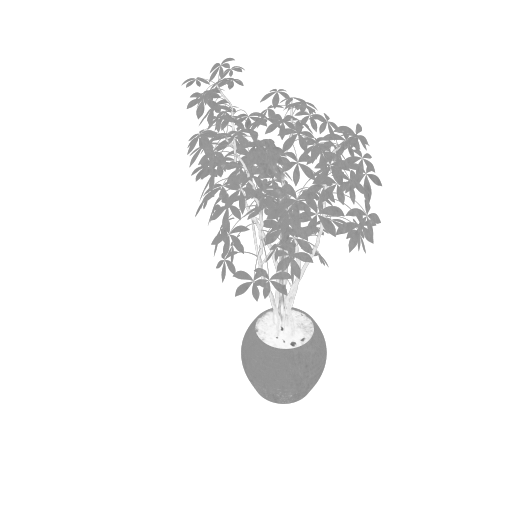}
 & 
\includegraphics[trim={50 50 50 50},clip,width=0.077\textwidth]{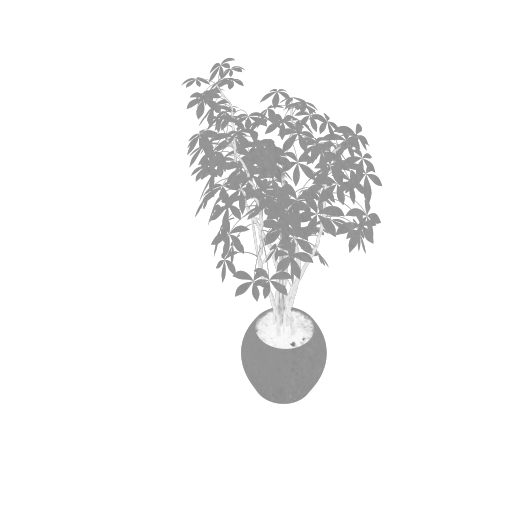}
 & 
\includegraphics[trim={50 50 50 50},clip,width=0.077\textwidth]{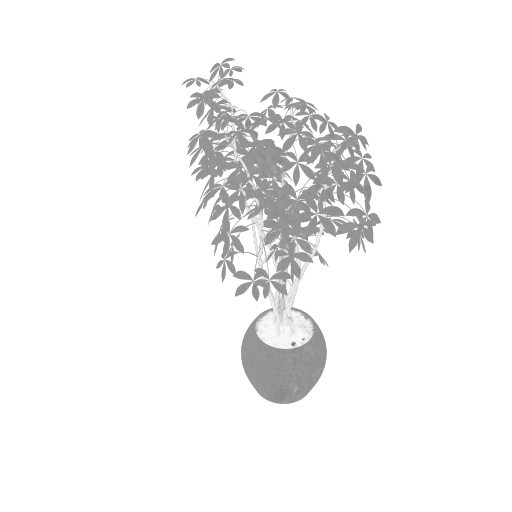}
 & 
\includegraphics[trim={50 50 50 50},clip,width=0.077\textwidth]{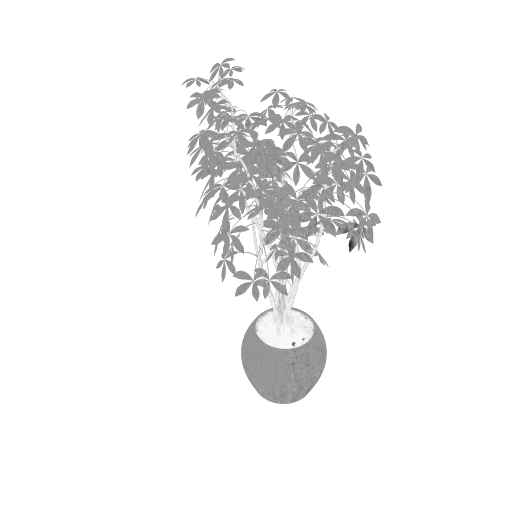}
 & 
\includegraphics[trim={50 50 50 50},clip,width=0.077\textwidth]{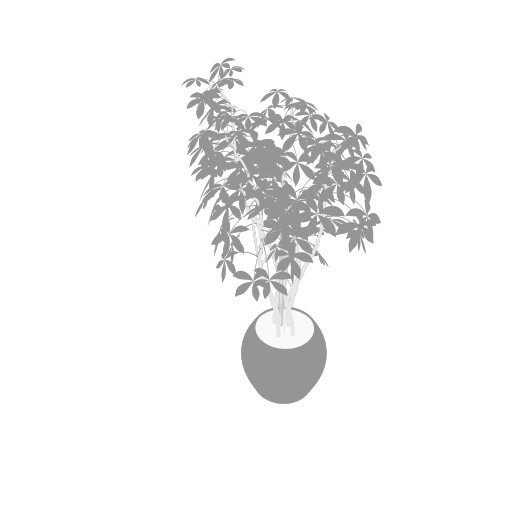}

\makebox[5pt]{\rotatebox{-90}{\hspace{-36pt}\footnotesize{Roughness}}}

 \\

{\footnotesize{37.99}}
 & 
{\footnotesize{\textbf{39.87}}}
 & 
{\footnotesize{39.09}}
 & 
{\footnotesize{38.46}}
 &  &  & 
{\footnotesize{35.75}}
 & 
{\footnotesize{36.09}}
 & 
{\footnotesize{36.32}}
 & 
{\footnotesize{\textbf{36.93}}}
 &  \\
\end{tabular}
\end{subfigure}

\endgroup

    \vspace{-0.3cm}
    \caption{
    \textbf{Main results for NeRF scenes.} \mycaption{For each scene, we compare the rendering, recovered albedo, and recovered roughness (top to bottom rows) for direct illumination, PRB, and AD-Ours. We also compare to the case where the radiance cache is trained without the prior. We show two different views of each scene, and report PSNR to ground truth (GT).}}
    \label{fig:results_nerf}
\end{figure*}

\section{Training Progress Curves}

We present how each method converges during training in Figure \ref{fig:recons} and \ref{fig:recons_nerf}.

\newcommand\cruve{0.33}

\begin{figure*}
\centering
\footnotesize

\begin{subfigure}{\textwidth}
    \centering
    \begin{subfigure}{\cruve\textwidth}
        \includegraphics[width=\textwidth]{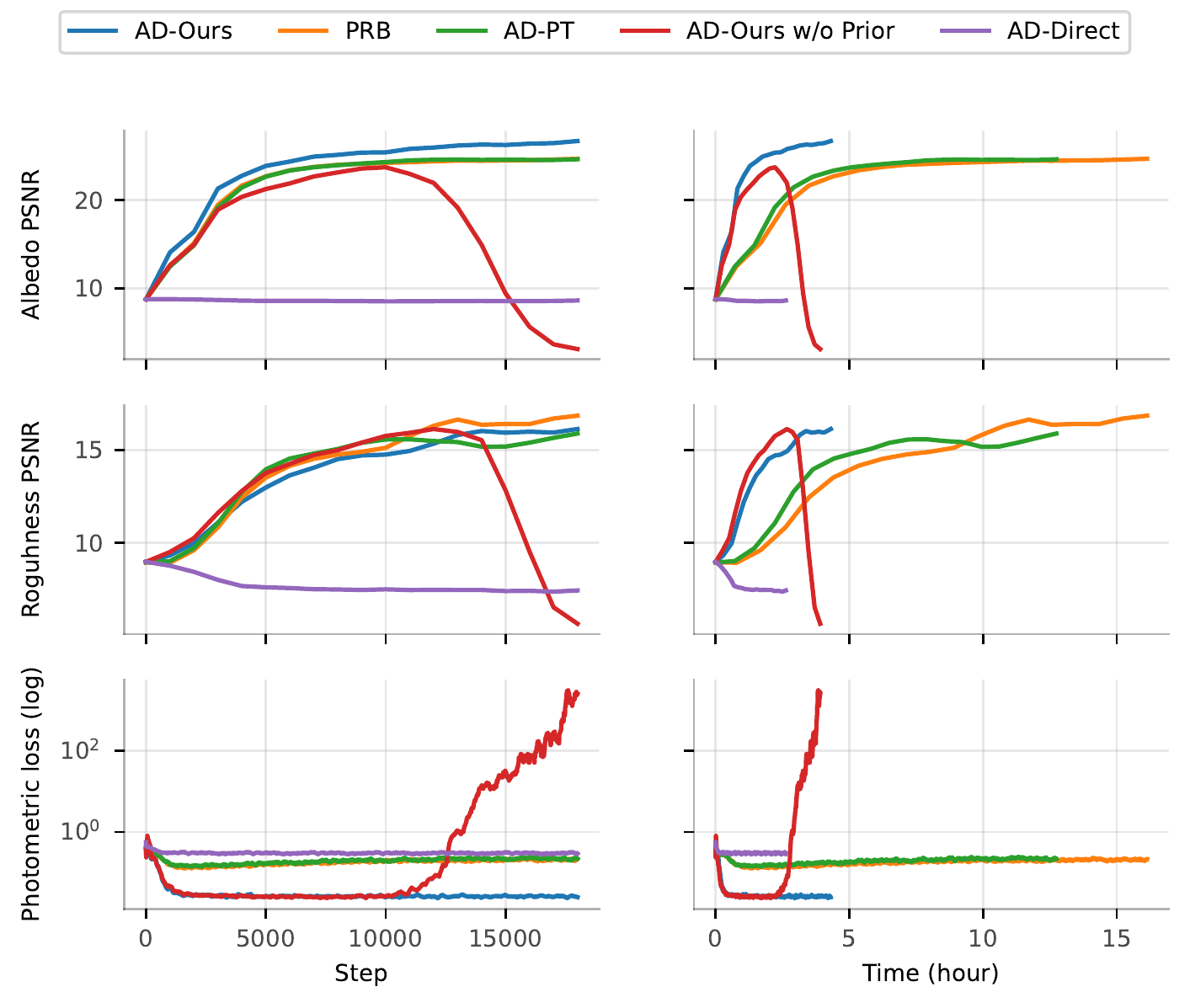}
        \caption{\Staircase}
    \end{subfigure}    
    \begin{subfigure}{\cruve\textwidth}
        \includegraphics[width=\textwidth]{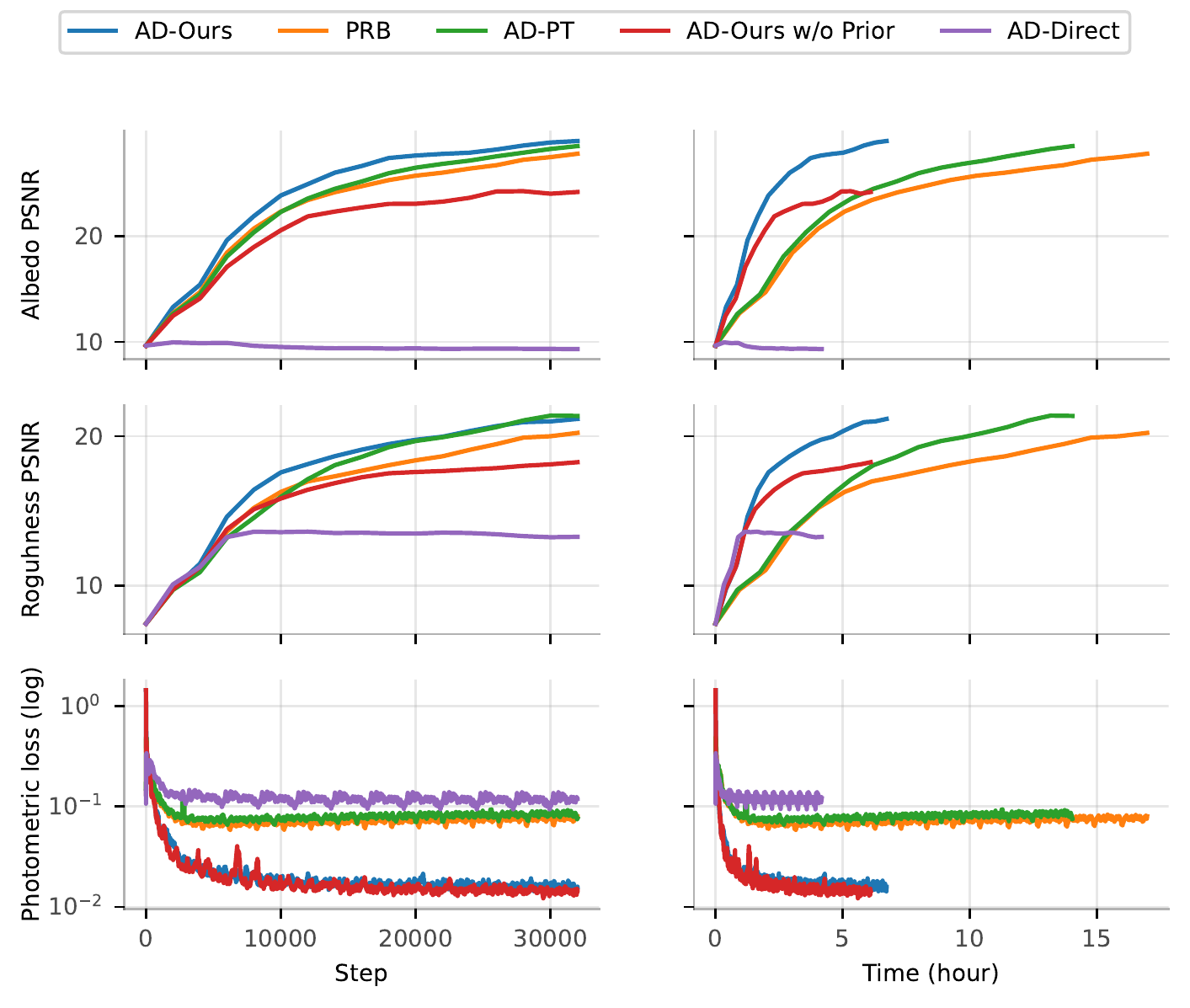}
        \caption{\Kitchen}
    \end{subfigure} \\
    
    \begin{subfigure}{\cruve\textwidth}
        \includegraphics[width=\textwidth]{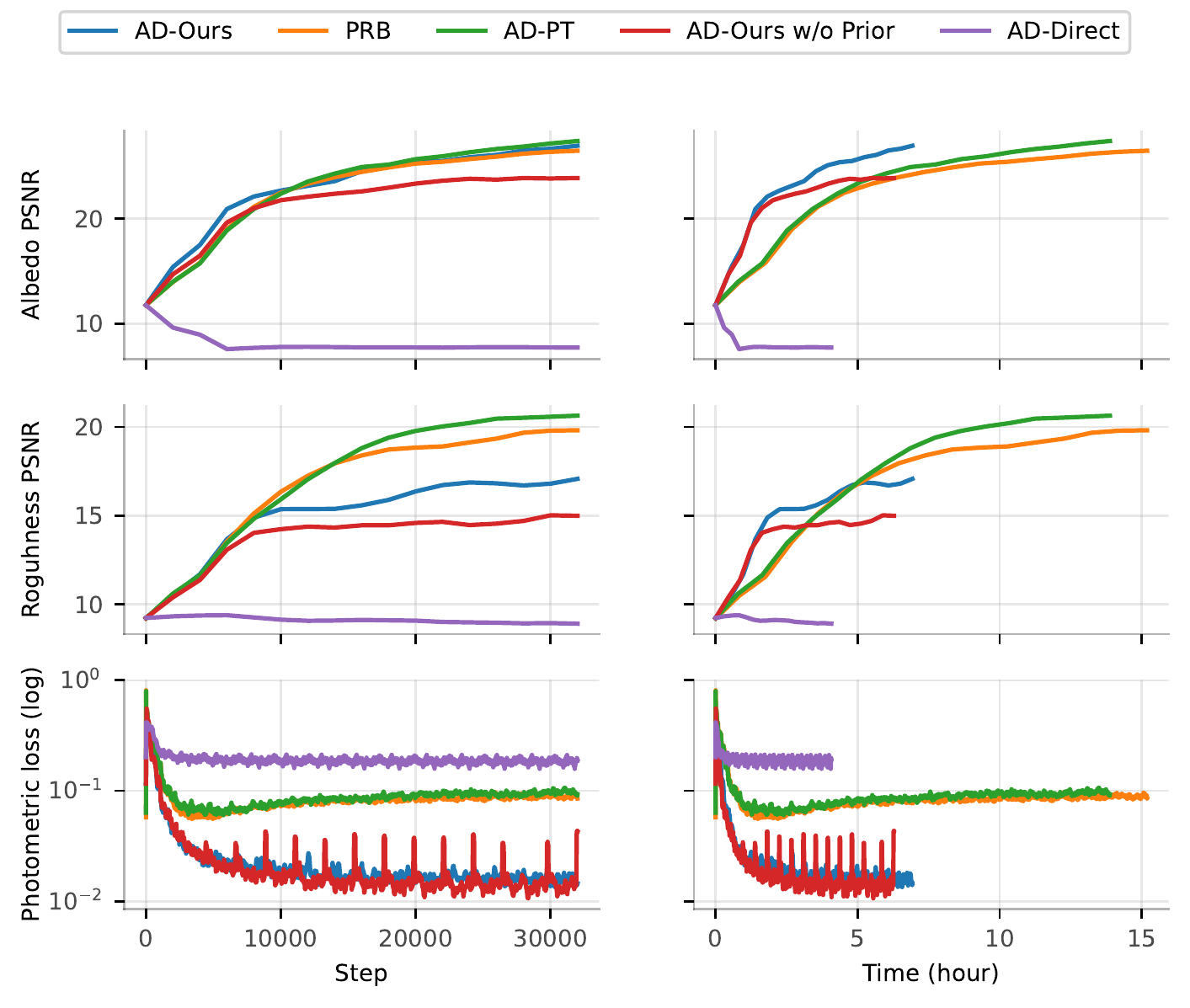}
        \caption{\LivingRoom}
    \end{subfigure}
    \begin{subfigure}{\cruve\textwidth}
        \includegraphics[width=\textwidth]{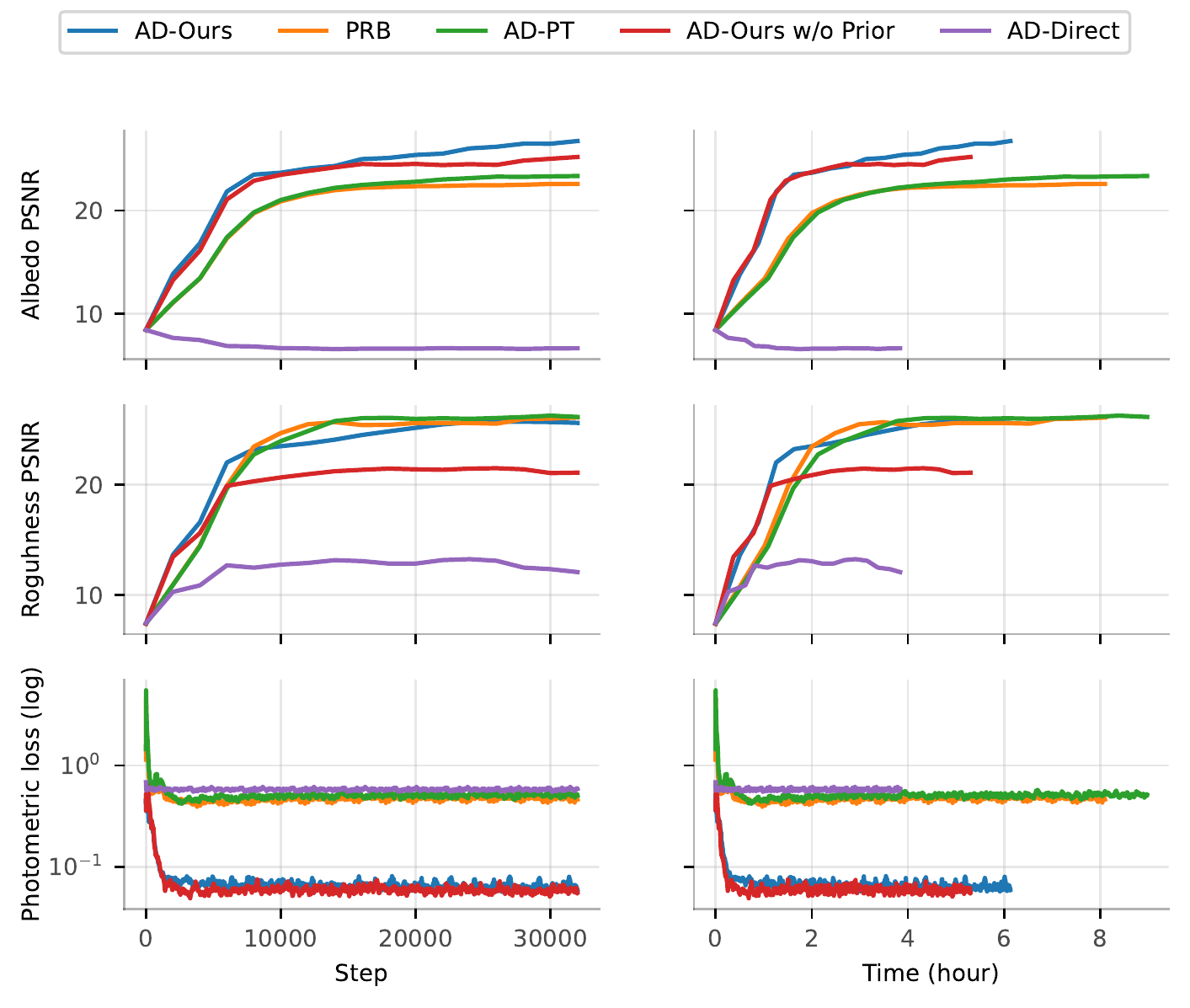}
        \caption{\VeachDoor}
    \end{subfigure}        
\end{subfigure}

\caption{ \textbf{Reconstruction accuracy}. \mycaption{Albedo, roughness and photometric error, for all the scenes, as a function of training steps and time. 
    Our method correctly accounts for global illumination thanks to our neural radiometric prior, resulting in comparable accuracy at low computational cost.}}
\label{fig:recons}
\end{figure*}

\begin{figure*}[ht]
\centering
\footnotesize

\begin{subfigure}{\textwidth}
    \centering
    \begin{subfigure}{\cruve\textwidth}
        \includegraphics[width=\textwidth]{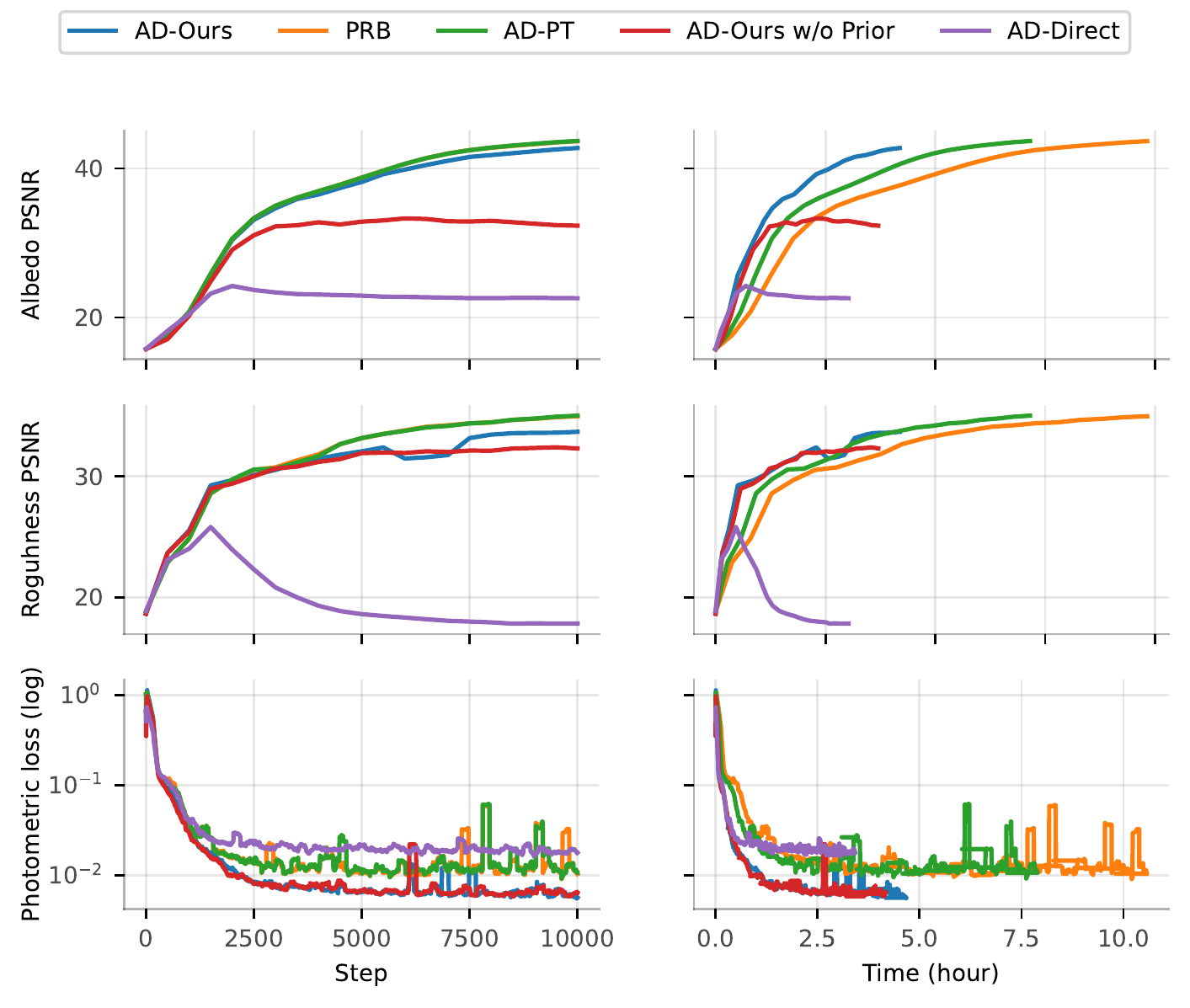}
        \caption{\Lego}
    \end{subfigure}    
    \begin{subfigure}{\cruve\textwidth}
        \includegraphics[width=\textwidth]{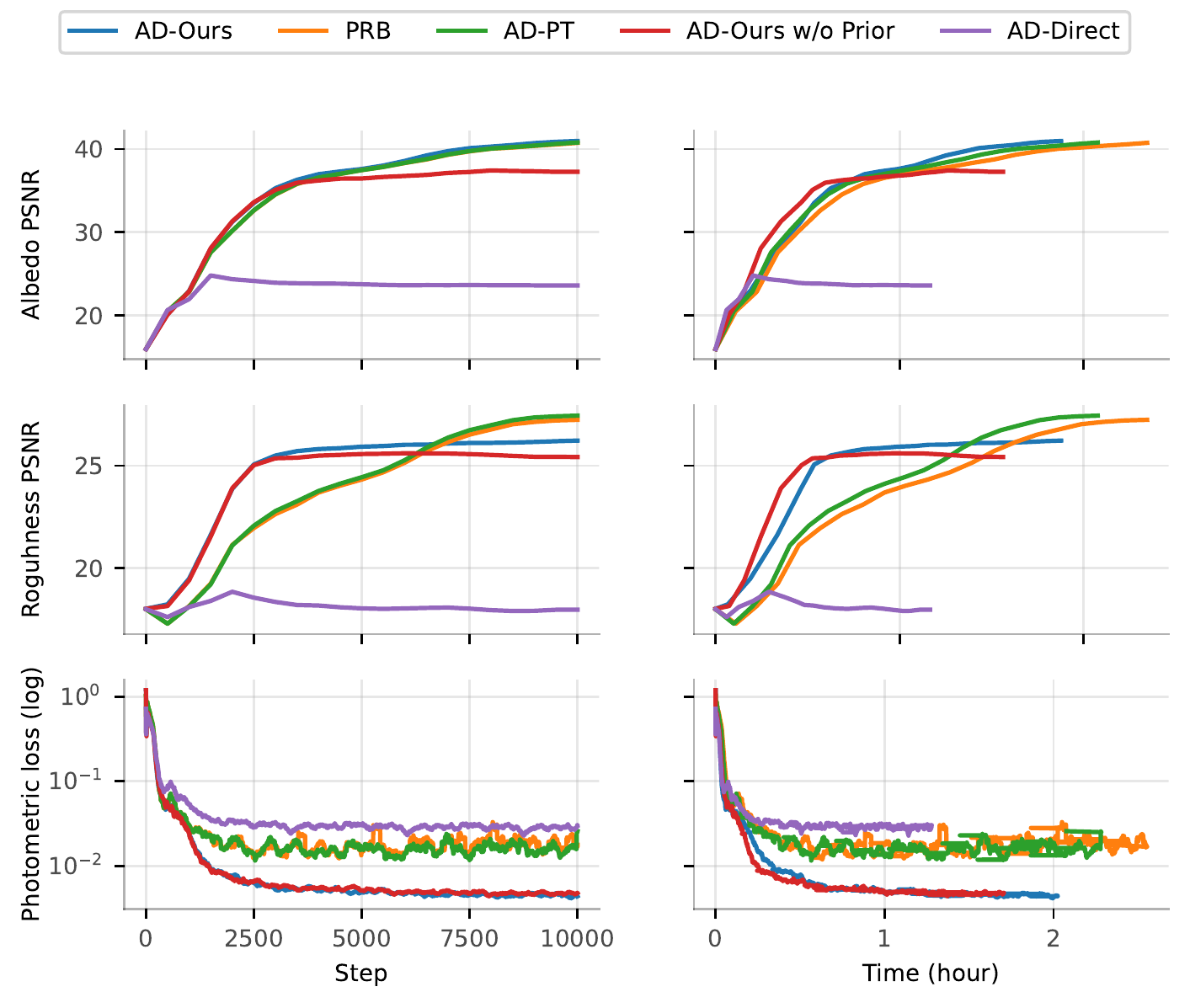}
        \caption{\Hotdog}
    \end{subfigure}    
    \begin{subfigure}{\cruve\textwidth}
        \includegraphics[width=\textwidth]{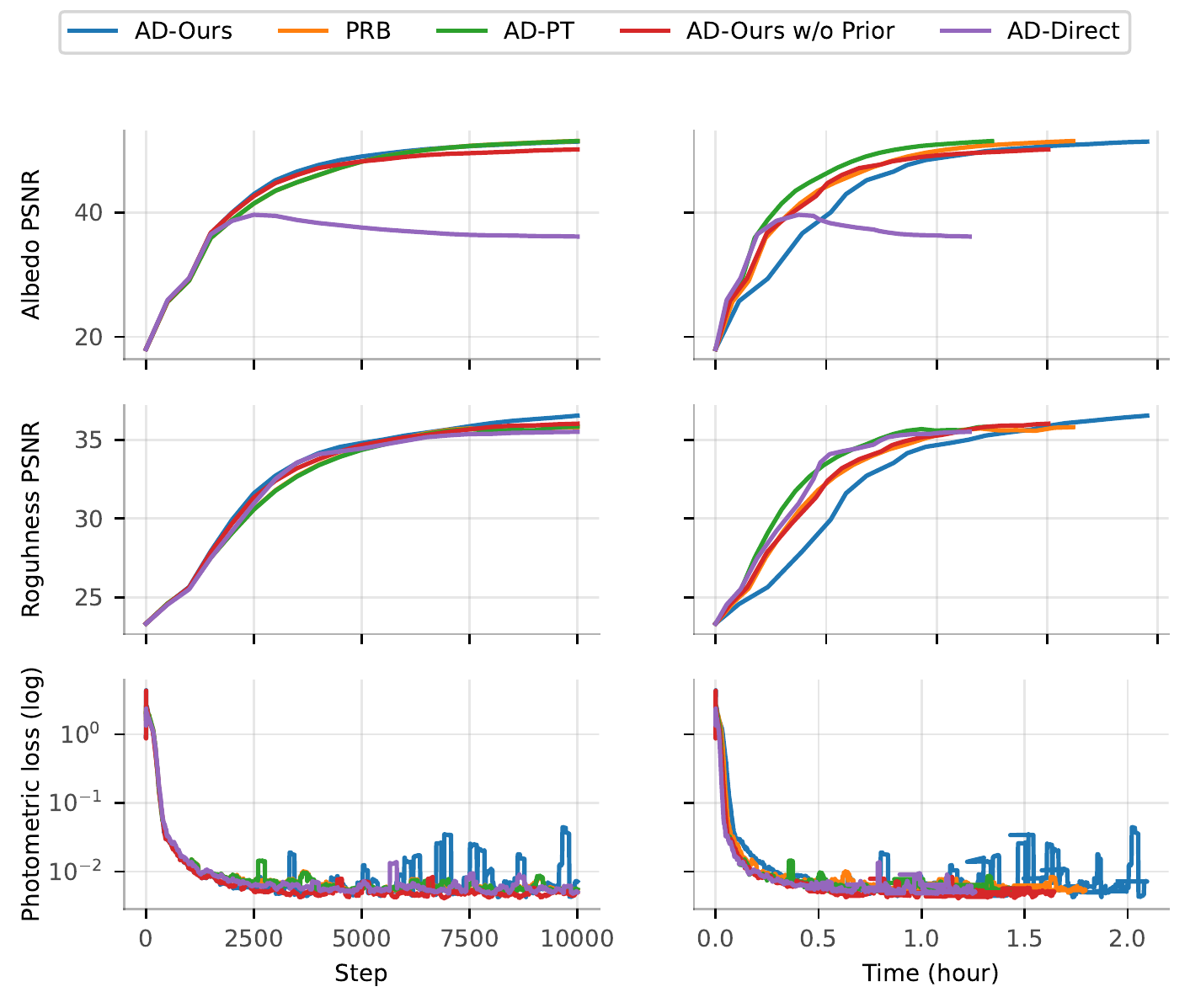}
        \caption{\Ficus}
    \end{subfigure}    
\end{subfigure}

\caption{ \textbf{Reconstruction accuracy for NeRF scenes}. \mycaption{Albedo, roughness and photometric error, for all the NeRF scenes,  as a function of training steps and time. 
    Our method correctly accounts for global illumination thanks to our neural radiometric prior, resulting in comparable accuracy at low computational cost.}}
\label{fig:recons_nerf}
\end{figure*}

\section{Ablation for more scenes}
We provide an ablation study for more scenes in Figure \ref{fig:suppl_ablation}.

\begin{figure*}
    \centering
    \captionsetup[subfigure]{labelformat=empty}
    \begingroup
\renewcommand{\arraystretch}{0.6}
\setlength{\tabcolsep}{0.1em}

\makebox[5pt]{\rotatebox{90}{\hspace{-10pt} \footnotesize{\Staircase}}}
\begin{subfigure}[b]{0.98\textwidth}
\begin{tabular}{cccccccc}

{\footnotesize{AD-Direct}}
 & 
{\footnotesize{\begin{tabular}{@{}c@{}}w/ Radiometric Prior \\ (Eq. (10)) \end{tabular}}}
 & 
{\footnotesize{\begin{tabular}{@{}c@{}}w/ Stop gradient \\ prior (Sec. 4.2)\end{tabular}}}
 & 
{\footnotesize{\begin{tabular}{@{}c@{}}w/ Second-bounce \\ prior (Sec. 4.2)\end{tabular}}}
 & 
{\footnotesize{\begin{tabular}{@{}c@{}}w/ LHS loss \\ (AD-Ours, Sec 4.3)\end{tabular}}}
 & 
{\footnotesize{\begin{tabular}{@{}c@{}}AD-Ours \\ w/o Prior\end{tabular}}}
 & 
{\footnotesize{GT}}
 \\

\includegraphics[width=0.135\textwidth]{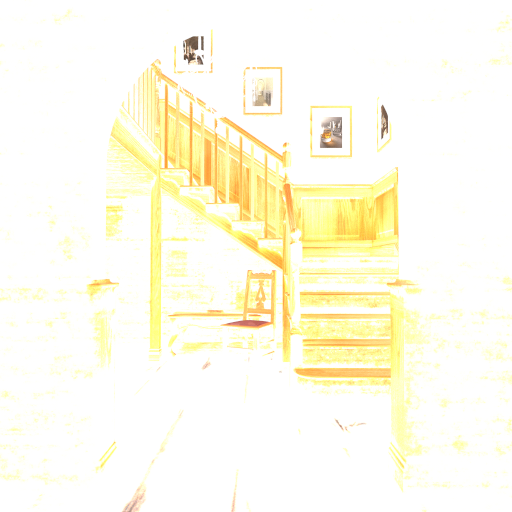}
 & 
\includegraphics[width=0.135\textwidth]{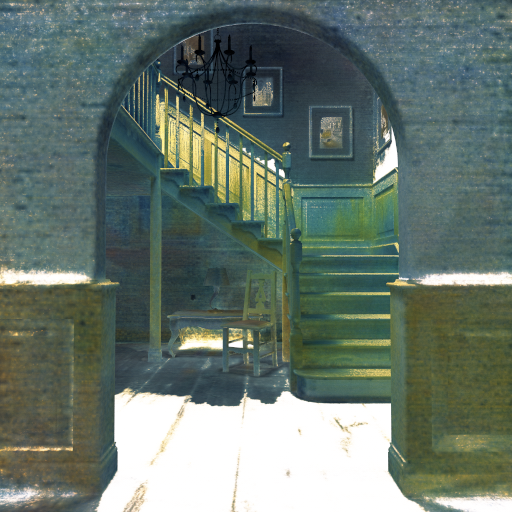}
 & 
\includegraphics[width=0.135\textwidth]{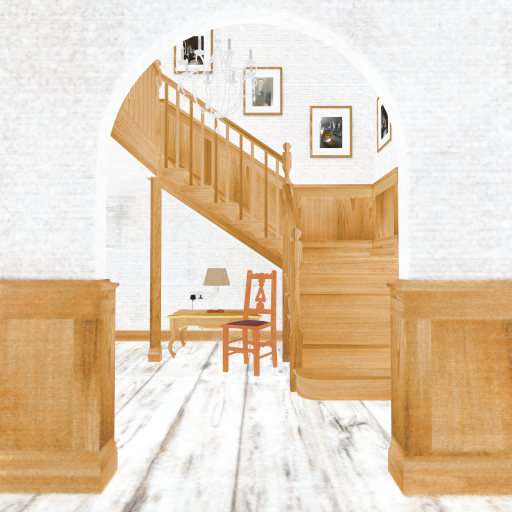}
 & 
\includegraphics[width=0.135\textwidth]{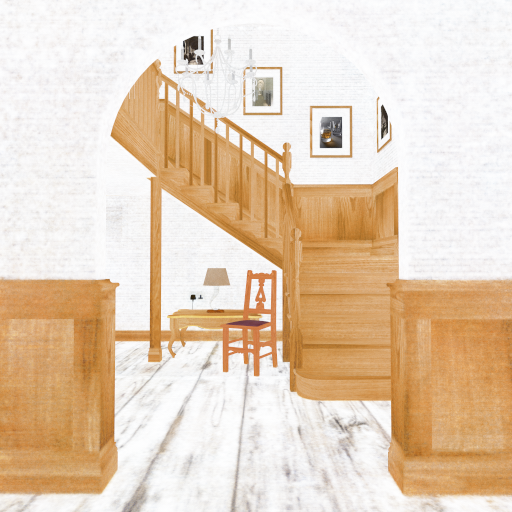}
 & 
\includegraphics[width=0.135\textwidth]{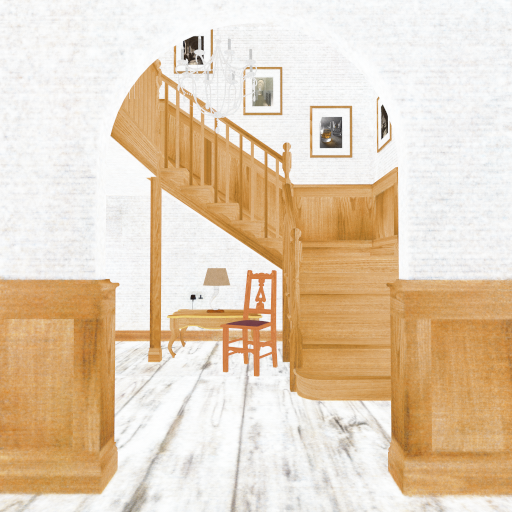}
 & 
\includegraphics[width=0.135\textwidth]{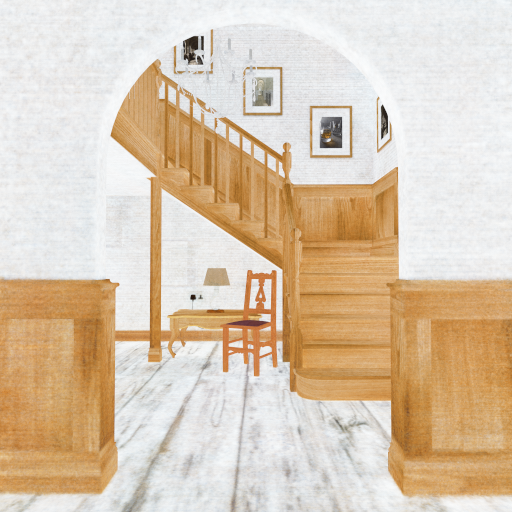}
 & 
\includegraphics[width=0.135\textwidth]{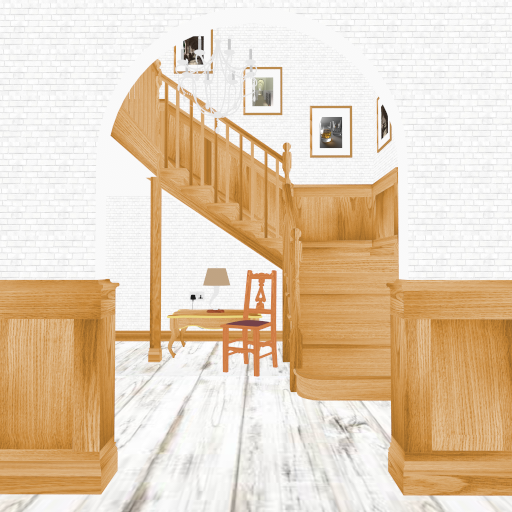}

\makebox[5pt]{\rotatebox{-90}{\hspace{-42pt}\footnotesize{Albedo}}}

 \\

{\footnotesize{MAPE: 1.1772}}
 & 
{\footnotesize{0.6864}}
 & 
{\footnotesize{0.0761}}
 & 
{\footnotesize{0.0668}}
 & 
{\footnotesize{\textbf{0.0657}}}
 & 
{\footnotesize{0.1162}}
 &  \\

\includegraphics[width=0.135\textwidth]{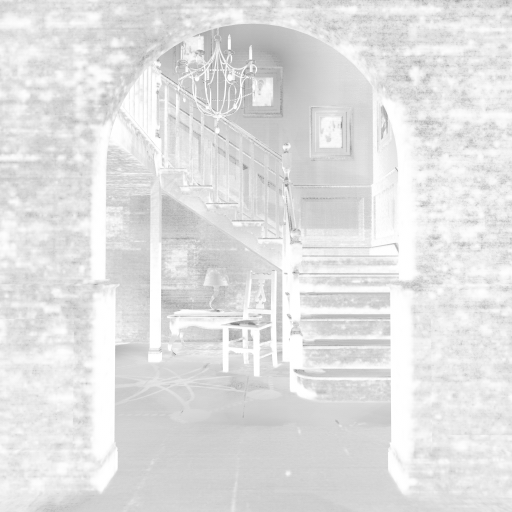}
 & 
\includegraphics[width=0.135\textwidth]{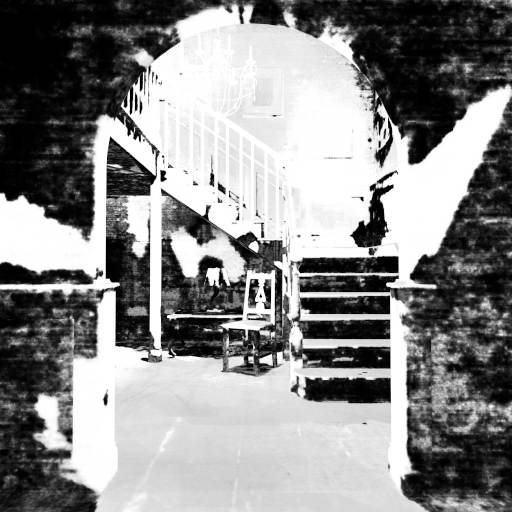}
 & 
\includegraphics[width=0.135\textwidth]{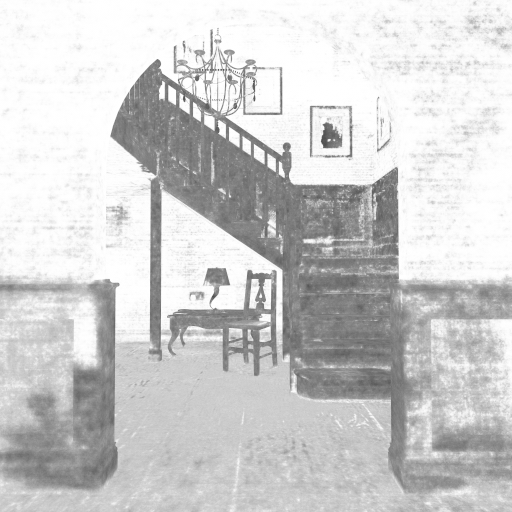}
 & 
\includegraphics[width=0.135\textwidth]{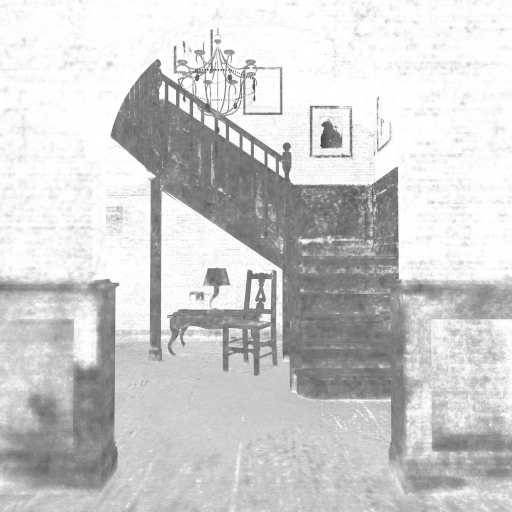}
 & 
\includegraphics[width=0.135\textwidth]{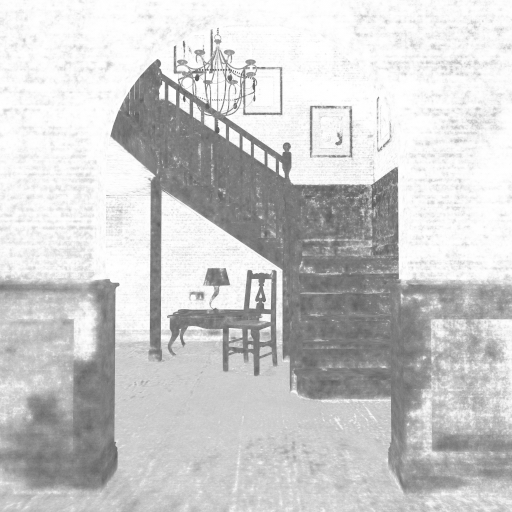}
 & 
\includegraphics[width=0.135\textwidth]{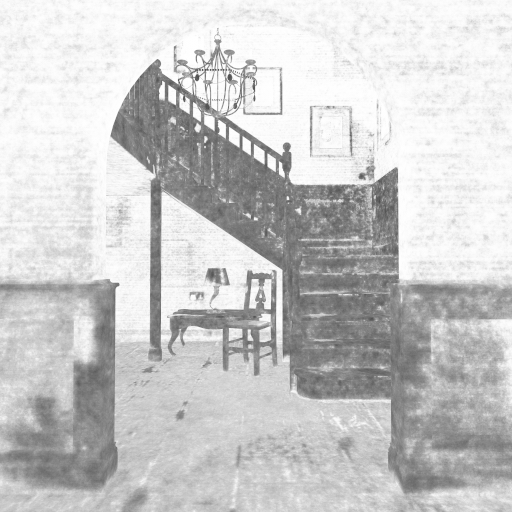}
 & 
\includegraphics[width=0.135\textwidth]{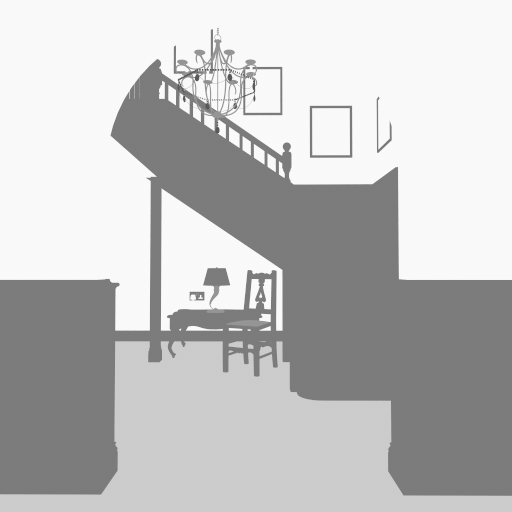}

\makebox[5pt]{\rotatebox{-90}{\hspace{-47pt}\footnotesize{Roughness}}}

 \\

{\footnotesize{1.2301}}
 & 
{\footnotesize{1.0540}}
 & 
{\footnotesize{0.3588}}
 & 
{\footnotesize{0.3558}}
 & 
{\footnotesize{\textbf{0.2985}}}
 & 
{\footnotesize{0.3102}}
 &  \\
\midrule
\end{tabular}
\end{subfigure}

\makebox[5pt]{\rotatebox{90}{\hspace{-10pt} \footnotesize{\Hotdog}}}
\begin{subfigure}[b]{0.98\textwidth}
\begin{tabular}{cccccccc}

\includegraphics[trim={50 50 50 50},clip,width=0.135\textwidth]{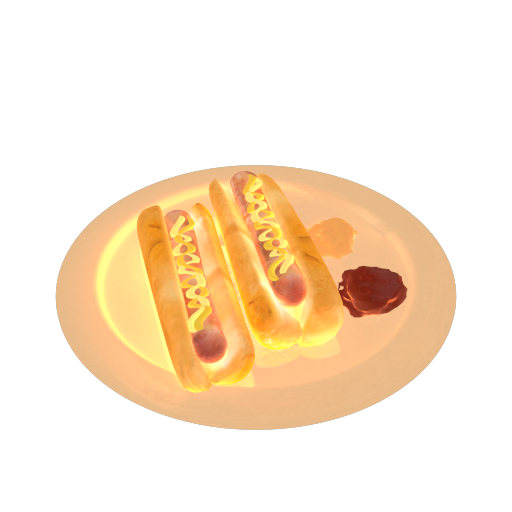}
 & 
\includegraphics[trim={50 50 50 50},clip,width=0.135\textwidth]{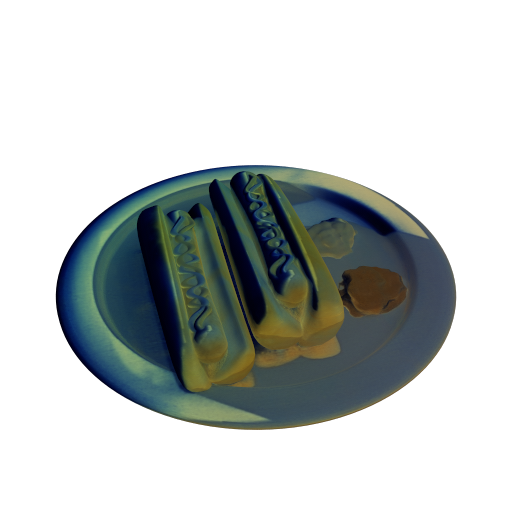}
 & 
\includegraphics[trim={50 50 50 50},clip,width=0.135\textwidth]{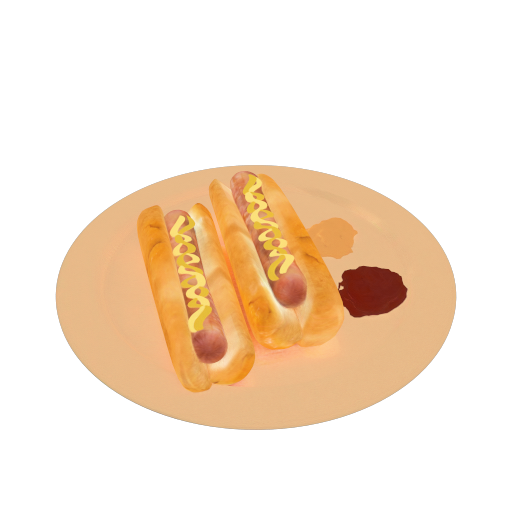}
 & 
\includegraphics[trim={50 50 50 50},clip,width=0.135\textwidth]{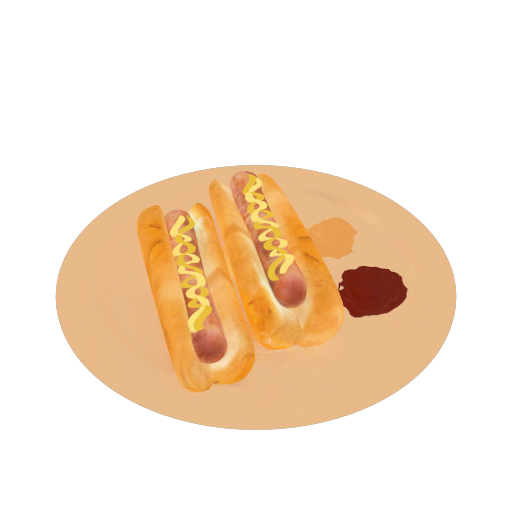}
 & 
\includegraphics[trim={50 50 50 50},clip,width=0.135\textwidth]{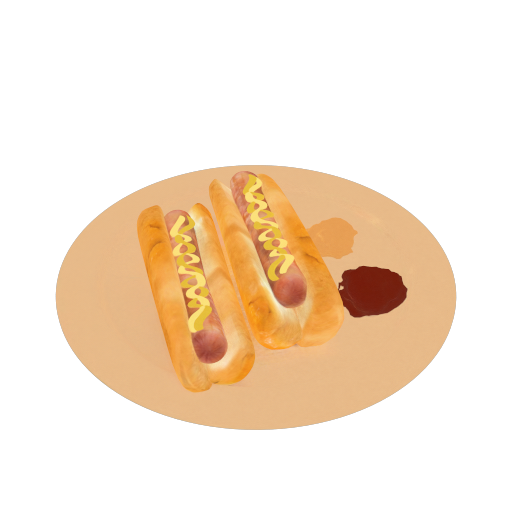}
 & 
\includegraphics[trim={50 50 50 50},clip,width=0.135\textwidth]{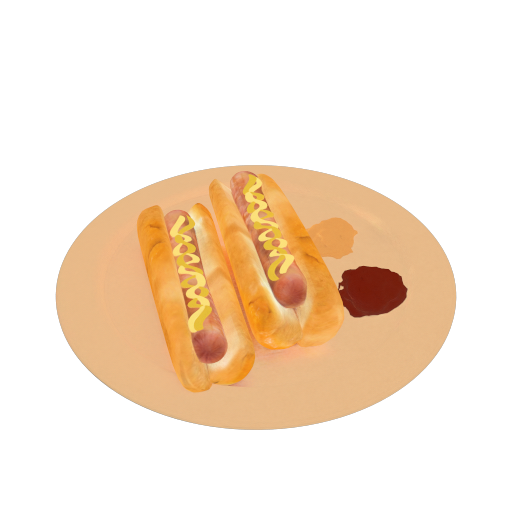}
 & 
\includegraphics[trim={50 50 50 50},clip,width=0.135\textwidth]{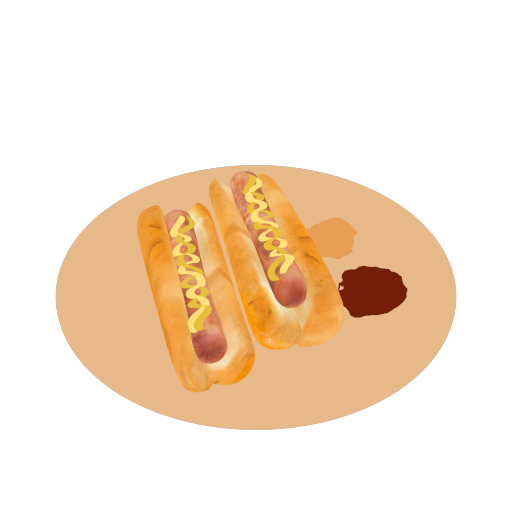}

\makebox[5pt]{\rotatebox{-90}{\hspace{-42pt}\footnotesize{Albedo}}}

 \\

{\footnotesize{MAPE: 0.0508}}
 & 
{\footnotesize{0.2343}}
 & 
{\footnotesize{0.0113}}
 & 
{\footnotesize{0.0091}}
 & 
{\footnotesize{\textbf{0.0080}}}
 & 
{\footnotesize{0.0097}}
 &  \\

\includegraphics[trim={50 50 50 50},clip,width=0.135\textwidth]{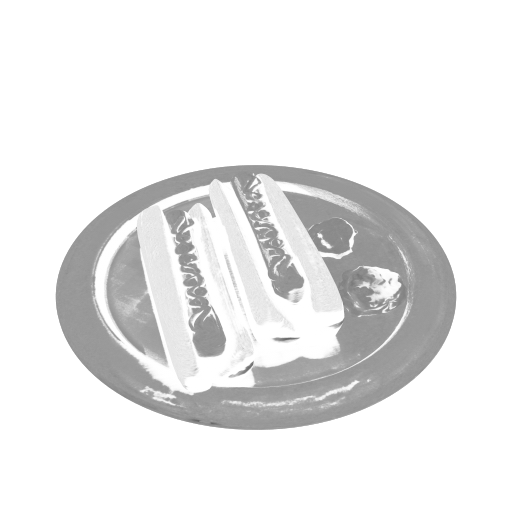}
 & 
\includegraphics[trim={50 50 50 50},clip,width=0.135\textwidth]{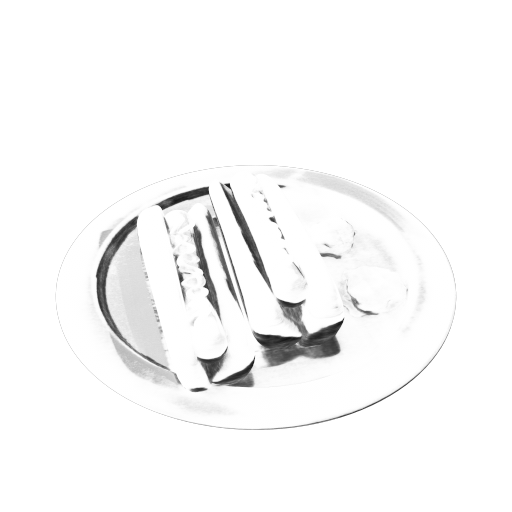}
 & 
\includegraphics[trim={50 50 50 50},clip,width=0.135\textwidth]{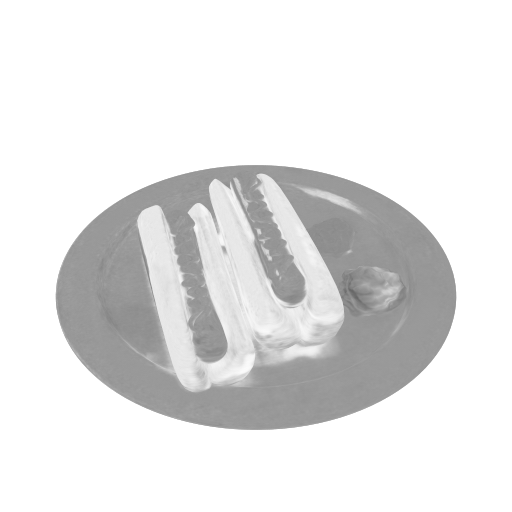}
 & 
\includegraphics[trim={50 50 50 50},clip,width=0.135\textwidth]{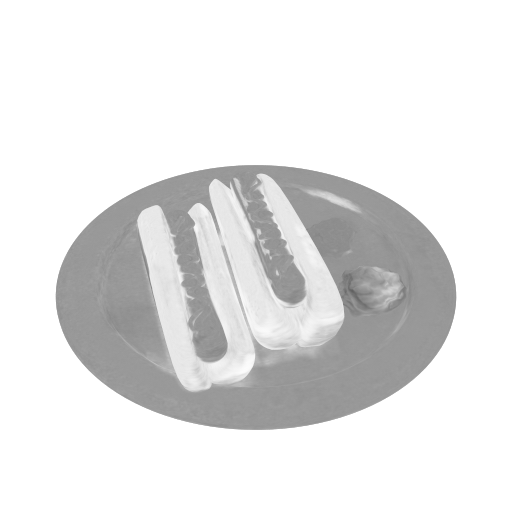}
 & 
\includegraphics[trim={50 50 50 50},clip,width=0.135\textwidth]{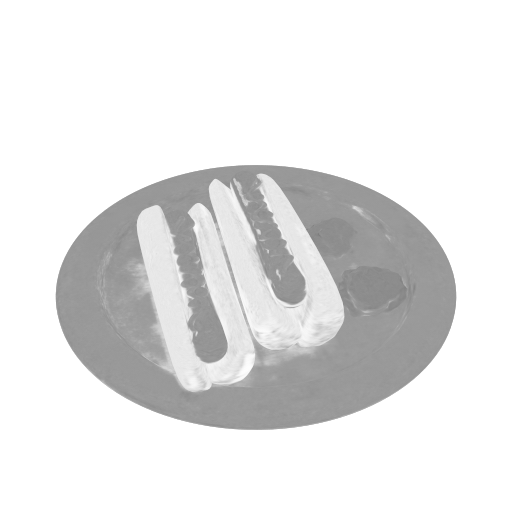}
 & 
\includegraphics[trim={50 50 50 50},clip,width=0.135\textwidth]{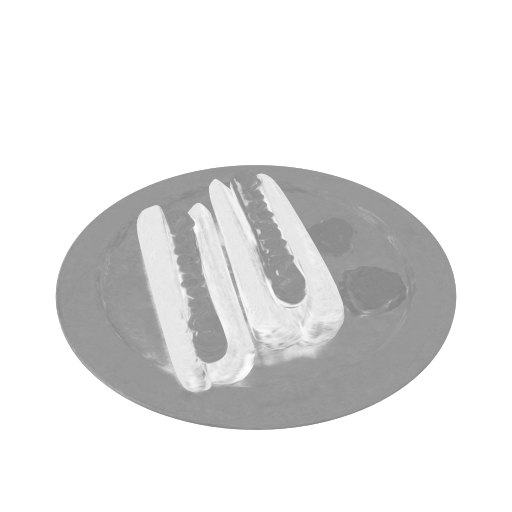}
 & 
\includegraphics[trim={50 50 50 50},clip,width=0.135\textwidth]{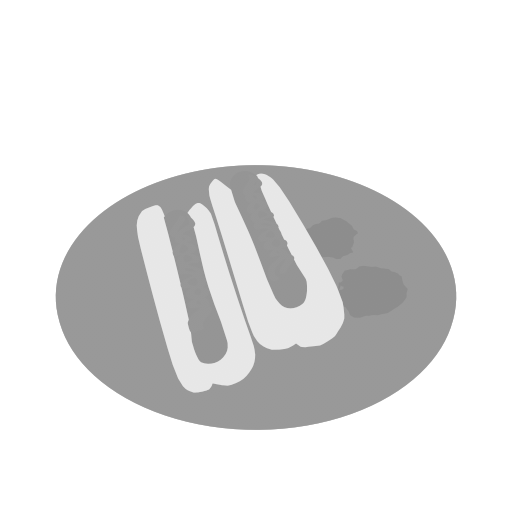}

\makebox[5pt]{\rotatebox{-90}{\hspace{-47pt}\footnotesize{Roughness}}}

 \\

{\footnotesize{0.1067}}
 & 
{\footnotesize{0.4695}}
 & 
{\footnotesize{0.0296}}
 & 
{\footnotesize{0.0245}}
 & 
{\footnotesize{\textbf{0.0215}}}
 & 
{\footnotesize{0.0218}}
 &  \\
\end{tabular}
\end{subfigure}

\endgroup

    \caption{{\textbf{Ablation for more scenes.}} \mycaption{We start with the direct illumination integrator (left), and add the radiometric prior to it. The results significantly improve when we ignore the gradients of the prior w.r.t scene parameters. Adding the prior to the second bounce better accounts for additional global illumination effects for areas unseen by the input cameras. Using ground truth data to improve the radiance field further improves the quality. Finally, the second column from the right shows our full method, except that we omit the prior.}}
\label{fig:suppl_ablation}
\end{figure*}

\fi
\end{document}